\documentclass[10pt,twocolumn,letterpaper]{article}

\usepackage{iccv}
\usepackage{times}
\usepackage{epsfig}
\usepackage{graphicx}
\usepackage{amsmath}
\usepackage{amssymb}
\usepackage{multirow}
\usepackage{comment}
\usepackage{booktabs}
\usepackage{enumitem}
\usepackage{xcolor}
\usepackage[symbol]{footmisc}
\usepackage[accsupp]{axessibility} %
\renewcommand{\thefootnote}{\fnsymbol{footnote}}

\definecolor{Mycolor1}{HTML}{FD836E}
\definecolor{Mycolor2}{HTML}{46CDD0}
\newcommand{\OURS}{Text2Room}
\newcommand{\mypar}[1]{\vspace{1mm}\noindent\textbf{#1.}}
\newcommand{\rwpar}[1]{\vspace{1mm}\noindent\textbf{#1}}

\usepackage[pagebackref=true,breaklinks=true,letterpaper=true,colorlinks,bookmarks=false]{hyperref}

\iccvfinalcopy

\begin{document}

\title{\OURS: Extracting Textured 3D Meshes from 2D Text-to-Image Models}

\author{
Lukas H{\"o}llein$^{1*}$ \quad Ang Cao$^{2*}$ \quad Andrew Owens$^2$ \quad Justin Johnson$^2$ \quad Matthias Nie{\ss}ner$^1$ \\
$^1$Technical University of Munich \quad $^2$University of Michigan
}

\ificcvfinal\thispagestyle{empty}\fi
{\let\thefootnote\relax\footnotetext{{$^*$ joint first authorship}}}

\begin{figure}

\twocolumn[{
\renewcommand\twocolumn[1][]{#1}
\maketitle

\centering
\setlength\tabcolsep{1pt}
\begin{tabular}{cc}
\includegraphics[width=0.766\textwidth]{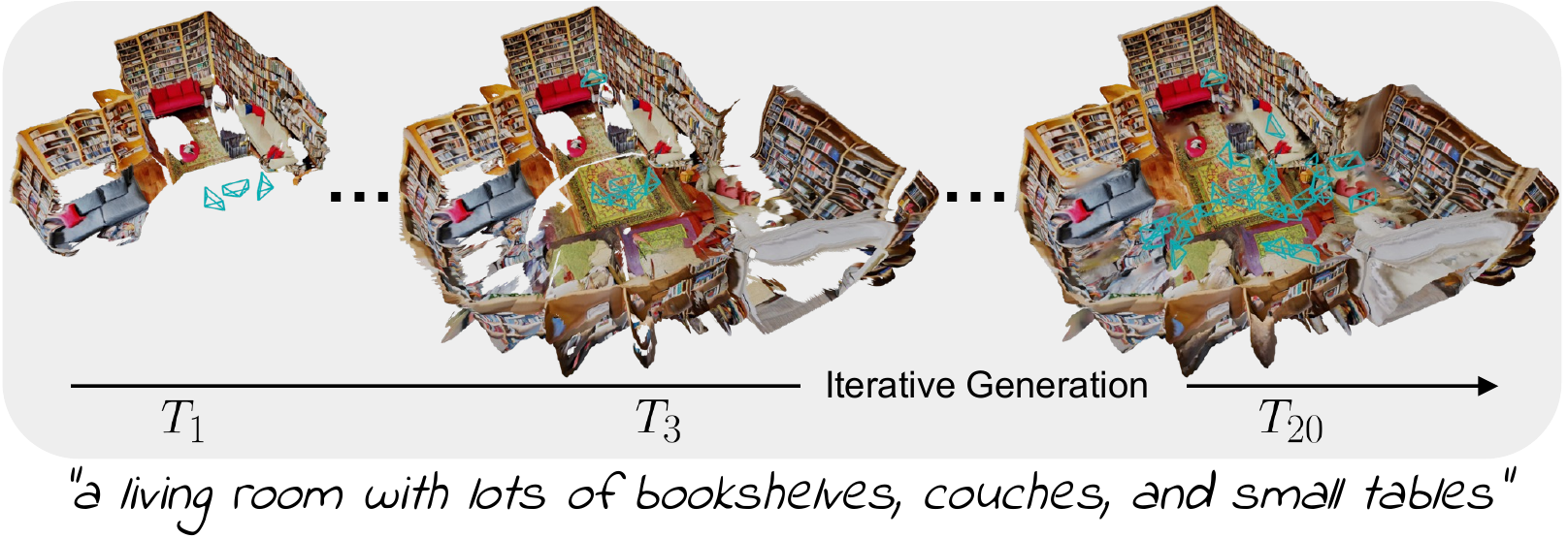} &
\includegraphics[width=0.233\textwidth]{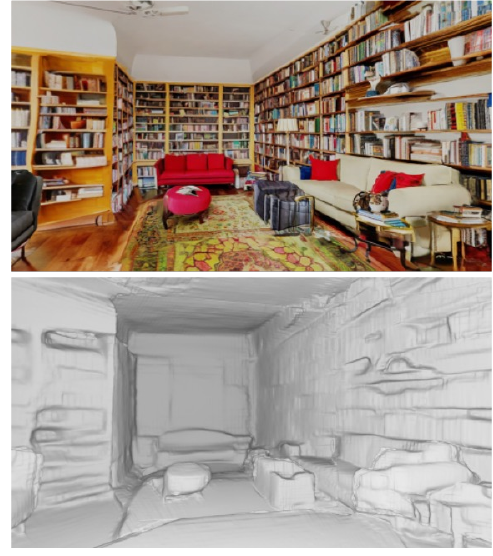} \\
(a) 3D Mesh Generation from Text &(b) Rendered Image + Mesh
\end{tabular}
\vspace{1mm}
\caption{
\textbf{Textured 3D mesh generation from text prompts.}
We generate textured 3D meshes from a given text prompt using 2D text-to-image models.
(a) The scene is iteratively created from different viewpoints (marked in blue).
(b) Our generated mesh contains compelling textures and geometry.
We remove the ceiling in the top-down views for better visualization of the scene layout.
}
\vspace{3mm}
\label{fig:teaser}
}]
\end{figure}

\begin{abstract}
We present \OURS \footnote[2]{\url{https://lukashoel.github.io/text-to-room}}, a method for generating room-scale textured 3D meshes from a given text prompt as input.
To this end, we leverage pre-trained 2D text-to-image models to synthesize a sequence of images from different poses.
In order to lift these outputs into a consistent 3D scene representation, we combine monocular depth estimation with a text-conditioned inpainting model.
The core idea of our approach is a tailored viewpoint selection such that the content of each image can be fused into a seamless, textured 3D mesh.
More specifically, we propose a continuous alignment strategy that iteratively fuses scene frames with the existing geometry to create a seamless mesh.
Unlike existing works that focus on generating single objects~\cite{poole2022dreamfusion, lin2022magic3d} or zoom-out trajectories~\cite{fridman2023scenescape} from text, our method generates complete 3D scenes with multiple objects and explicit 3D geometry.
We evaluate our approach using qualitative and quantitative metrics, demonstrating it as the first method to generate room-scale 3D geometry with compelling textures from only text as input.

\end{abstract}
\section{Introduction}

Mesh representations of 3D scenes are a crucial component for many applications, from AR/VR asset creation to computer graphics, yet creating these 3D assets remains a painstaking process that requires considerable expertise.
In the 2D domain, recent works have successfully created high-quality images from text using generative models, such as diffusion models~\cite{Rombach2021HighResolutionIS, Ramesh2022HierarchicalTI, Saharia2022PhotorealisticTD}.
These methods significantly reduce the barriers to creating images that contain a user's desired content, effectively helping towards the democratization of content creation.
An emerging line of work has sought to apply similar methods to create 3D models from text~\cite{Chen2018Text2ShapeGS,poole2022dreamfusion, jain2021dreamfields, lin2022magic3d, lee2022understanding}, yet existing approaches come with a number of significant limitations and lack the generality of 2D text-to-image models.

One of the core challenges of generating 3D models is coping with the lack of available 3D training data, as 3D datasets are vastly smaller than those available in many other applications, such as 2D image synthesis.
For example, methods that directly use 3D supervision, such as Chen~\etal~\cite{Chen2018Text2ShapeGS}, are often limited to datasets of simple shapes, such as ShapeNet~\cite{Chang2015ShapeNetAI}.
To address these data limitations, recent methods~\cite{poole2022dreamfusion, jain2021dreamfields, lin2022magic3d, lee2022understanding, wang2022score} lift the expressive power of 2D text-to-image models into 3D by formulating 3D generation as an iterative optimization problem in the image domain.
This allows them to generate 3D objects stored in a radiance field representation, demonstrating the ability to generate arbitrary (neural) shapes from text.
However, these methods cannot easily be extended to create room-scale 3D structure and texture.
The challenge of generating large scenes is ensuring that the generated output is dense and coherent across outward-facing viewpoints, and that these views contain all of the required structures, such as walls, floors, and furniture.
Additionally, a mesh remains a desired representation for many end-user tasks, such as rendering on commodity hardware (which requires an additional conversion step as presented in Lin~\etal~\cite{lin2022magic3d}).

To address these shortcomings, we propose a method that extracts scene-scale 3D meshes from off-the-shelf 2D text-to-image models. 
Our method iteratively generates a scene through inpainting and monocular depth estimation.
We produce an initial mesh by generating an image from text, and backproject it into 3D using a depth estimation model.
Then, we iteratively render the mesh from novel viewpoints.
From each one, we fill in holes in the rendered images via inpainting, then fuse the generated content into the mesh (Fig.~\ref{fig:teaser}a).

Our iterative generation scheme has two important design considerations: how we choose the viewpoints, and how we merge generated scene content with the existing mesh.
We first select viewpoints from predefined trajectories that will cover large amounts of scene content, then adaptively select viewpoints that close remaining holes.
When merging generated content with the mesh, we align the two depth maps to create smooth transitions, and remove parts of the mesh that contain distorted textures.
Together, these decisions lead to large, scene-scale 3D meshes with compelling textures and consistent geometry (Fig.~\ref{fig:teaser}b), that can represent a wide range of rooms.

\noindent To summarize, our contributions are:
\begin{itemize}[leftmargin=*,topsep=1pt, noitemsep]
    \item Generating 3D meshes of room-scale indoor scenes with compelling textures and geometry from any text input.
    \item A method that leverages 2D text-to-image models and monocular depth estimation to lift frames into 3D in an iterative scene generation. Our proposed depth alignment and mesh fusion steps, enable us to create seamless and undistorted geometry and textures.
    \item A two-stage tailored viewpoint selection that samples camera poses from optimal positions to first create the room layout and furniture and then close any remaining holes, creating a watertight mesh.
\end{itemize}

\section{Related Work}

\par \noindent{\bf Text-based Generation}
has seen significant advances due to large-scale image-text datasets~\cite{sharma2018conceptual, Schuhmann2021LAION400MOD, Desai2021RedCapsWI, Schuhmann2022LAION5BAO} and scalable generative model architectures~\cite{Esser2020TamingTF, Ronneberger2015UNetCN, Razavi2019GeneratingDH, Karras2017ProgressiveGO}, enabling synthesis of novel images from text~\cite{Gu2021VectorQD, Avrahami2021BlendedDF, Patashnik2021StyleCLIPTM}.

Recently, diffusion models~\cite{sohl2015deep, Ho2020DenoisingDP, Song2019GenerativeMB, Song2020ImprovedTF, Song2020ScoreBasedGM} achieved impressive results on image synthesis~\cite{Dhariwal2021DiffusionMB, Rombach2021HighResolutionIS, Saharia2022PhotorealisticTD, Nichol2021GLIDETP, Ramesh2022HierarchicalTI} through improvements like latent space denoising~\cite{Rombach2021HighResolutionIS, Vahdat2021ScorebasedGM}, faster sampling~\cite{Ho2020DenoisingDP, Song2020DenoisingDI, Nichol2021ImprovedDD, Kong2021OnFS}, and better guidance~\cite{Ho2022ClassifierFreeDG}. 

In particular, \emph{text-to-image} methods like Stable Diffusion~\cite{Rombach2021HighResolutionIS}, Imagen~\cite{Saharia2022PhotorealisticTD}, GLIDE~\cite{Nichol2021GLIDETP} and DALL$\cdot$E 2~\cite{Ramesh2022HierarchicalTI} yield diverse, high-fidelity, and controllable~\cite{Brooks2022InstructPix2PixLT,zhang2023adding} outputs.
Text-based generation has been extended to other modalities including audio~\cite{Kong2020DiffWaveAV, Forsgren_Martiros_2022, Huang2023Noise2MusicTM, Schneider2023MosaiTG}, video~\cite{Singer2022MakeAVideoTG, Wu2022TuneAVideoOT, villegas2022phenaki, Ho2022VideoDM}, and 4D fields~\cite{Singer2023TextTo4DDS}.
We use \emph{text-to-image} models by lifting their generated output into complete 3D scene meshes.

\rwpar{Text-to-3D.}
Several methods use 3D data for supervised training of text-to-3D models~\cite{Chen2018Text2ShapeGS, nichol2022point, Bautista2022GAUDIAN};
however this direction remains challenging due to the lack of large-scale aligned datasets of text and 3D.

Alternative approaches use 2D vision-language models like CLIP~\cite{radford2021learning} to create 3D content by formulating the generation as an optimization problem in the image domain~\cite{Wang2021CLIPNeRFTD, jain2021dreamfields, lee2022understanding, khalid2022clipmesh, jiang20223d} or as object alignment~\cite{Sanghi2021CLIPForgeTZ}.
Related methods refine existing 3D input through text guidance in a similar fashion~\cite{michel2022text2mesh, chen2022tango, wang2022nerf, richardson2023texture}.

Recent methods~\cite{poole2022dreamfusion, lin2022magic3d, metzer2022latent, wang2022score, MelasKyriazi2023RealFusion3R} combine large text-to-image diffusion models~\cite{Rombach2021HighResolutionIS, Saharia2022PhotorealisticTD} and neural radiance fields~\cite{Mildenhall2020NeRFRS} to generate 3D objects without training.
Other approaches train custom diffusion models on a similar text-to-3D task~\cite{li20223ddesigner, nam20223d, cheng2022sdfusion}.
In contrast, we use a fixed text-to-image model and extract a 3D mesh representing entire scenes of many objects and structural elements like walls.

\rwpar{3D-Consistent View Synthesis from a Single Image.}
Several methods have been proposed that perform novel-view-synthesis from a single image~\cite{Rockwell2021PixelSynthGA, Wiles2019SynSinEV, Shih20203DPU, Ren2022LookOT, Gu2023NerfDiffSV}.
Others optimize a neural 3D representation of an object, that can be viewed from arbitrary novel view points~\cite{Xu_2022_neuralLift, watson2022novel, anciukevicius2022renderdiffusion}.
Another line of work performs \emph{perpetual view generation}~\cite{Sivic2008InfiniteIC, infinite_nature_2020, li2022_infinite_nature_zero, cai2022diffdreamer}, synthesizing videos via a \emph{render-refine-repeat} pattern from a single RGB image that depict a scene along a forward-facing camera trajectory.
In very recent concurrent work, Fridman~\etal~\cite{fridman2023scenescape} create 3D scenes from text, but focus on this type of 3D-consistent ``zoom-out'' video generation.
Instead, we generate complete, textured 3D room geometry from arbitrary trajectories.

\begin{figure*}[ht]
\centering
\setlength\tabcolsep{0pt}
\begin{tabular}{cc}
\includegraphics[width=0.712\textwidth]{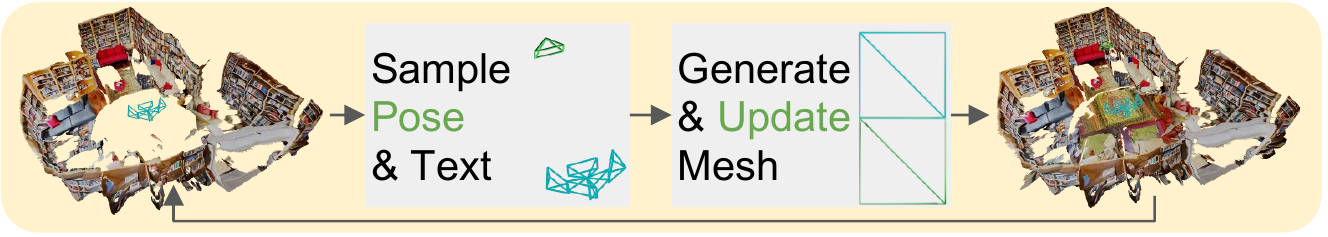} &
\includegraphics[width=0.287\textwidth]{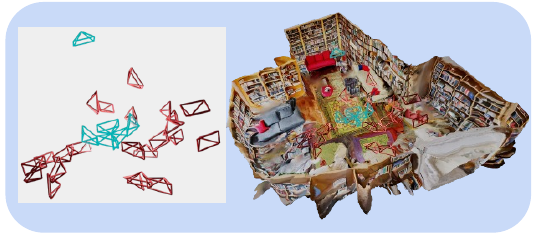} \\
(a) Scene Generation Stage & (b) Scene Completion Stage
\end{tabular}
\vspace{0.5mm}
\caption{\textbf{Method overview}. We iteratively create a textured 3D mesh in two stages. (a) First, we sample predefined poses and text to generate the complete scene layout and furniture. 
Each new pose (marked in green) adds newly generated geometry to the mesh (depicted by green triangles) in an iterative scene generation scheme (see Figure~\ref{fig:iterative-gen} for details).
Blue poses/triangles denote viewpoints that created geometry in a previous iteration.
(b) Second, we fill in the remaining unobserved regions by sampling additional poses (marked in red) after the scene layout is defined.}
\label{fig:pipeline}
\end{figure*}

\section{Method}

Our method creates a textured 3D mesh of a complete scene from text input.
To this end, we continuously fuse generated frames from a 2D text-to-image model at different poses into a joint 3D mesh, creating the scene over time.
The core idea of our approach is a two-stage tailored viewpoint selection, that first generates the scene layout and objects and then closes remaining holes in the 3D geometry (Section~\ref{subsec:Trajectory Generation}).
We visualize this workflow in Figure~\ref{fig:pipeline}.
For each pose in both stages, we apply an iterative scene generation scheme to update the mesh (Section~\ref{subsec:Perpetual Generation}).
We first align each frame with the existing geometry with a depth alignment strategy (Section~\ref{subsec:Depth Alignment}).
Next, we triangulate and filter the novel content to merge it into the mesh (Section~\ref{subsec:3D Mesh Generation}).

\subsection{Iterative 3D Scene Generation}
\label{subsec:Perpetual Generation}

\begin{figure}
\centering
\includegraphics[width=0.5\textwidth]{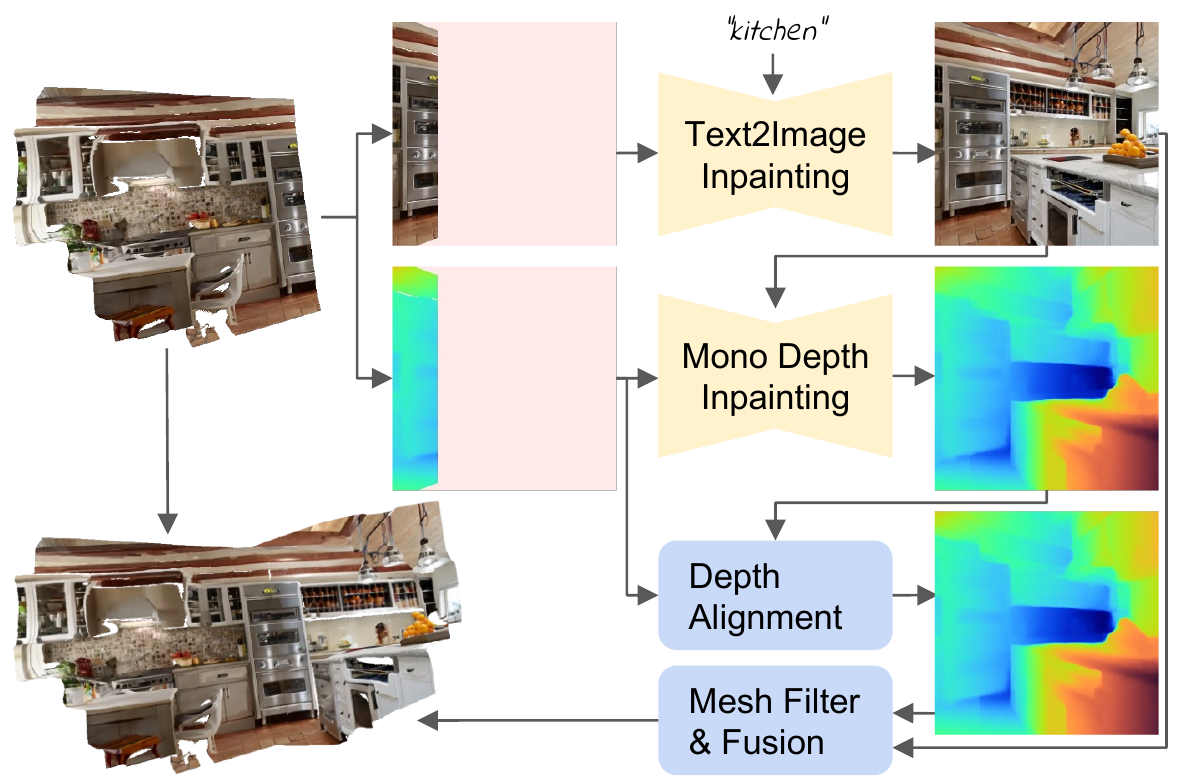}
\caption{\textbf{Iterative scene generation}. For each new pose, we render the current mesh to obtain partial RGB and depth renderings. We complete both, utilizing respective inpainting models and the text prompt. Next, we perform depth alignment (see Section~\ref{subsec:Depth Alignment}) and mesh filtering (see Section~\ref{subsec:3D Mesh Generation}) to obtain an optimal next mesh patch, that is finally fused with the existing geometry.}
\label{fig:iterative-gen}
\end{figure}

Our scene is represented as a mesh $\mathcal{M}=(\mathcal{V}, \mathcal{C}, \mathcal{S})$ where the vertices $\mathcal{V} \in \mathbb{R}^{N \times 3}$, vertex colors $\mathcal{C} \in \mathbb{R}^{N \times 3}$ and the face set $\mathcal{S} \in \mathbb{N}_0^{M\times 3}$ are generated over time.
Input to our method is a set of arbitrary text prompts $\{P_t\}_{t=1}^T$ that corresponds to our selected poses $\{E_t\}_{t=1}^T \in \mathbb{R}^{3\times 4}$ in both stages.
Inspired by recent methods~\cite{infinite_nature_2020, li2022_infinite_nature_zero}, we iteratively build up the scene, following a \textit{render-refine-repeat} pattern.
We summarize this iterative scene generation process in Figure~\ref{fig:iterative-gen}.
Formally, for each step of generation $t$, we first render the current scene from a novel viewpoint: 
\begin{equation}
\label{eq:step-1}
    I_t, d_t, m_t = \textit{r}(\mathcal{M}_t, E_t),
\end{equation}
where $\textit{r}$ is a classical rasterization function without shading, $I_t$ is the rendered image, $d_t$ the rendered depth and $m_t$ the image-space mask, that marks pixels without observed content.
We then use a fixed text-to-image model $\mathcal{F}_{t2i}$ to inpaint unobserved pixels according to the text prompt:
\begin{equation}
\label{eq:step-2}
    \hat{I}_t = \mathcal{F}_{t2i}(I_t, m_t, P_t).
\end{equation}
Next, we inpaint unobserved depth by applying a monocular depth estimator $\mathcal{F}_{d}$ in our depth alignment (see Section~\ref{subsec:Depth Alignment}):
\begin{equation}
\label{eq:step-3}
    \hat{d}_t = \textit{predict-and-align}(\mathcal{F}_{d}, I_t, d_t, m_t).
\end{equation}
Finally, we combine the novel content $\{\hat{I}_t, \hat{d}_t, m_t\}$ with the existing mesh by our fusion scheme (see Section~\ref{subsec:3D Mesh Generation}):
\begin{equation}
\label{eq:step-4}
    \mathcal{M}_{t{+}1} = \textit{fuse}(\mathcal{M}_{t}, \hat{I}_t, \hat{d}_t, m_t, E_t).
\end{equation}
\subsection{Depth Alignment Step}
\label{subsec:Depth Alignment}

To lift a 2D image $I$ into 3D, we predict the per-pixel depth.
To correctly combine old and new content, it is necessary that both align with each other.
In other words, similar regions in a scene like walls or furniture should be placed at similar depth.
However, directly using the predicted depth for backprojection leads to hard cuts and discontinuities in the 3D geometry, since the depth is inconsistent in scale between subsequent viewpoints (see Figure~\ref{fig:ours-ablation}a).

To this end, we perform depth alignment in two-stages.
First, we use a state-of-the-art depth inpainting network~\cite{Bae2022} that takes ground-truth depth $d$ for known parts in the image as input and aligns the prediction to it: $\hat{d}_{p} = \mathcal{F}_d(I, d).$

Inspired by Liu~\etal~\cite{infinite_nature_2020} we then improve the result by optimizing for scale and shift parameters $\gamma, \beta \in \mathbb{R}$, aligning predicted and rendered disparity in the least squares sense:
\begin{equation}
\label{eq:align-step-2}
    \min_{\gamma, \beta} \quad \left \lVert m \odot \left( \frac{\gamma}{\hat{d}_{p}} + \beta - \frac{1}{d} \right) \right \rVert^2,
\end{equation}
where we mask out unobserved pixels via $m$.
We can then extract the aligned depth as $\hat{d} = (\frac{\gamma}{\hat{d}_{p}} + \beta)^{-1}$.
Finally, we smooth $\hat{d}$ by applying a $5\times5$ Gaussian kernel at the mask edges (see supplemental material for more details).

\subsection{Mesh Fusion Step}
\label{subsec:3D Mesh Generation}

\begin{figure}
 \centering
 \setlength\tabcolsep{1pt}
 \begin{tabular}{ccc}
 \includegraphics[width=0.1448\textwidth]{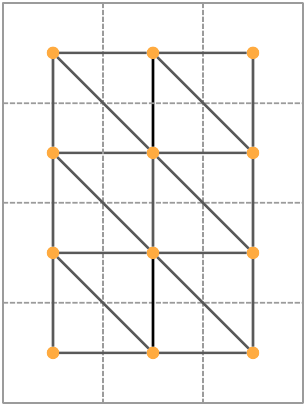} &
 \includegraphics[width=0.1448\textwidth]{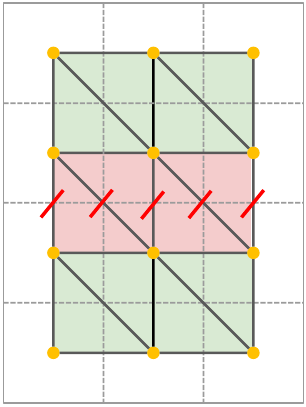} &
 \includegraphics[width=0.1096\textwidth]{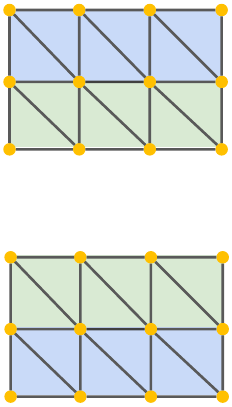} \\
 (a) Pixel Triangulation & (b) Face Filtering & (c) Mesh Fusion
 \end{tabular}
 \vspace{1mm}
 \caption{\textbf{Visualization of our mesh fusion step.}
(a) We triangulate an image, such that 4 neighboring pixels (orange dots) create two faces.
(b) We filter a face (marked in red), if its surface normal forms a small grazing angle with the viewing direction or if any edge in world space is too long.
(c) We fuse the remaining faces (marked in green) with the existing geometry (marked in blue).
 }
 \label{fig:fusion}
\end{figure}

At each step, we insert new content $\{\hat{I}_t, \hat{d}_t, m_t\}$ into the scene.
For that, we first backproject the image-space pixels into a world-space point cloud:
\begin{equation}
\label{eq:unproj}
\mathcal{P}_t = \{E_t^{-1} K^{-1} \cdot \hat{d}_t[u,v] \cdot (u, v, 1)^T\}_{u=0,v=0}^{W, H},
\end{equation}
where $K \in \mathbb{R}^{3\times 3}$ are the camera intrinsics and $W, H$ are image width and height, respectively.
We then use a simple triangulation scheme (Figure~\ref{fig:fusion}a), where each four neighboring pixels $\{(u, v), (u{+}1, v), (u, v{+}1), (u{+}1, v{+}1)\}$ in the image form two triangles.
Since the estimated depth is noisy, this na\"ive triangulation creates stretched out 3D geometry (see Figure~\ref{fig:ours-ablation}b).
To alleviate this problem, we propose two filters that remove stretched out faces (Figure~\ref{fig:fusion}b).

First, we filter faces based on their edge length.
We remove a face if the Euclidean distance of any face edge is larger than a threshold $\delta_{edge}$.
Second, we filter faces based on the angle between surface normal and viewing direction:
\begin{equation}
\label{eq:sn-filter}
\mathcal{S} = \{(i_0, i_1, i_2) | n^T v > \delta_{sn} \}
\end{equation}
where $\mathcal{S}$ is the face set, $(i_0, i_1, i_2)$ are the vertex indices of the triangle, $\delta_{sn}$ is the threshold, $n \in \mathbb{R}^3$ is the normalized face normal, and $v \in \mathbb{R}^3$ is the normalized view direction in world space from the camera center towards the average pixel location from which the triangle originated. 
This avoids creating texture for large regions of the mesh from a comparatively small number of pixels from an image.

Finally, we fuse together the newly generated mesh patch and the existing geometry (Figure~\ref{fig:fusion}c).
All faces that are backprojected from pixels falling into the inpainting mask $m_t$ are stitched together with their neighboring faces, which are already part of the mesh.
Precisely, we continue the triangulation scheme at all edges of $m_t$, but use the existing vertex positions of $\mathcal{M}_t$ to create the corresponding faces.

\subsection{Two-Stage Viewpoint Selection}
\label{subsec:Trajectory Generation}

A key part of our method is the choice of text prompts and camera poses from which the scene is synthesized.
Users can in principle choose these inputs arbitrarily to create any desired indoor scene.
However, the generated scene can degenerate and contain stretch and hole artifacts, if poses are chosen carelessly (see Figure~\ref{fig:ours-ablation} and supplemental material).
To this end, we propose a two-stage viewpoint selection strategy, that samples each next camera pose from optimal positions and refines empty regions subsequently.

\mypar{Generation Stage}
\label{para:gen-stage}
In the first stage, we create the main parts of the scene, including the general layout and furniture.
We subsequently render \emph{predefined} trajectories in different directions that eventually cover the whole room.
We found generation works best, if each trajectory starts off from a viewpoint with mostly unobserved regions.
This generates the outline of the next chunk, while still being connected to the rest of the scene (e.g., see Figure~\ref{fig:iterative-gen}).
Then, we complete the 3D structure of that chunk by moving and rotating into it subsequently until the end of the trajectory.

Additionally, we ensure an optimal observation distance for each pose.
We translates camera positions $T_0{\in}\mathbb{R}^3$ along the look-at direction $L{\in}\mathbb{R}^3$ uniformly: $T_{i+1}{=}T_{i}{-}0.3 L$.
We stop if the mean rendered depth is larger than $0.1$ or discard the camera after $10$ steps.
This avoids views too close to existing geometry.
For example, the green pose in Figure~\ref{fig:pipeline}a is moved back as far as possible into the existing geometry such that it views most of the empty floor region.

We create closed room layouts following this principle, by choosing trajectories that generate the next chunks in a circular motion, roughly centered around the origin.
We found it helpful to discourage the text-to-image generator from generating furniture in unwanted regions by engineering the text prompts accordingly.
For example, for poses looking at the floor or ceiling, we choose text prompts that only contain the words ``floor'' or ``ceiling'', respectively.

\mypar{Completion Stage}
\label{para:complete-stage}
After the first stage, the scene layout and furniture is defined.
However, it is impossible to choose sufficient poses \emph{a-priori}.
Since the scene is generated on-the-fly, the mesh contains holes that were not observed by any camera (see Figure~\ref{fig:ours-ablation}c).
We complete the scene by sampling additional poses \emph{a-posteriori}, looking at those holes.

Inspired by trajectory optimization~\cite{hepp2018plan3d, roberts2017submodular}, we voxelize the scene into dense uniform cells.
We sample random poses in each cell, discarding those being too close to existing geometry.
We select one pose per cell that views most unobserved pixels (e.g., see the red poses in Figure~\ref{fig:pipeline}b).

Next, we inpaint the scene from all chosen camera poses following Section~\ref{subsec:Perpetual Generation}.
Similar to Fridman~\etal~\cite{fridman2023scenescape}, we observe it is important to clean the inpainting masks, because our text-to-image generator can generate better results for large connected regions.
Thus, we first inpaint small holes with a classical inpainting algorithm~\cite{telea2004image} and dilate the remaining holes.
We additionally remove all faces that fall into the dilated region and are close to the rendered depth.
Please see the supplemental material for more details.

Finally, we run Poisson surface reconstruction~\cite{kazhdan2006poisson} on the scene mesh.
This closes any remaining holes after completion and smoothens out discontinuities. 
The result is a watertight mesh of the generated scene, that can be rendered with classical rasterization.

\section{Results}
\mypar{Implementation Details}
We implement mesh rasterization and fusion with Pytorch3D~\cite{ravi2020pytorch3d}.
As our text-to-image model $\mathcal{F}_{t2i}$, we utilize a Stable Diffusion~\cite{Rombach2021HighResolutionIS} model, that is finetuned on the image inpainting task, using additional mask input.
We generate a single inpainting proposal and employ a state-of-the-art guided diffusion sampler~\cite{lu2022dpm}.
As our monocular depth estimator $\mathcal{F}_{d}$, we employ an IronDepth~\cite{Bae2022} model, that is trained on indoor scenes from the ScanNet dataset~\cite{dai2017scannet} and augment it for depth inpainting according to Bae~\etal~\cite{Bae2022}.
We set $\delta_{edge}{=}0.1$ and $\delta_{sn}{=}0.1$ in all our experiments.
During generation, we use $20$ different trajectories with $10$ frames each sampled between the respective start and end poses.
We construct prompts using the guidelines suggested by Pierre~\cite{twitter2023}.
Creating one scene takes approximately 50 minutes on one RTX 3090 GPU.

\mypar{Baselines}
To the best of our knowledge, there are no direct baselines that generate textured 3D room geometry from text.
We compare against four related methods (please see the supplemental material for more details about baselines).
\begin{itemize}[leftmargin=*,topsep=0pt, noitemsep]
    \item \emph{PureClipNeRF}~\cite{lee2022understanding}: We compare against text-to-3D methods for generating objects~\cite{poole2022dreamfusion, lin2022magic3d, jain2021dreamfields, lee2022understanding, wang2022score} and choose Lee~\etal~\cite{lee2022understanding} as open-source representative.
    \item \emph{Outpainting}~\cite{Ramesh2022HierarchicalTI, OpenAI2022}: We combine outpainting from a Stable Diffusion~\cite{Rombach2021HighResolutionIS} model with depth estimation and triangulation to create a mesh from an enlarged viewpoint.
    \item \emph{Text2Light}~\cite{chen2022text2light}: We generate RGB panoramas from text using Chen~\etal~\cite{chen2022text2light}. Estimating 3D mesh structure from a panorama is difficult. Related approaches estimate room layout~\cite{xu2021layout}, perform view synthesis ~\cite{kulkarni2022360fusionnerf, hsu2021moving, hara2022enhancement, huang2022360roam} or predict $360^\circ$ depth~\cite{area2022360monodepth, jin2020geometric}. We perform depth prediction and subsequently apply our mesh fusion step.
    \item \emph{Blockade}~\cite{blockade}: We apply \emph{Blockade}~\cite{blockade}, which uses a text-to-image diffusion model to produce more expressive RGB panoramas. We then extract the mesh similarly.
\end{itemize}

\mypar{Evaluation Metrics}
The generated 3D geometry is evaluated both quantitatively and qualitatively. 
We calculate CLIP Score (CS)~\cite{radford2021learning} and Inception Score (IS)~\cite{salimans2016improved} on RGB renderings of the respective scenes.
Additionally, we conduct a user study and ask $n{=}61$ users to score Perceptual Quality (\emph{PQ}) and 3D Structure Completeness (\emph{3DS}) of the whole scene on a scale of $1{-}5$.

\subsection{Qualitative Results}

\begin{figure*}
\centering
\setlength\tabcolsep{1pt}
\begin{tabular}{ccc}
\multirow{2}{*}[0.8in]{\includegraphics[height=48mm]{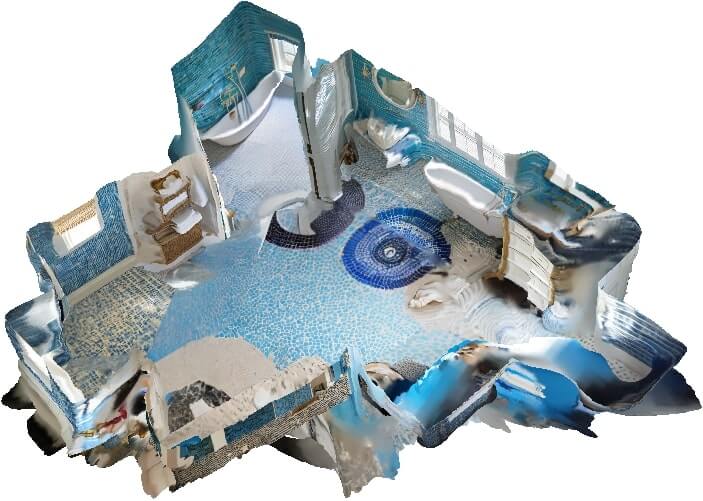}} & 
\includegraphics[height=24mm]{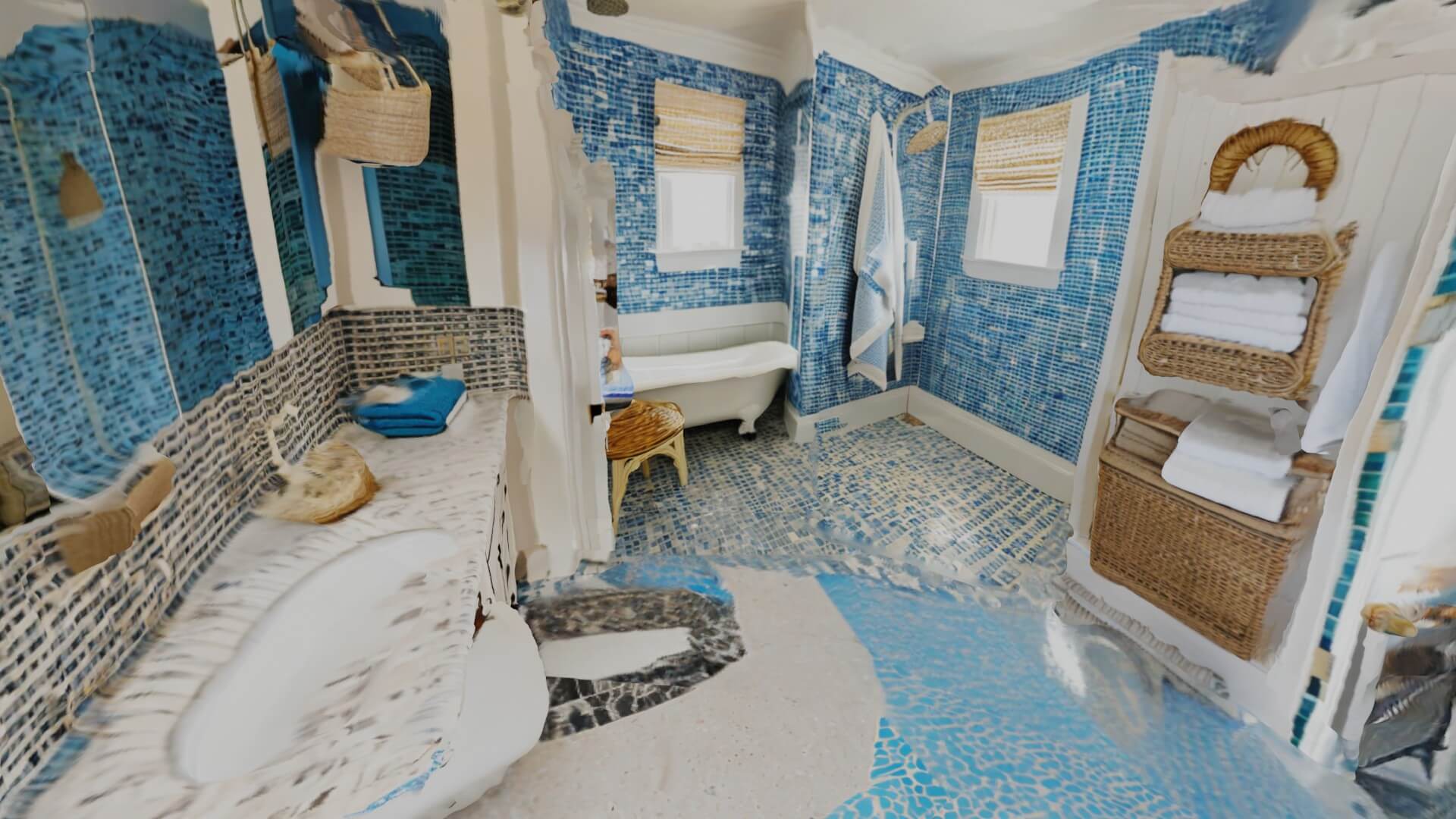} & 
\includegraphics[height=24mm]{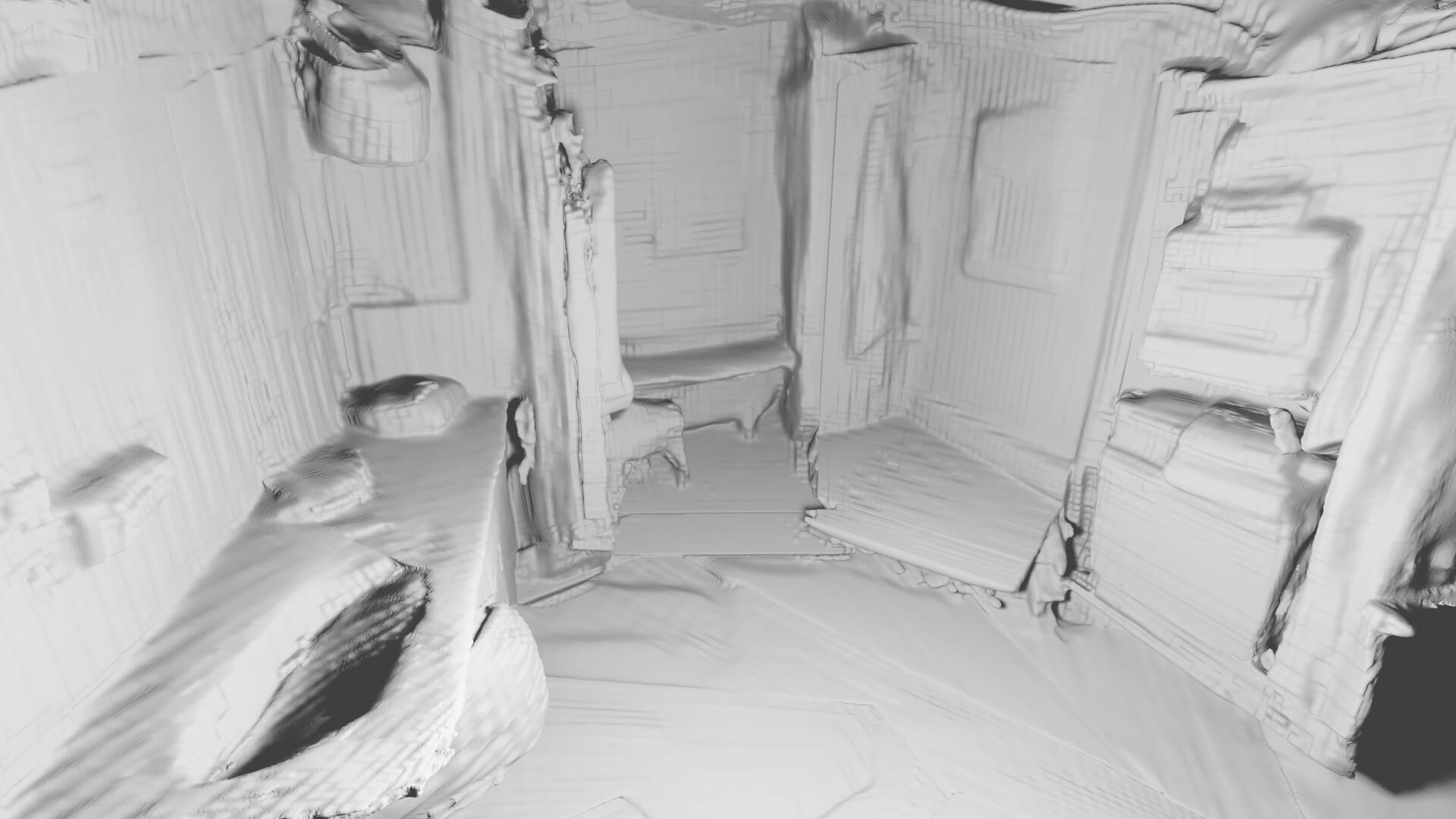} \\ 
& \includegraphics[height=24mm]{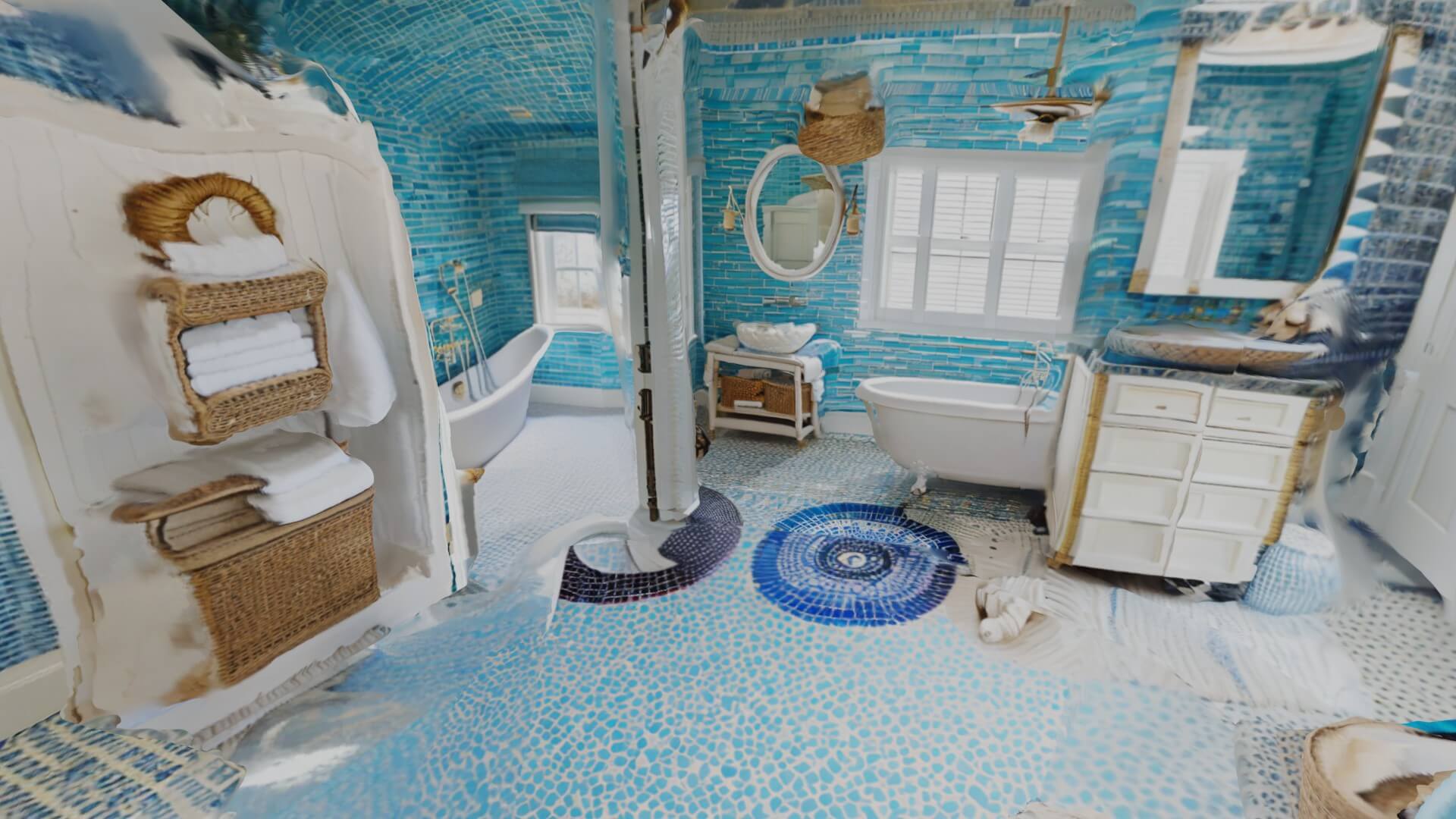} & 
\includegraphics[height=24mm]{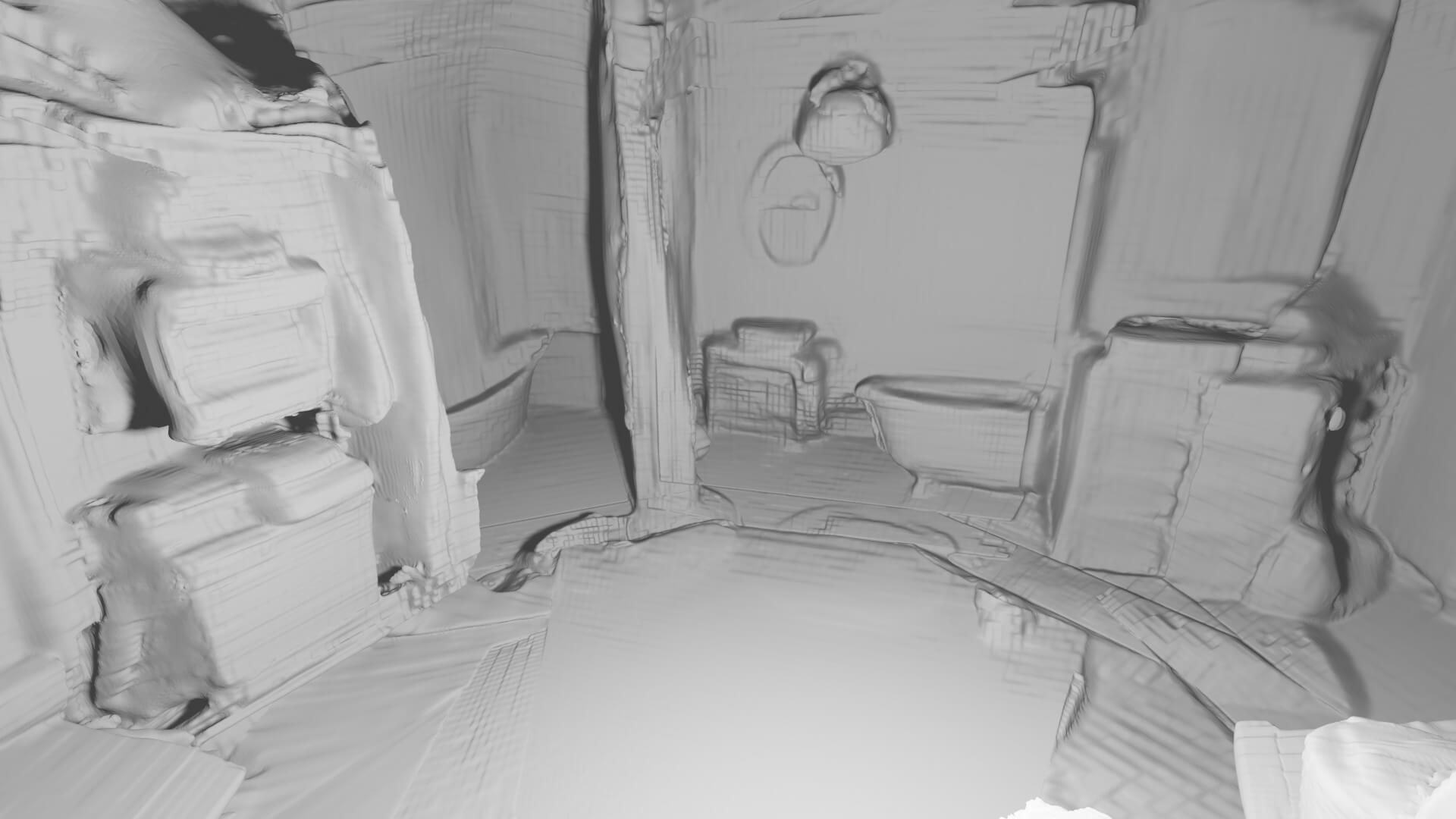} \\
\multicolumn{3}{c}{\textit{Editorial Style Photo, Coastal Bathroom, Clawfoot Tub, Seashell, Wicker, Mosaic Tile, Blue and White}} \\

\multirow{2}{*}[0.8in]{\includegraphics[height=48mm]{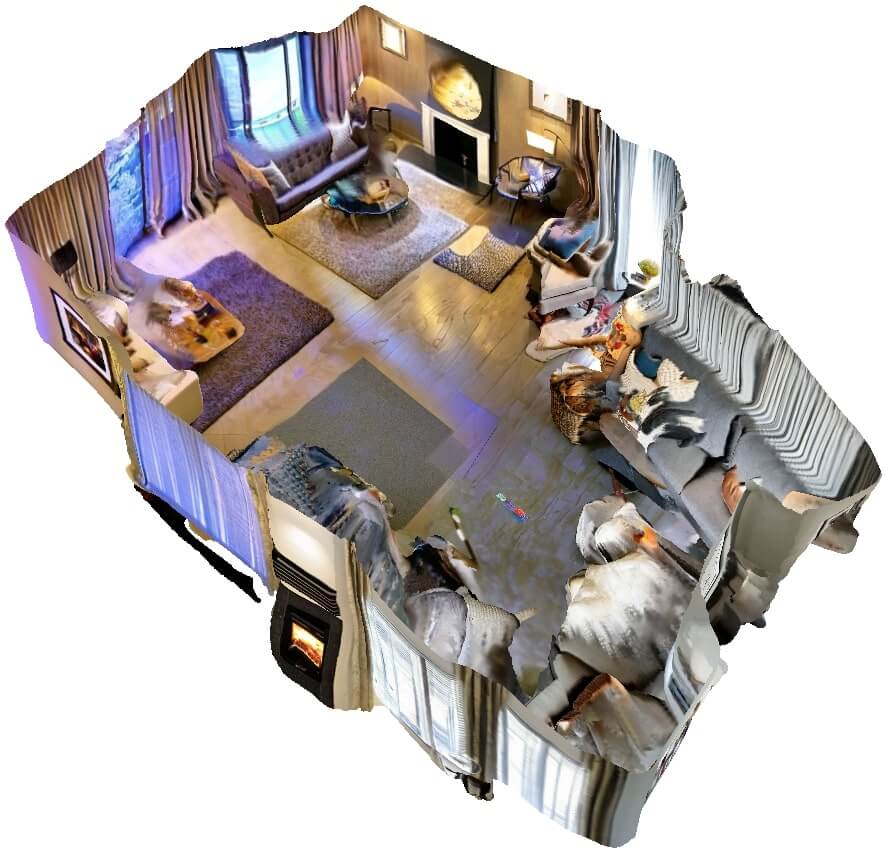}} & 
\includegraphics[height=24mm]{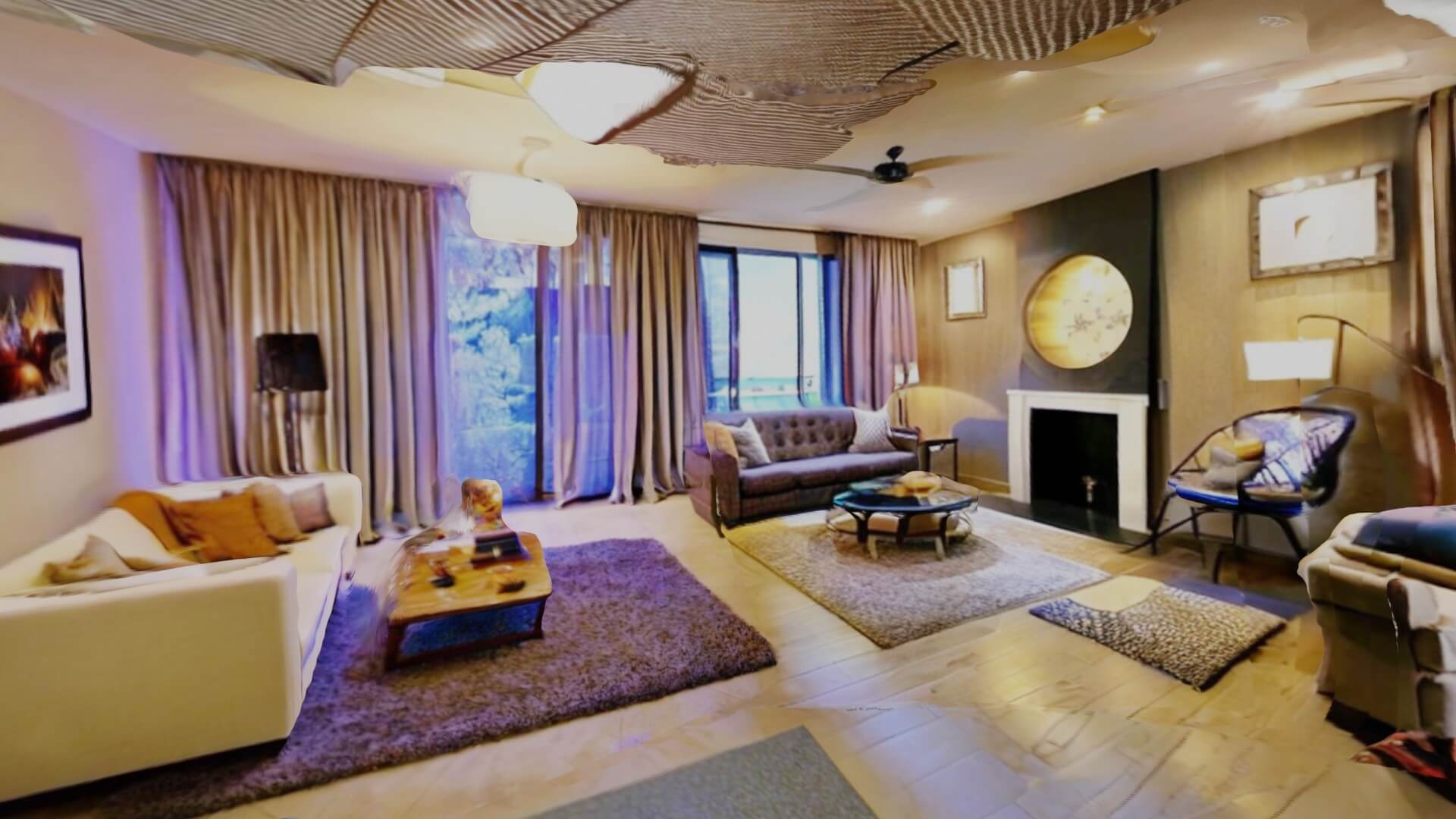} & 
\includegraphics[height=24mm]{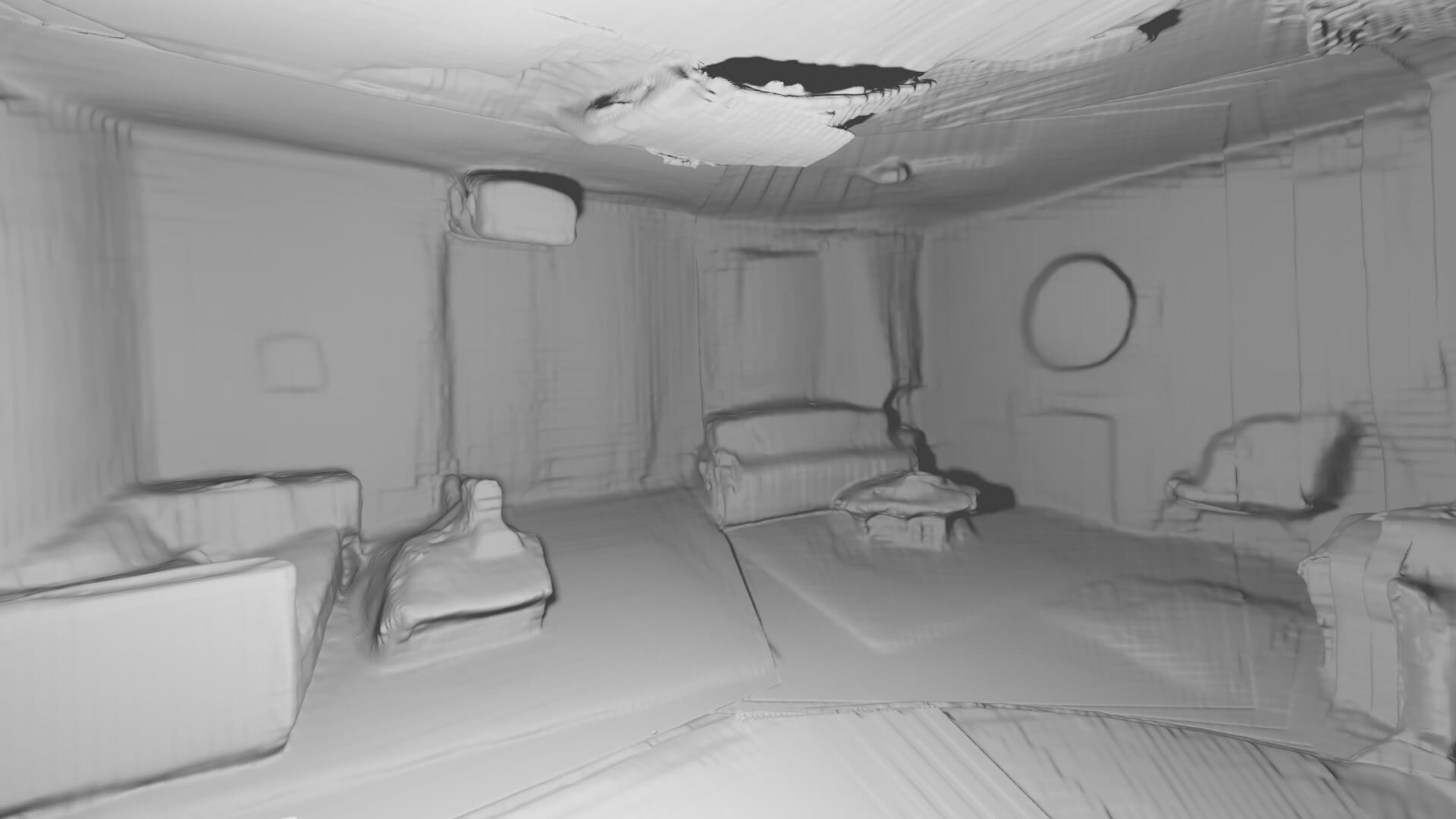} \\ 
& \includegraphics[height=24mm]{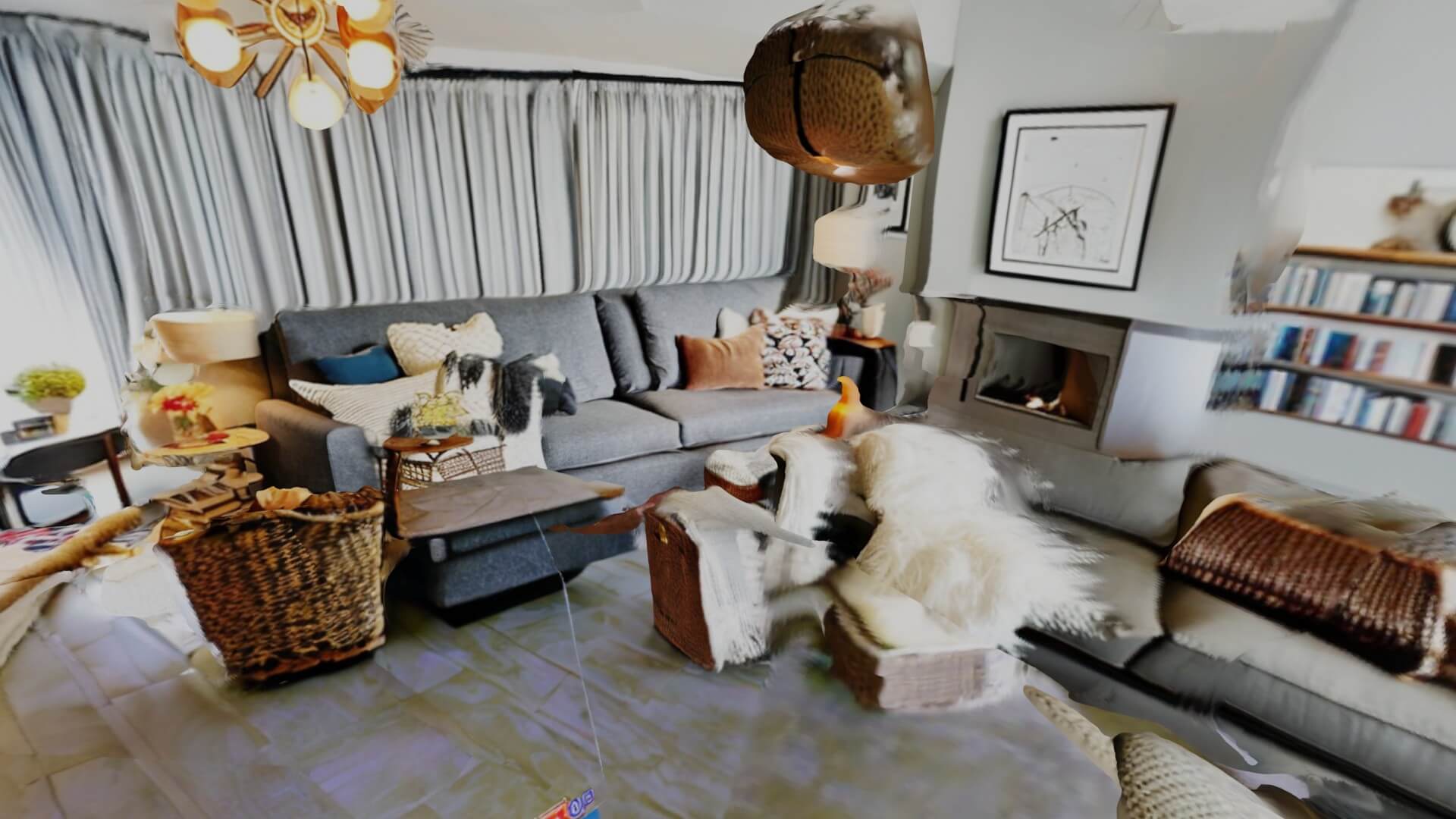} & 
\includegraphics[height=24mm]{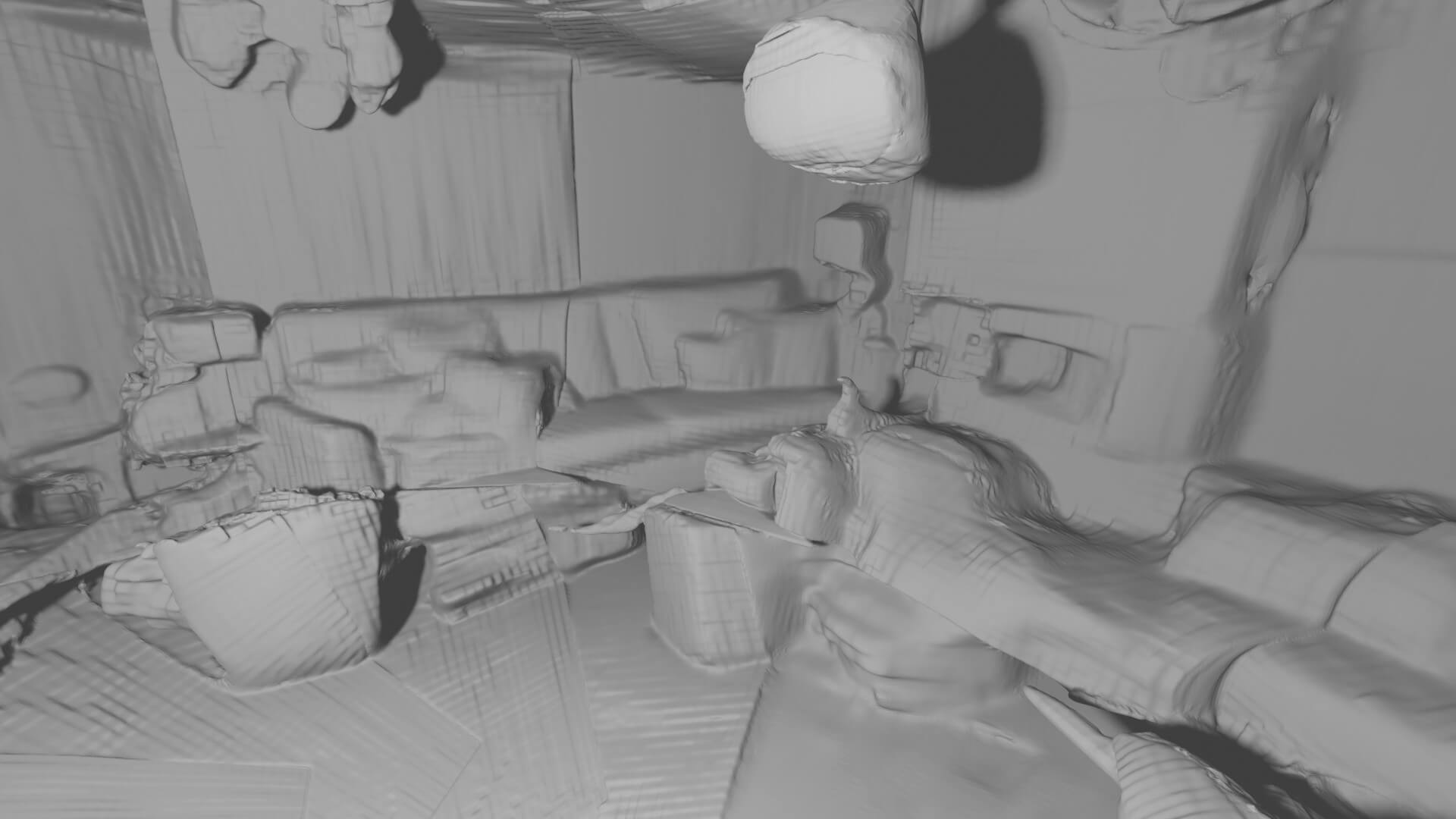} \\
\multicolumn{3}{c}{\textit{A living room with a lit furnace, couch, and cozy curtains, bright lamps that make the room look well-lit}} \\

\multirow{2}{*}[0.8in]{\includegraphics[height=48mm]{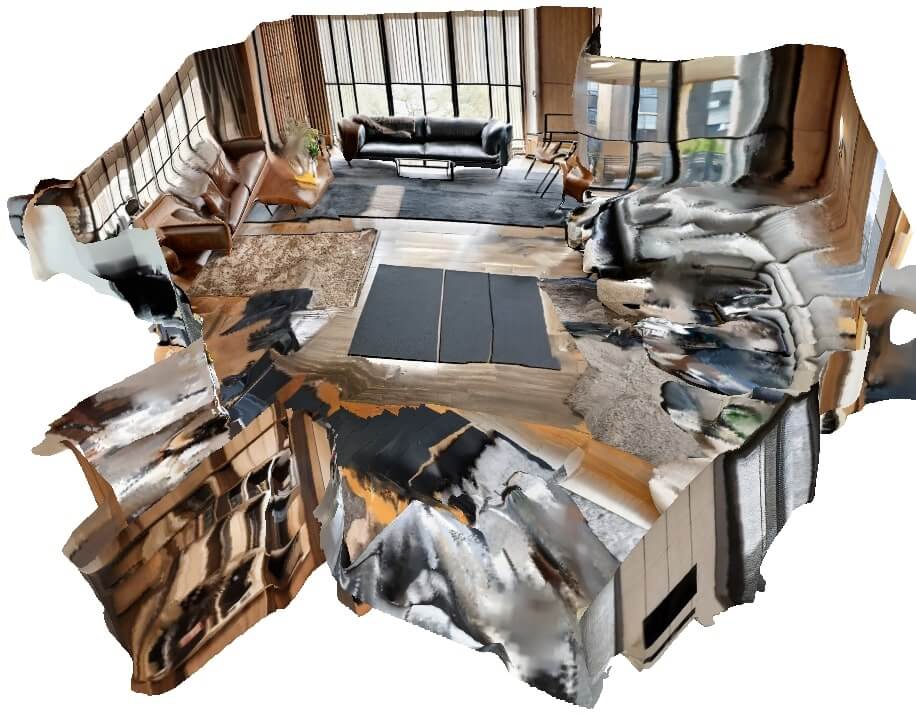}} & 
\includegraphics[height=24mm]{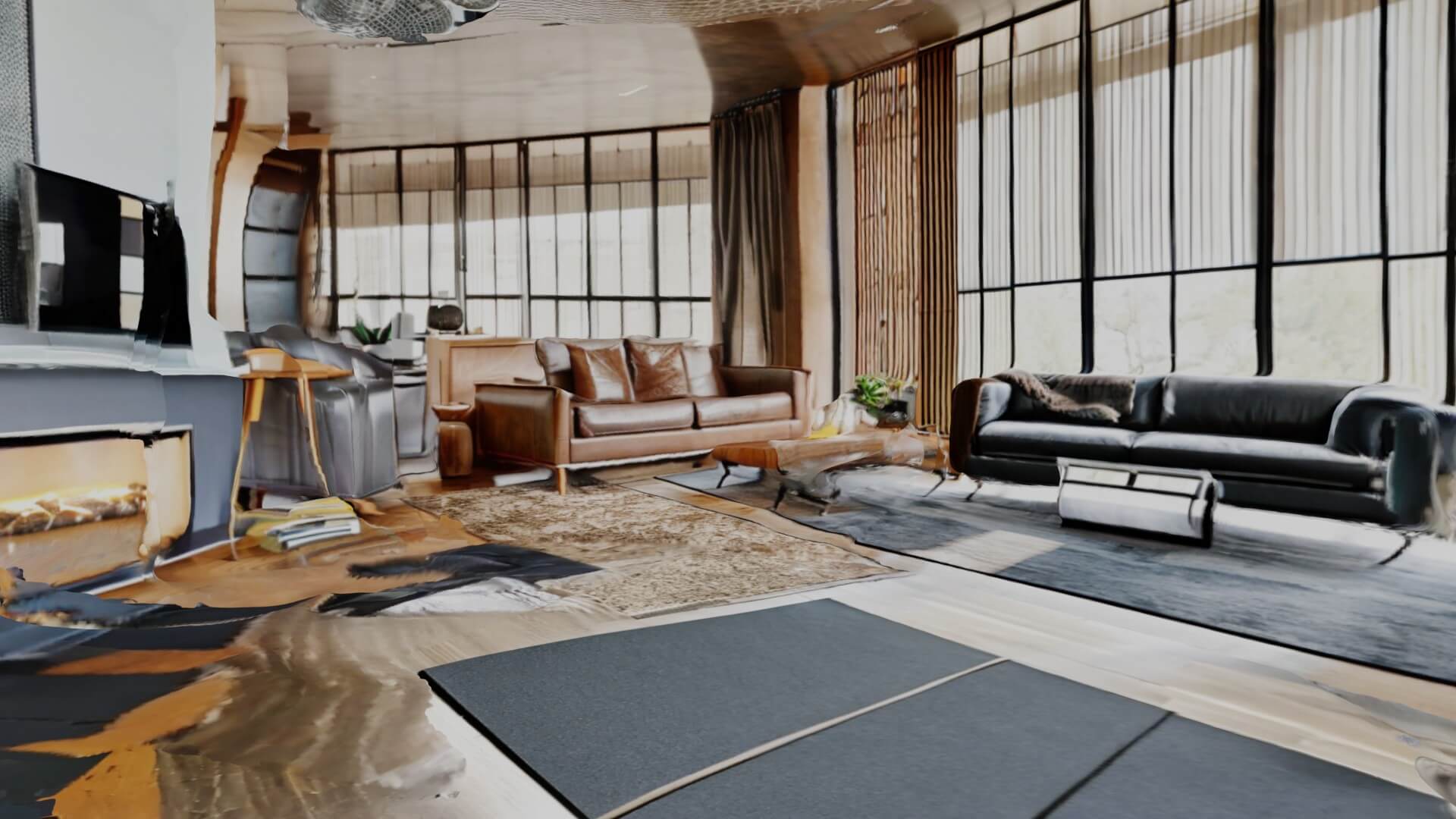} & 
\includegraphics[height=24mm]{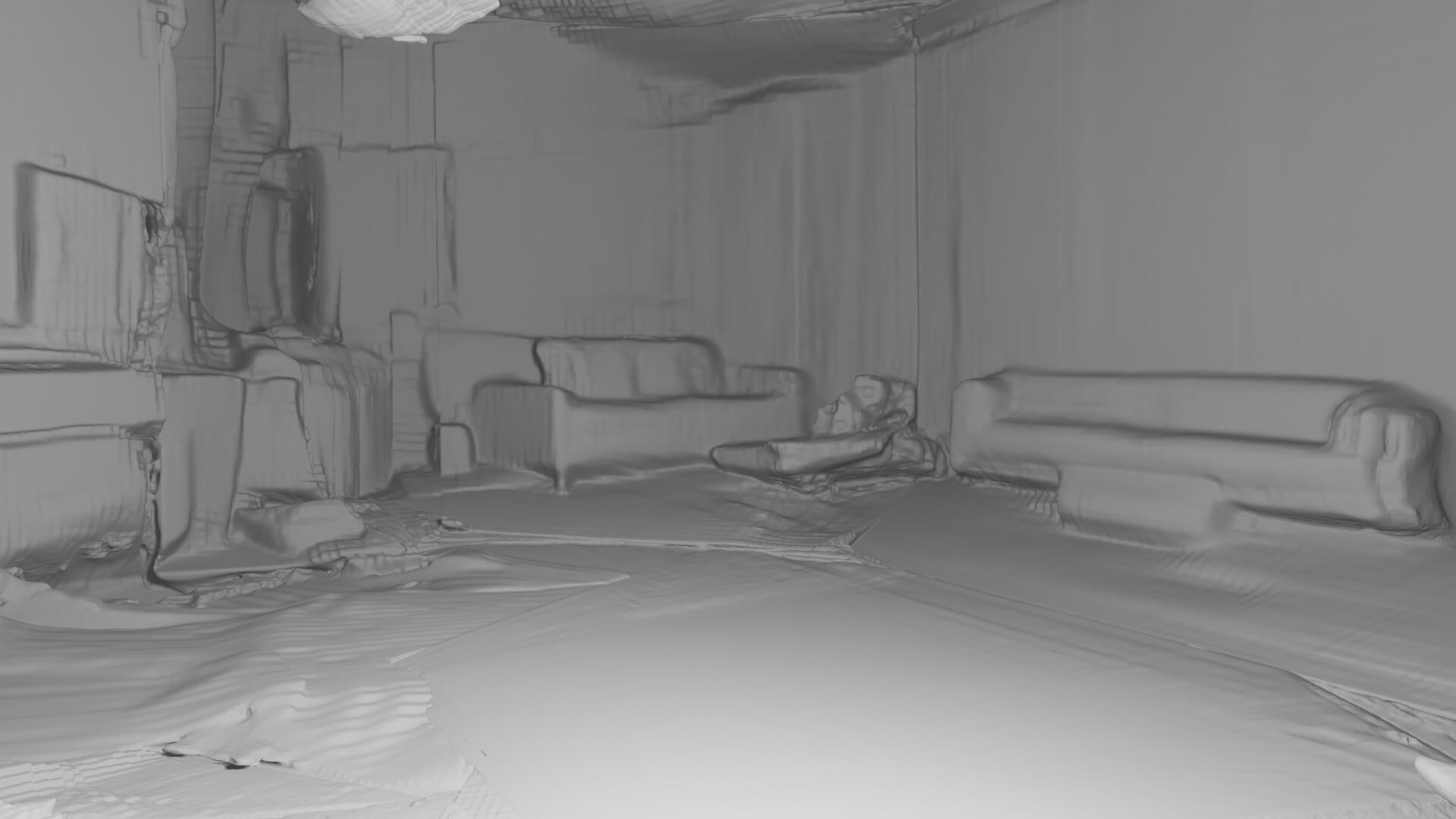} \\ 
& \includegraphics[height=24mm]{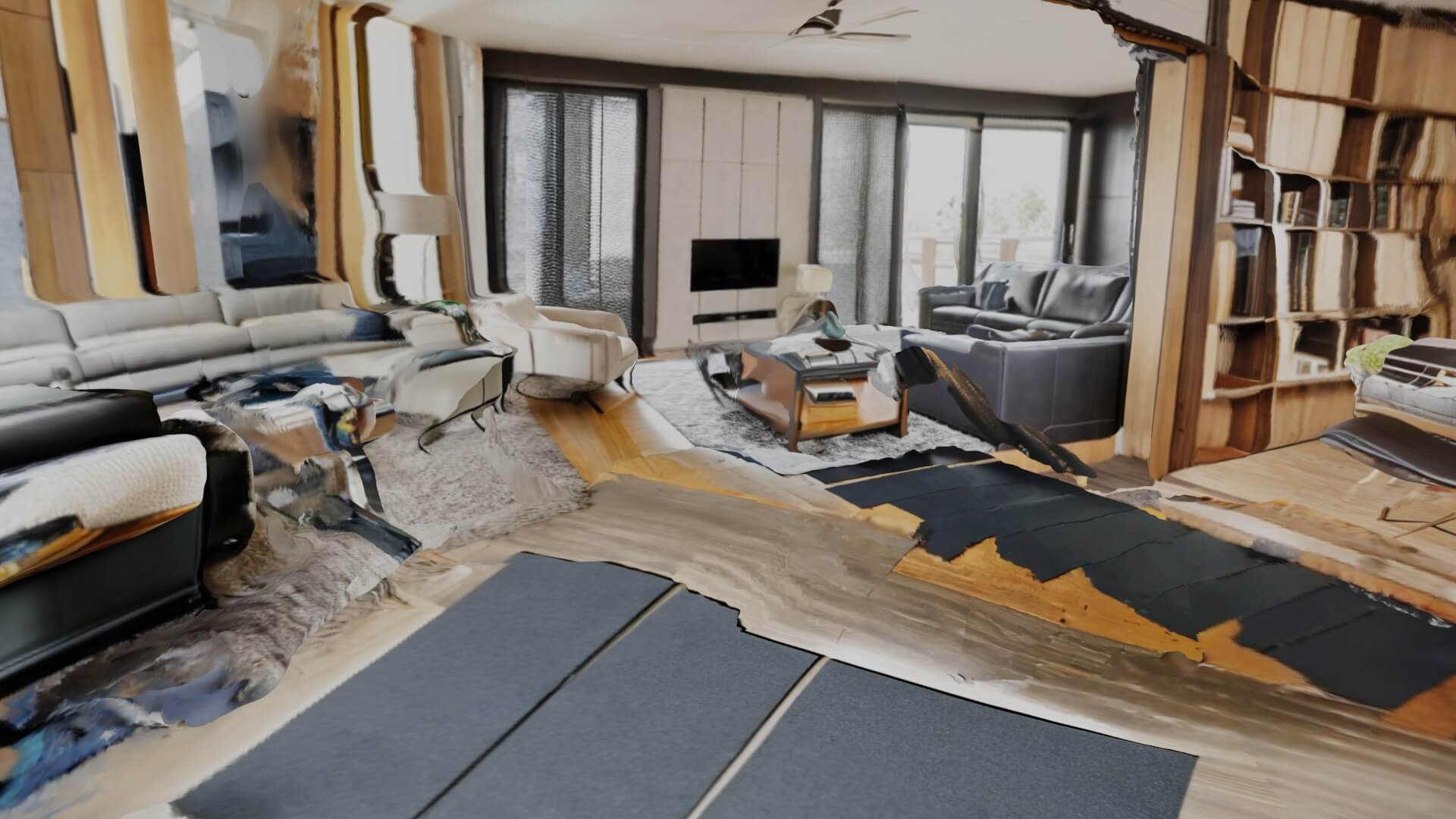} & 
\includegraphics[height=24mm]{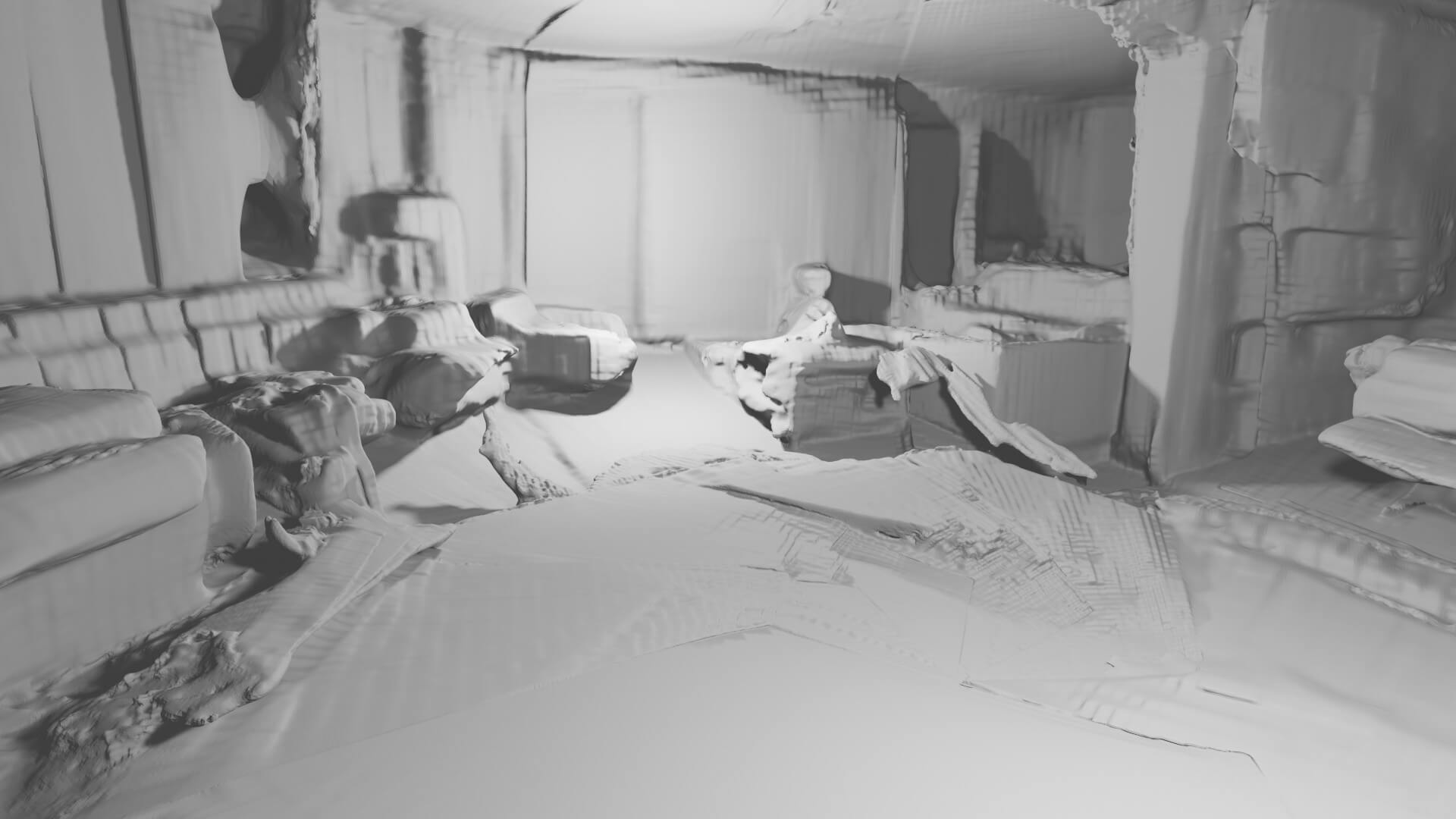} \\
\multicolumn{3}{c}{\textit{Editorial Style Photo, Modern Living Room, Large Window, Leather, Glass, Metal, Wood Paneling, Apartment}} \\

\multirow{2}{*}[0.8in]{\includegraphics[height=48mm]{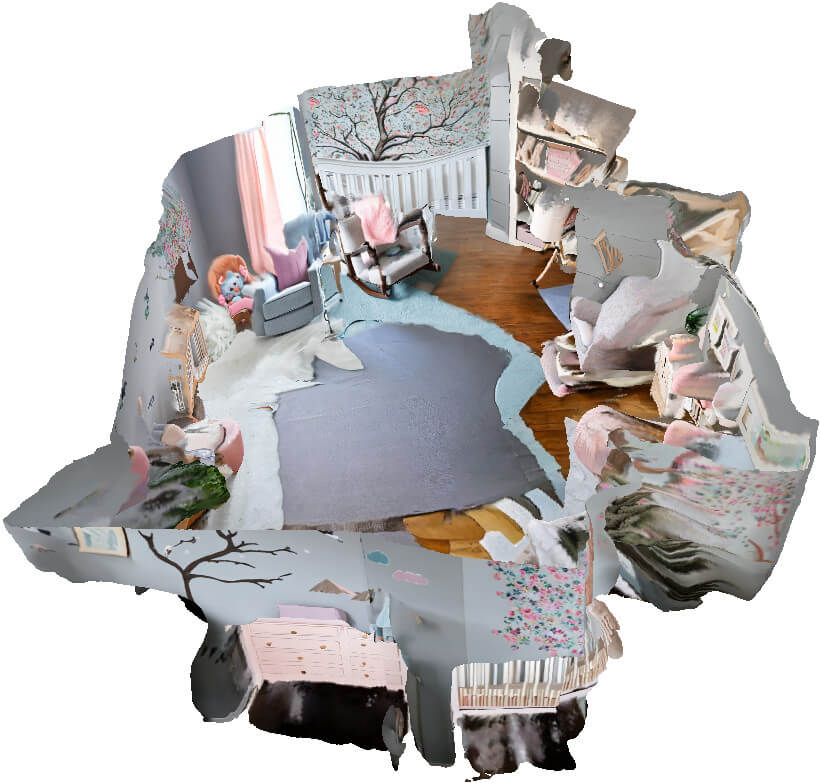}} & 
\includegraphics[height=24mm]{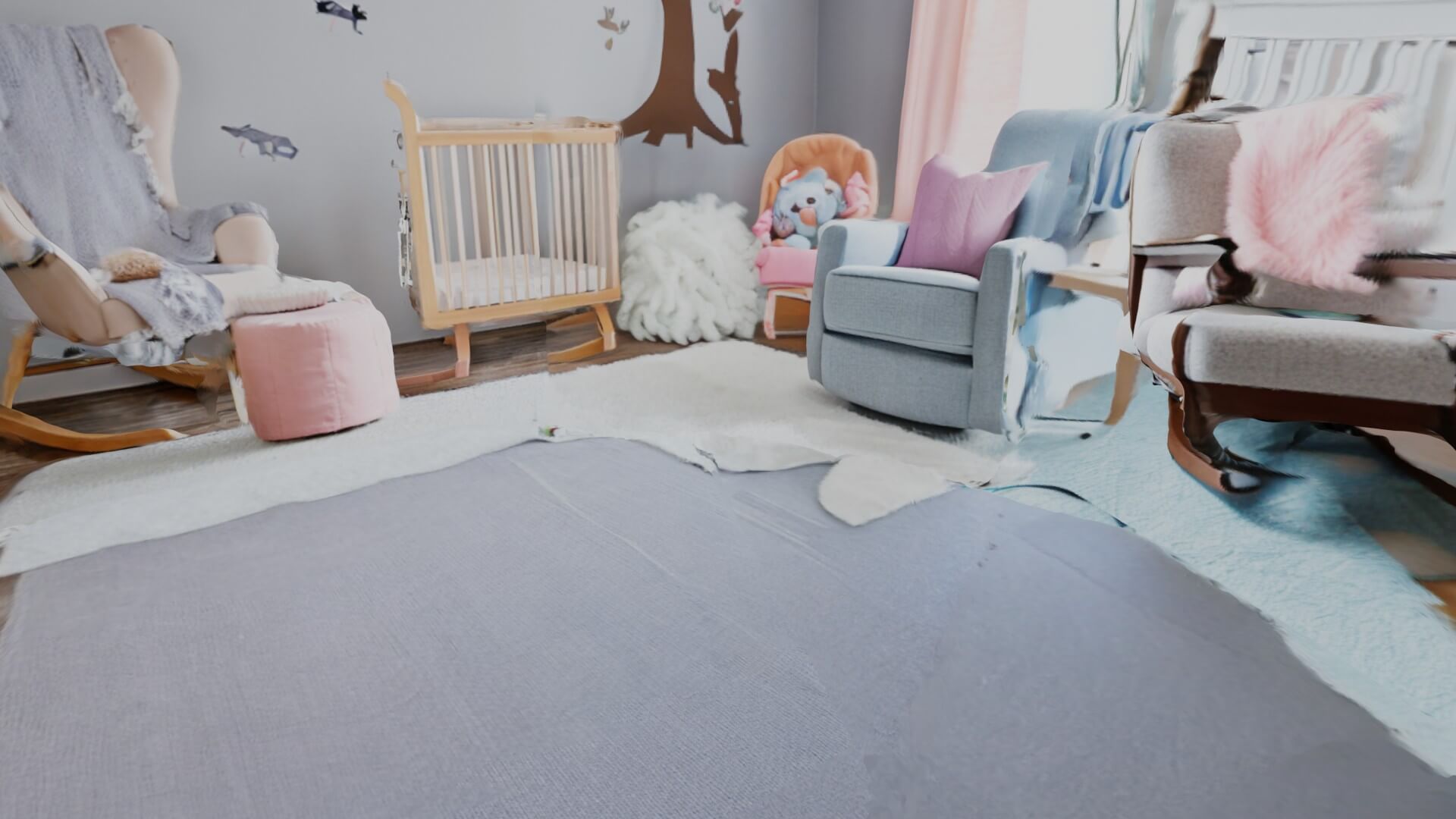} & 
\includegraphics[height=24mm]{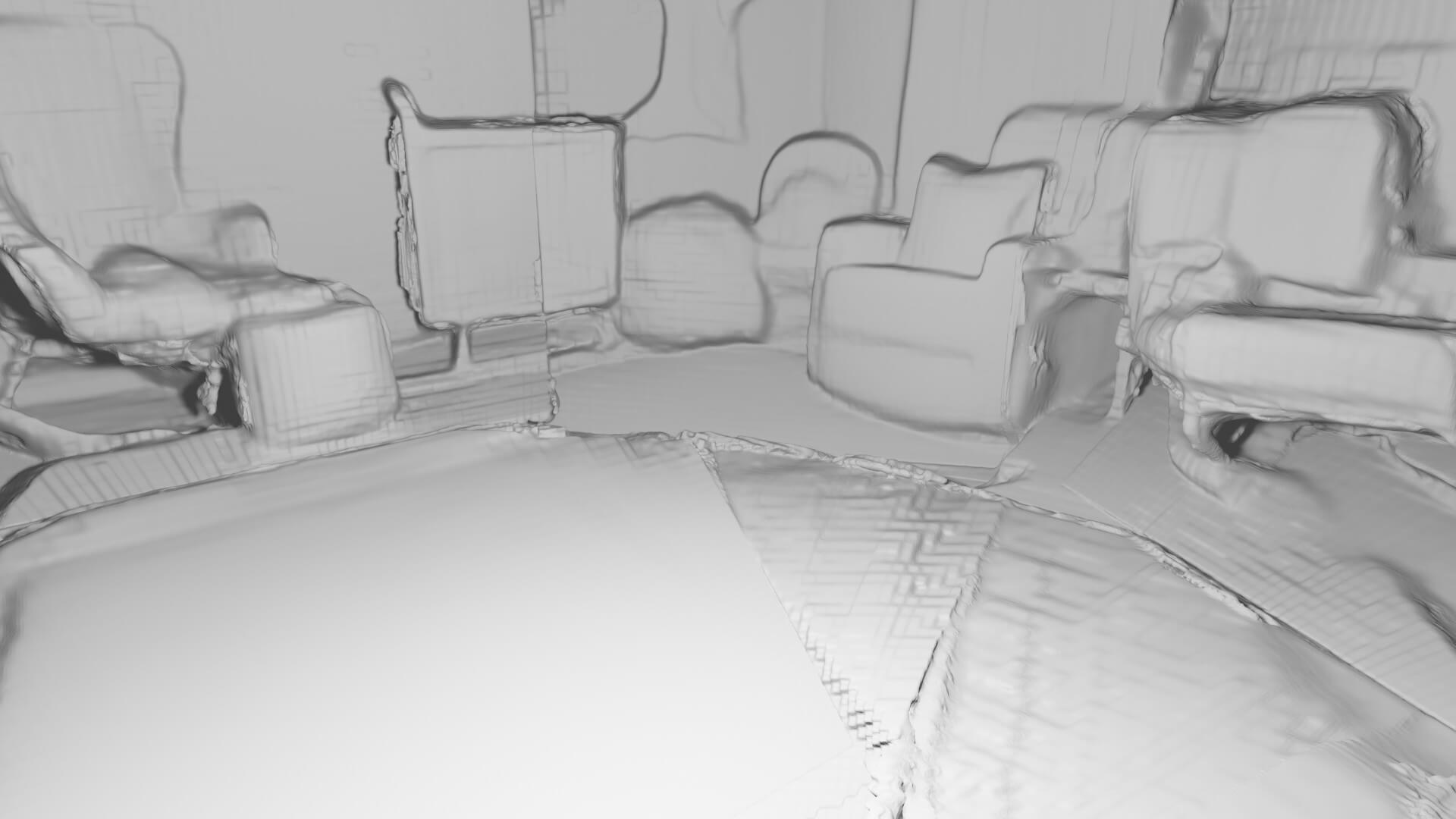} \\ 
& \includegraphics[height=24mm]{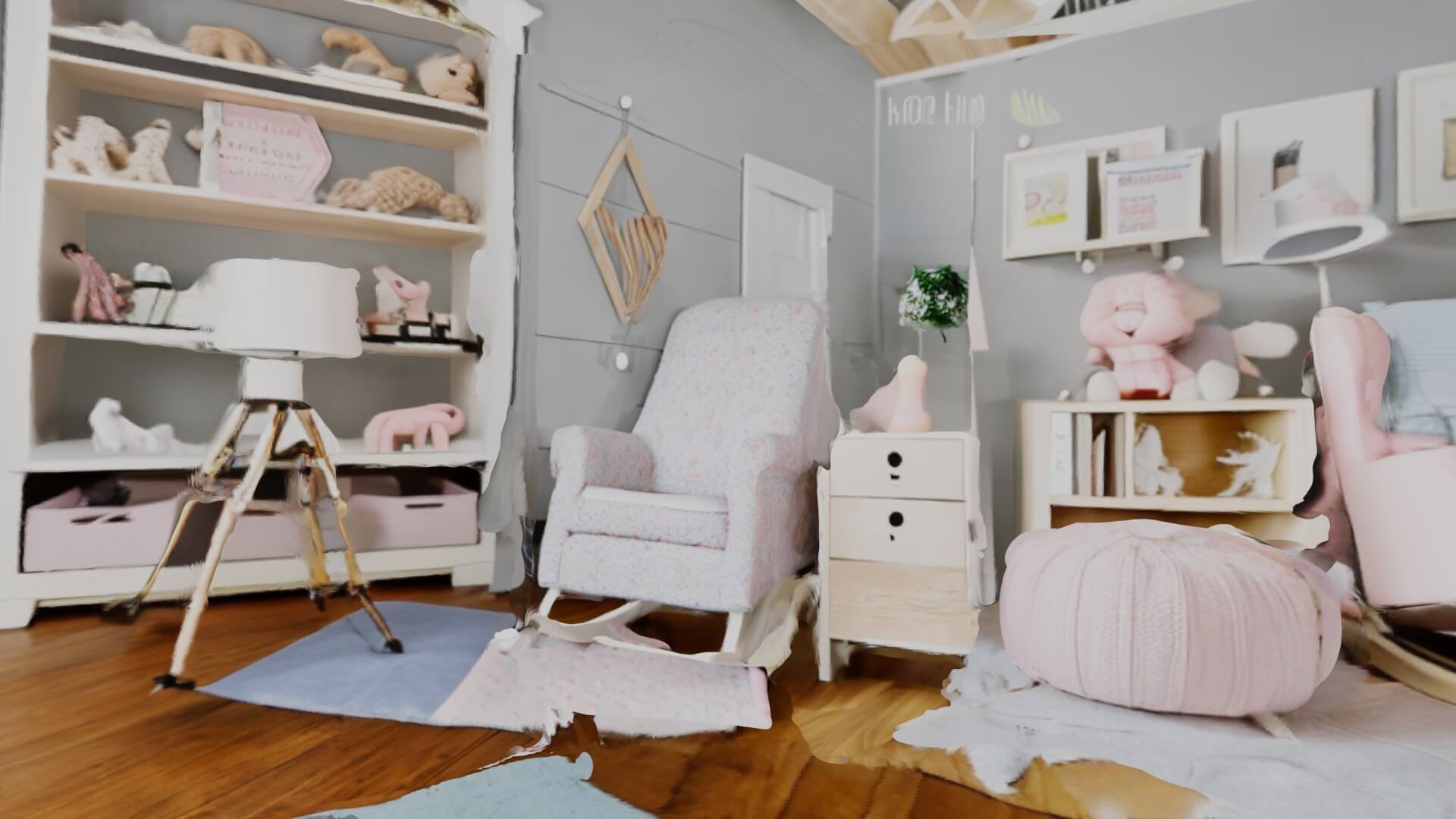} & 
\includegraphics[height=24mm]{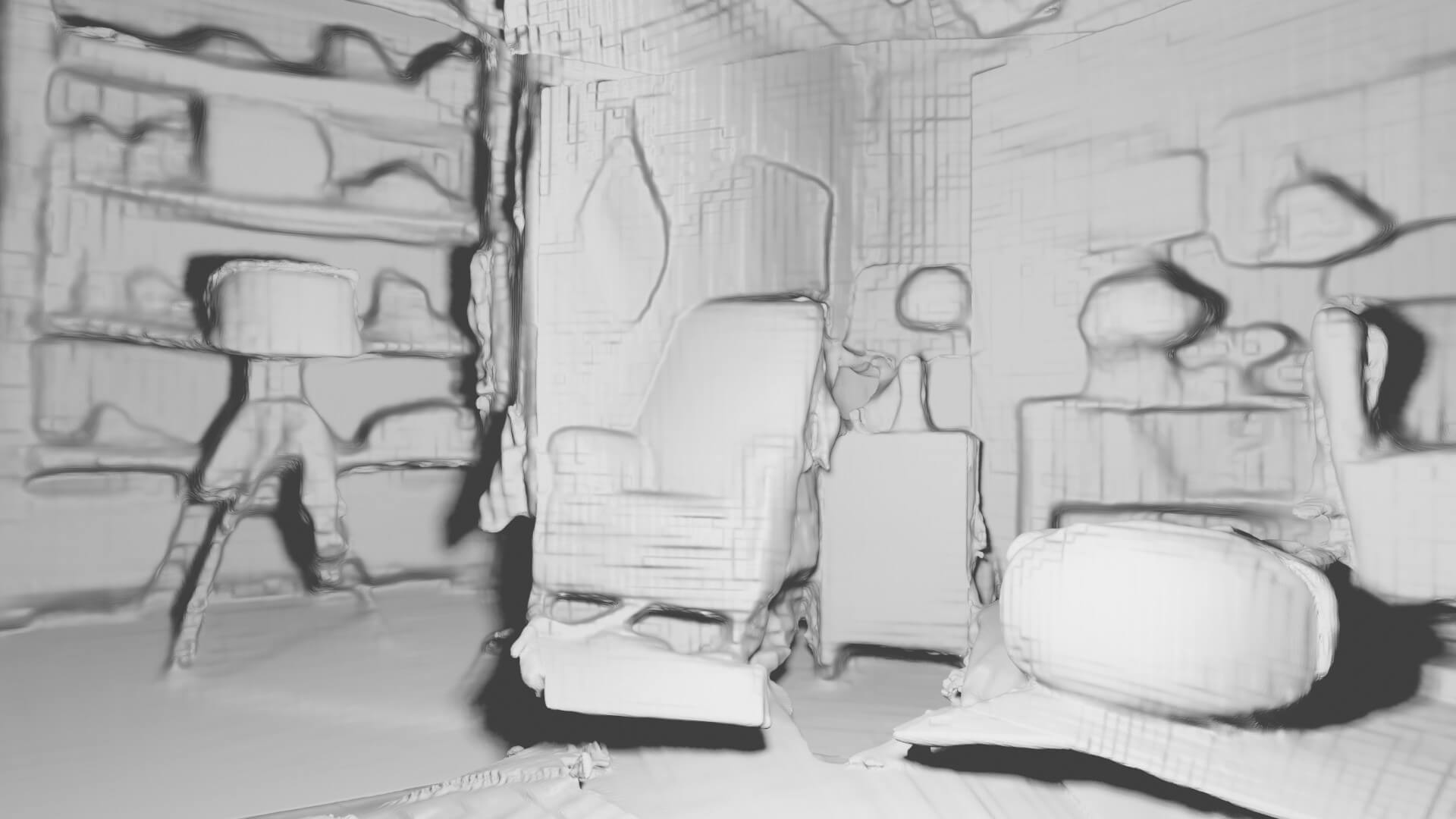} \\
\multicolumn{3}{c}{\textit{Editorial Style Photo, Modern Nursery, Table Lamp, Rocking Chair, Tree Wall Decal, Wood, Cotton, Faux Fur}} \\

\end{tabular}
\caption{
\textbf{3D scene generation results of our method.}
We show color and shaded geometry renderings from generated scenes with corresponding text prompts.
Our method synthesizes realistic meshes satisfying text descriptions. 
We remove the ceiling in the top-down view for better visualization of the scene layout.}
\label{fig:ours-only}
\end{figure*}

\begin{figure*}
\centering
\setlength\tabcolsep{1pt}
\begin{tabular}{ccccc}
\includegraphics[width=0.194\textwidth]{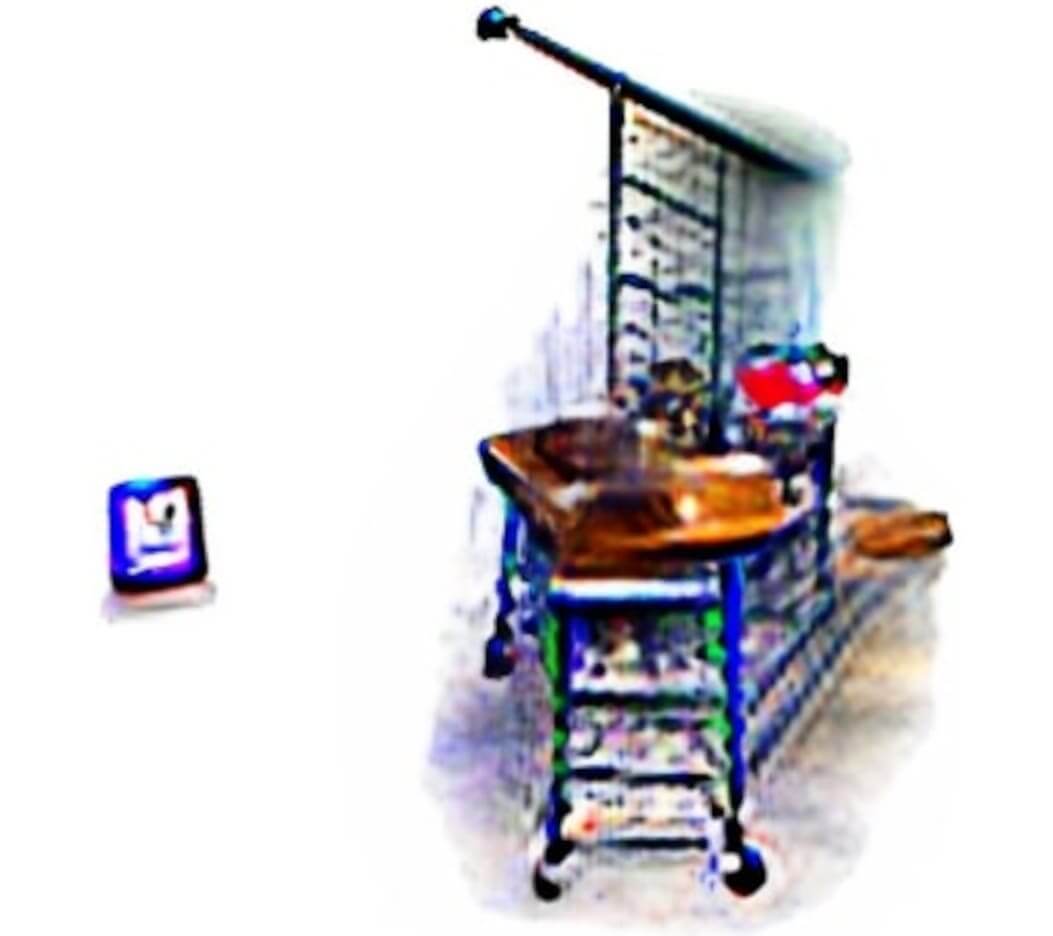} &
\includegraphics[width=0.194\textwidth]{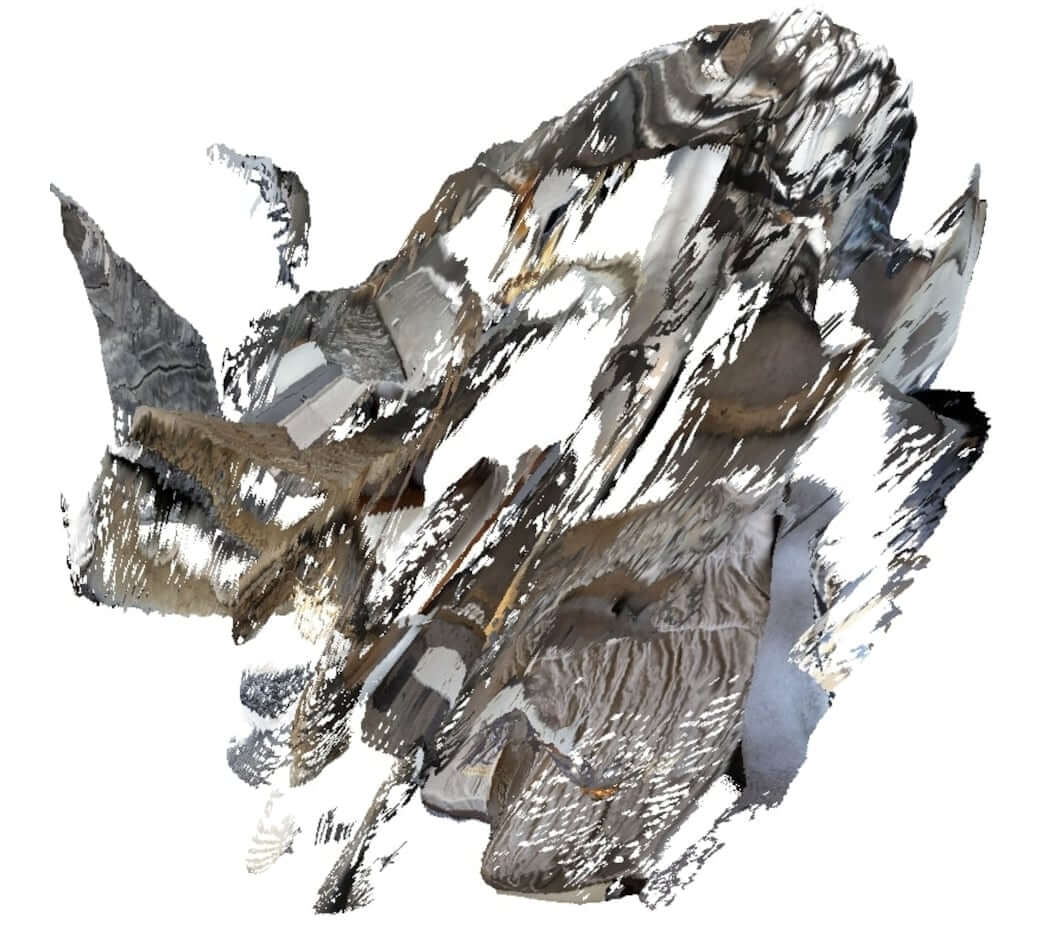} &
\includegraphics[width=0.194\textwidth]{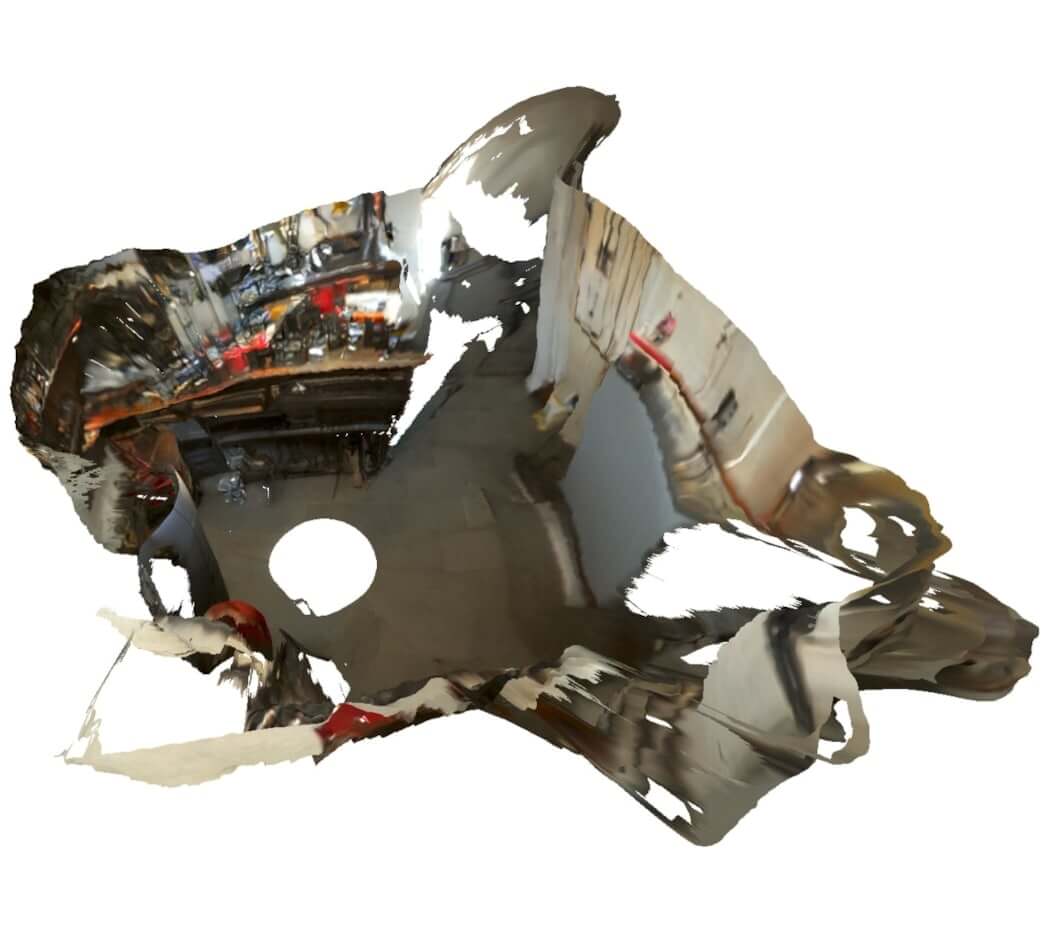} &
\includegraphics[width=0.194\textwidth]{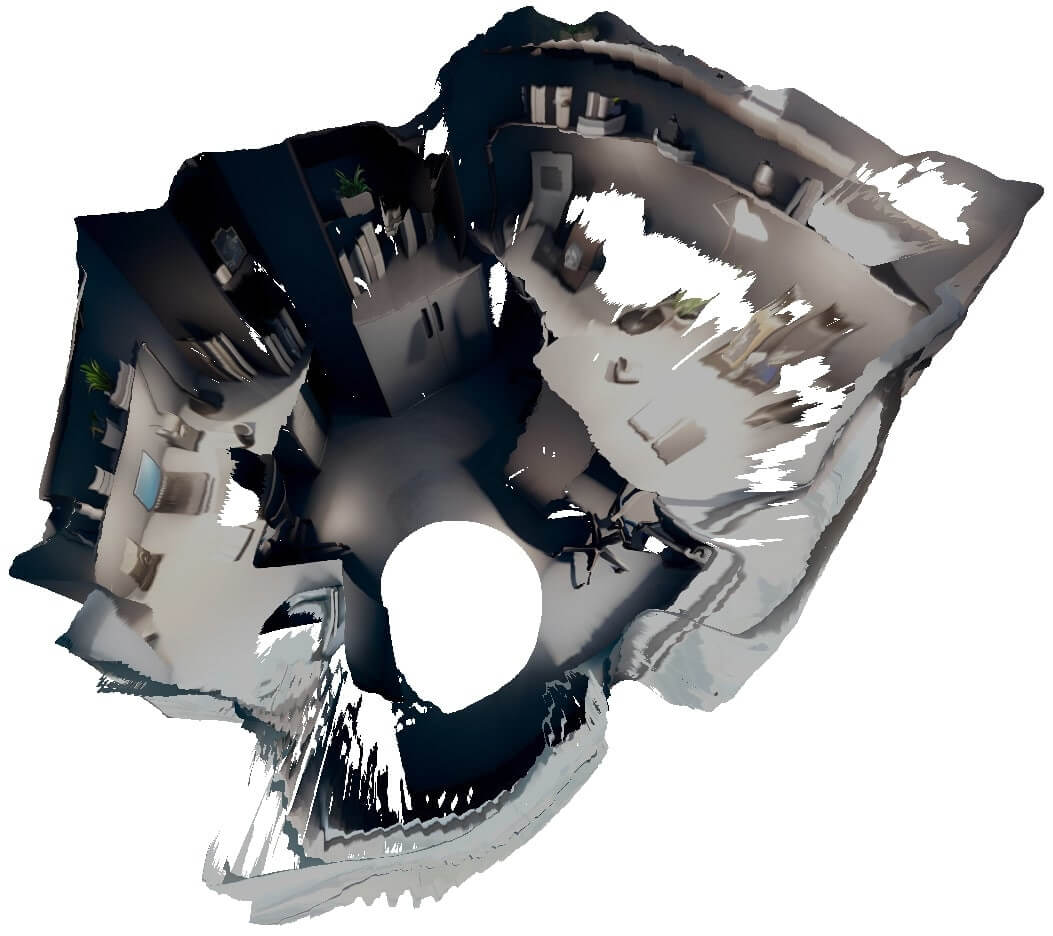} &
\includegraphics[width=0.194\textwidth]{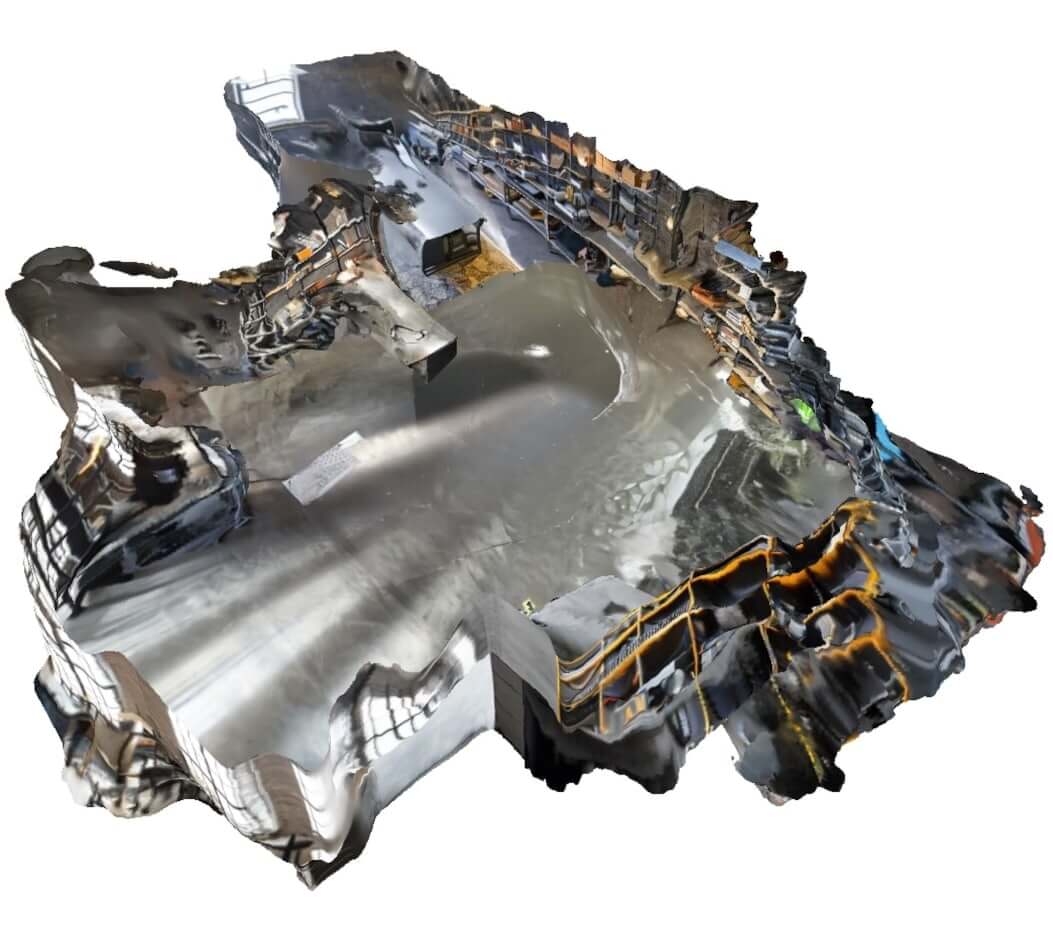} \\ 
\includegraphics[width=0.194\textwidth]{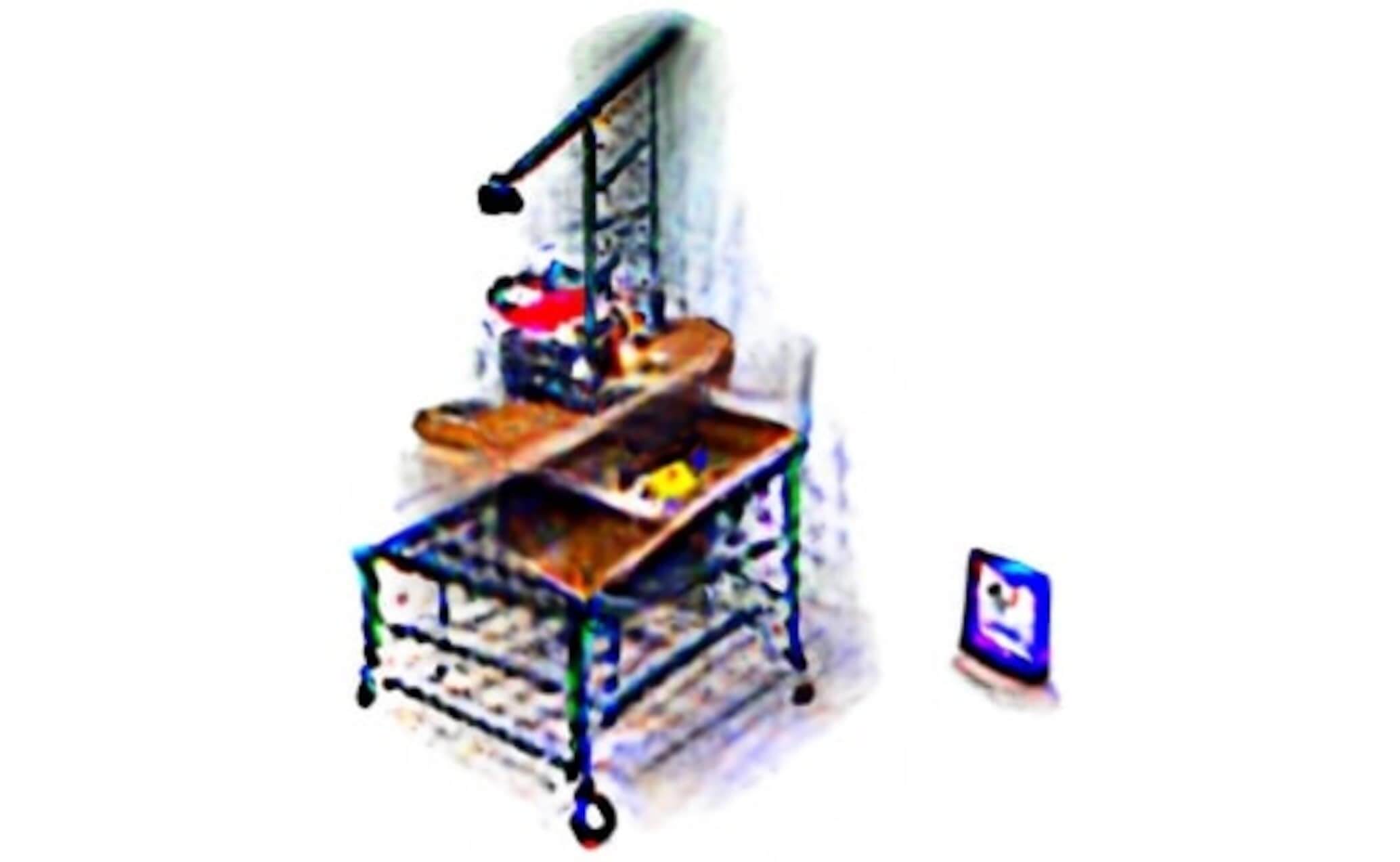} &
\includegraphics[width=0.194\textwidth]{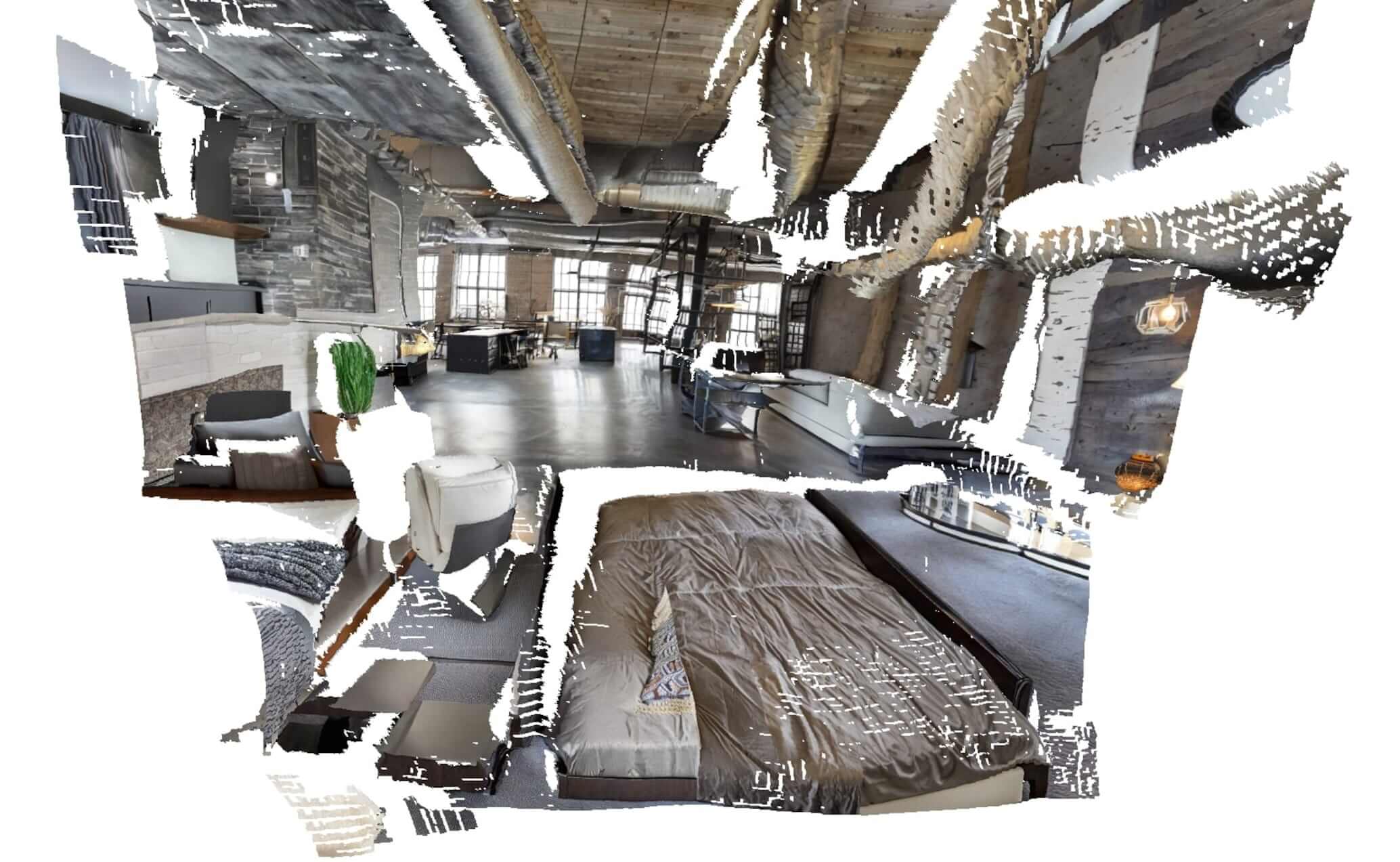} &
\includegraphics[width=0.194\textwidth]{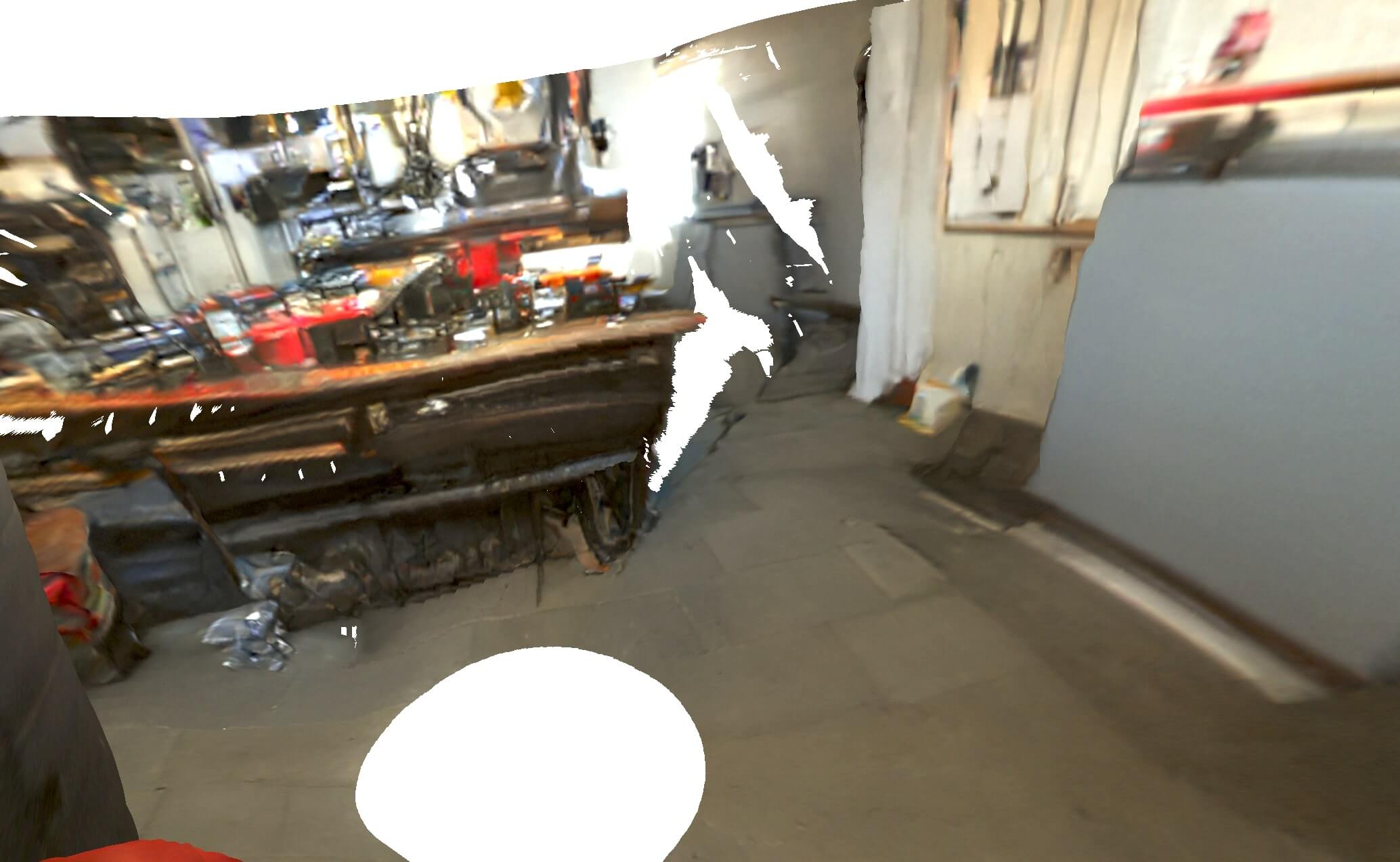} &
\includegraphics[width=0.194\textwidth]{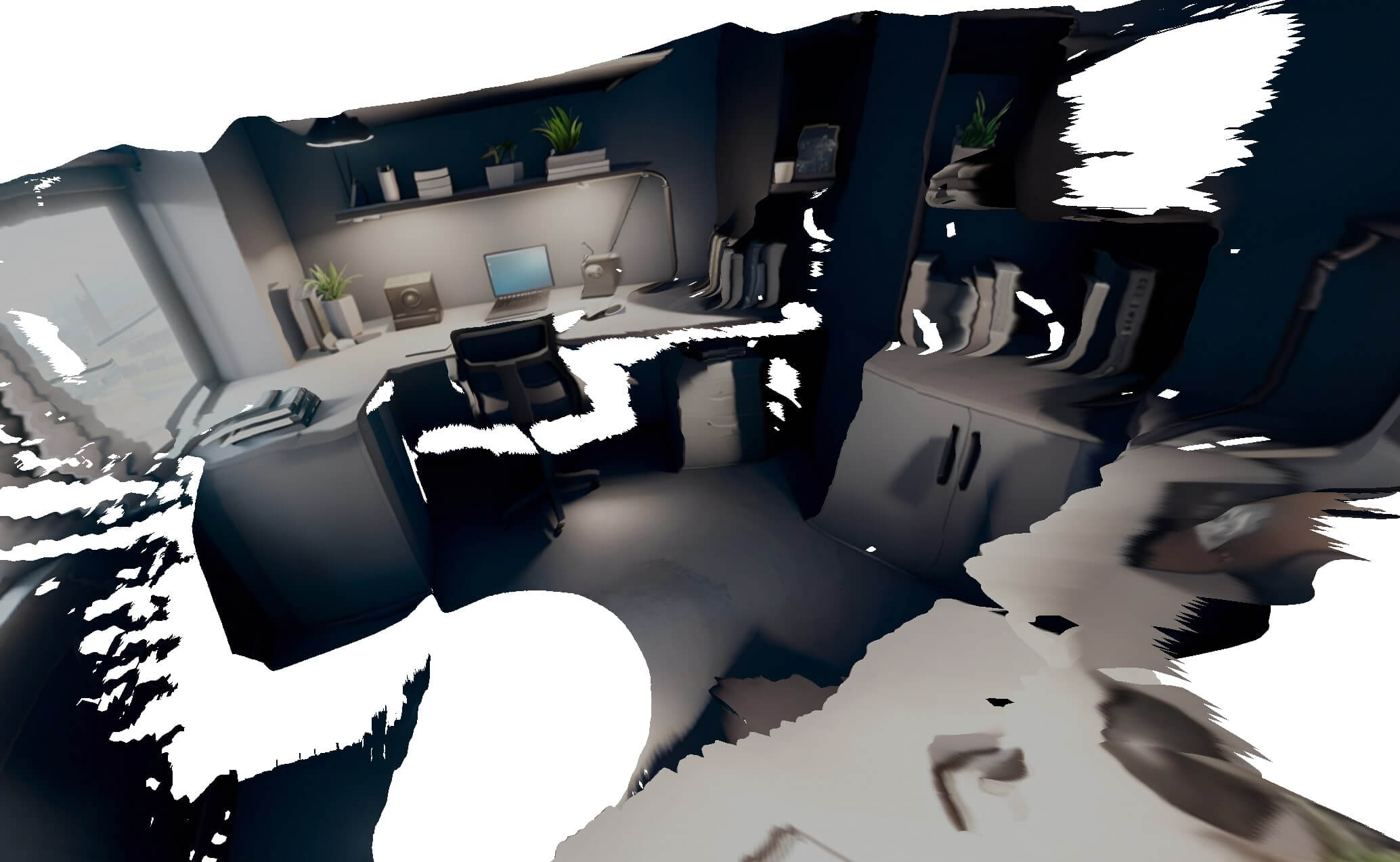} &
\includegraphics[width=0.194\textwidth]{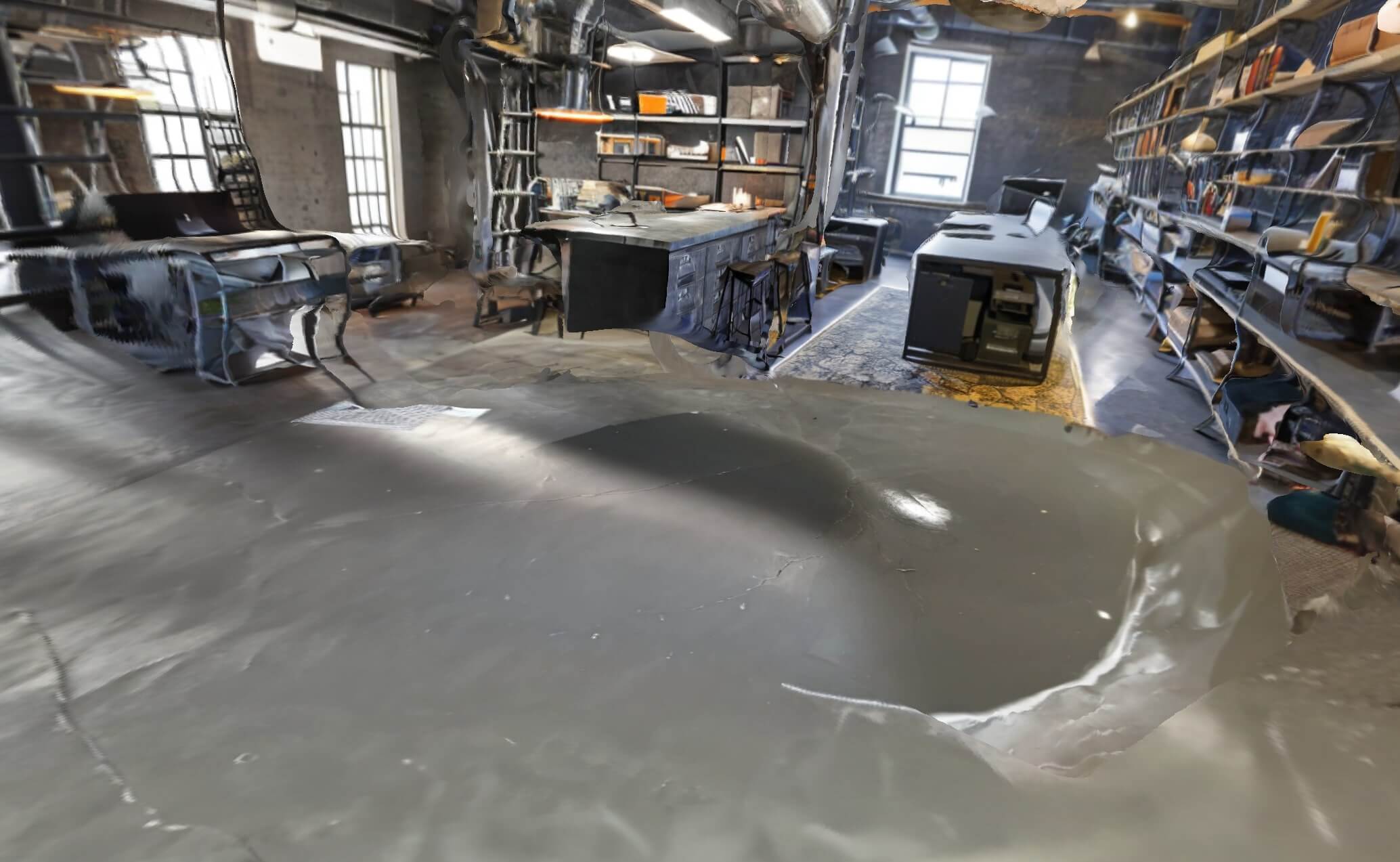} \\ 
\multicolumn{5}{c}{\textit{Editorial Style Photo, Industrial Home Office, Steel Shelves, Concrete, Metal, Edison Bulbs, Exposed Ductwork}} \\

\includegraphics[width=0.194\textwidth]{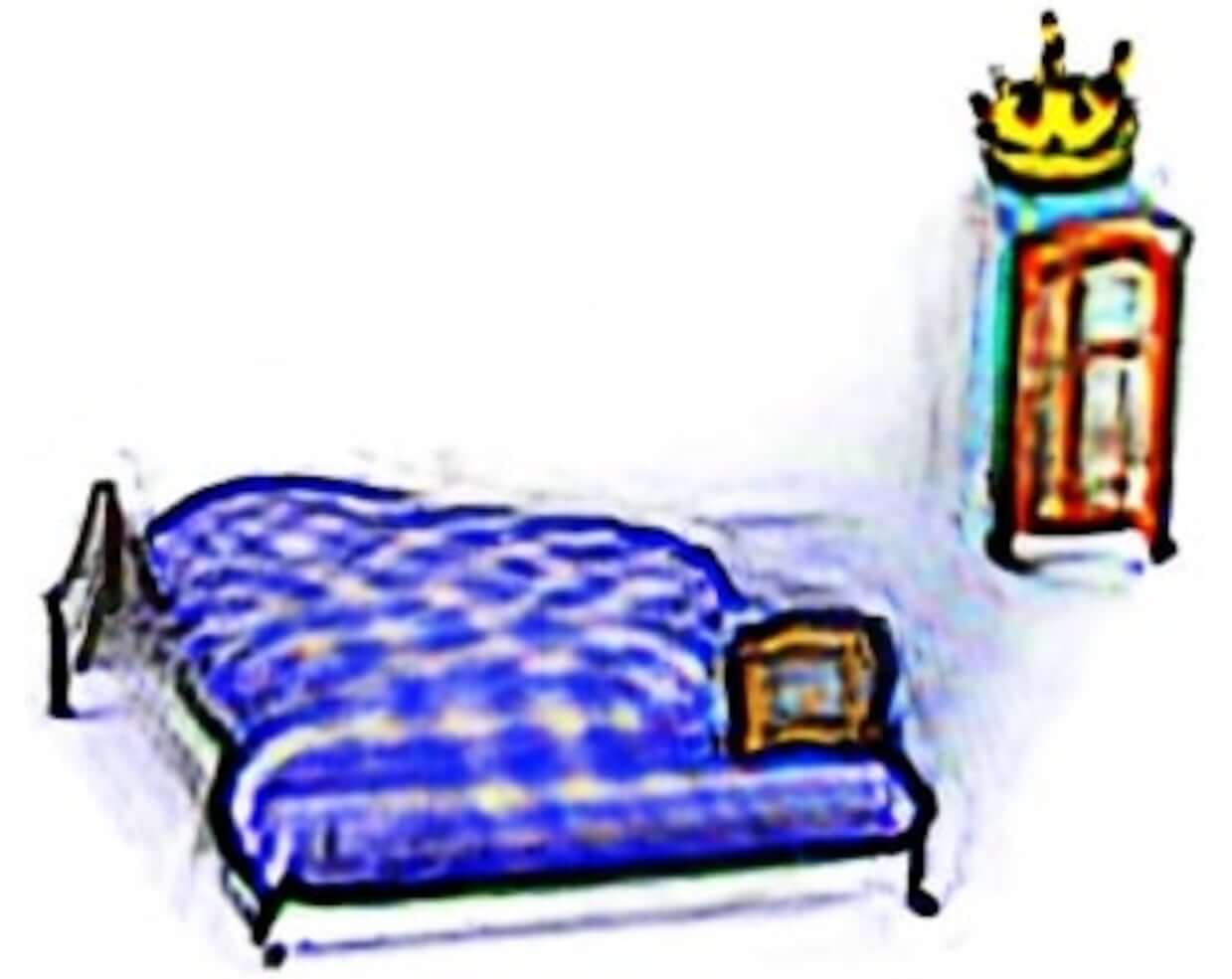} &
\includegraphics[width=0.194\textwidth]{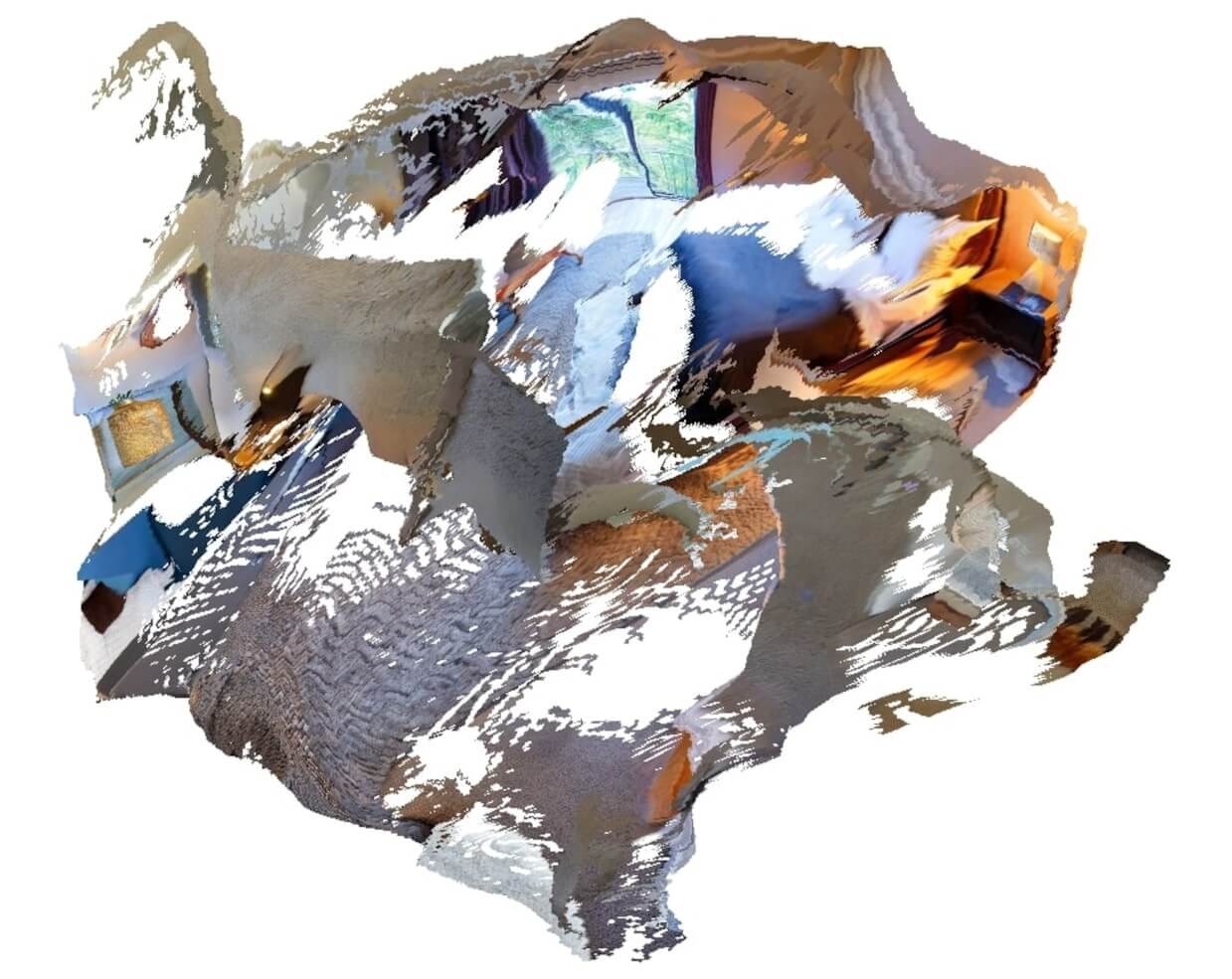} &
\includegraphics[width=0.194\textwidth]{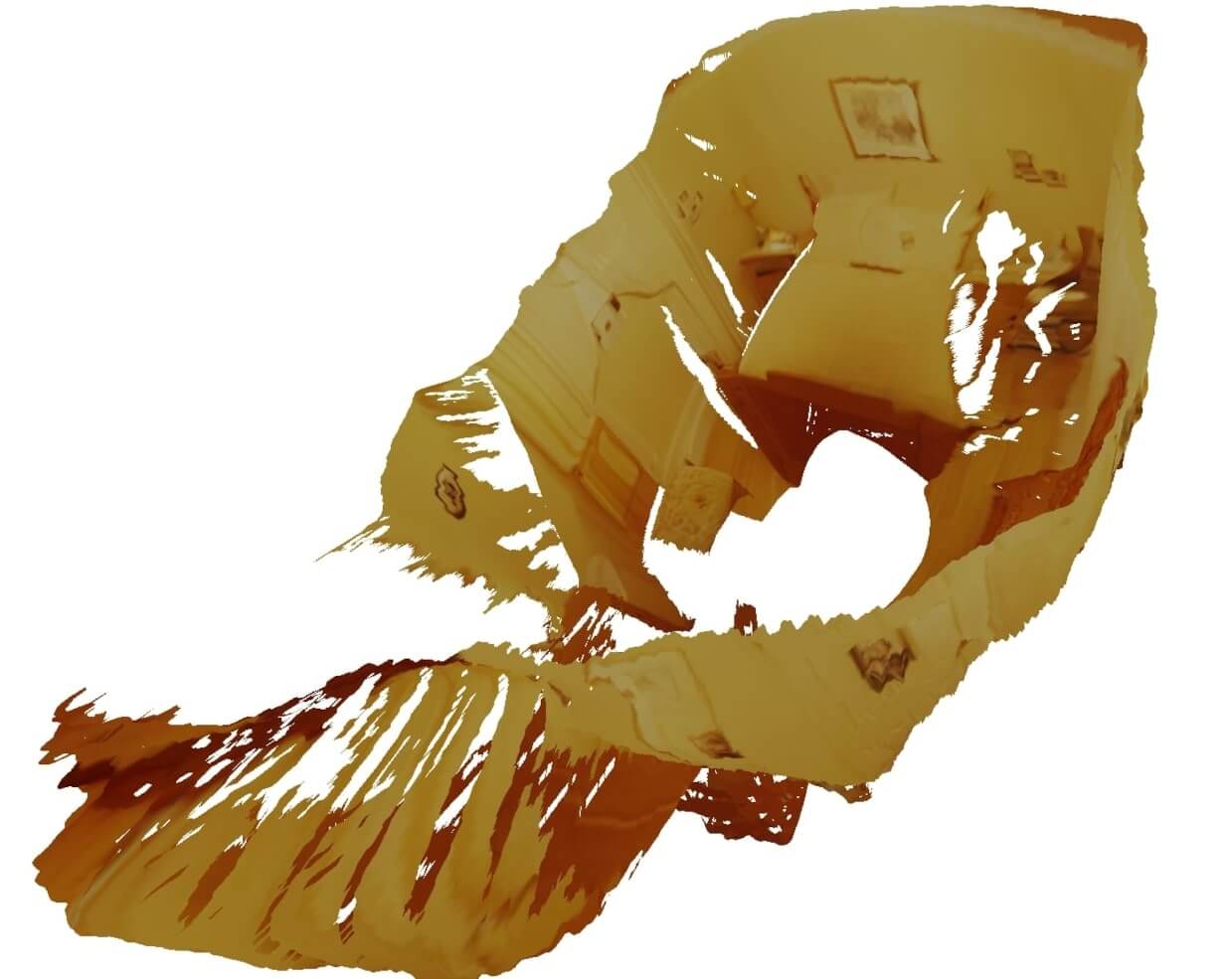} &
\includegraphics[width=0.194\textwidth]{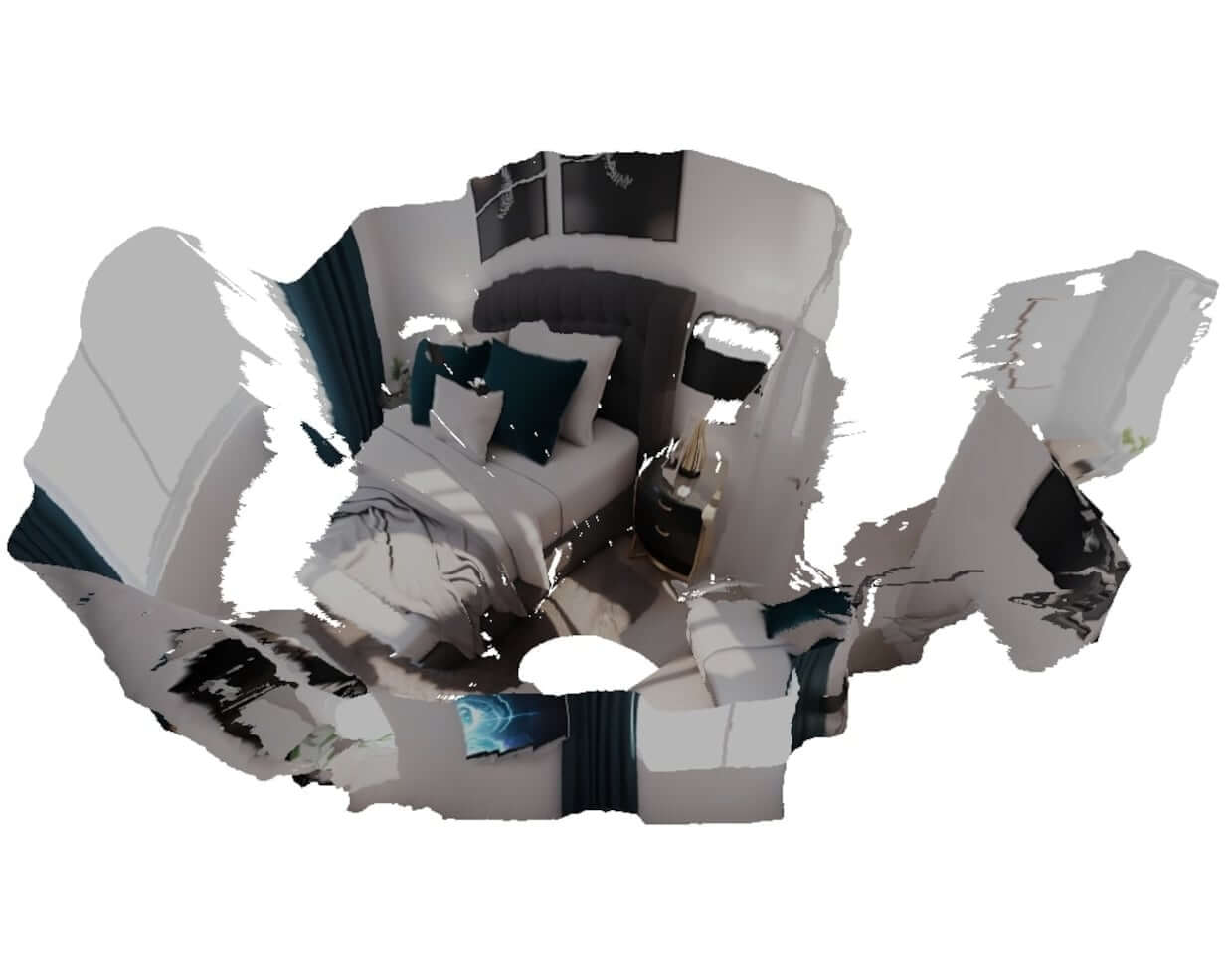} &
\includegraphics[width=0.194\textwidth]{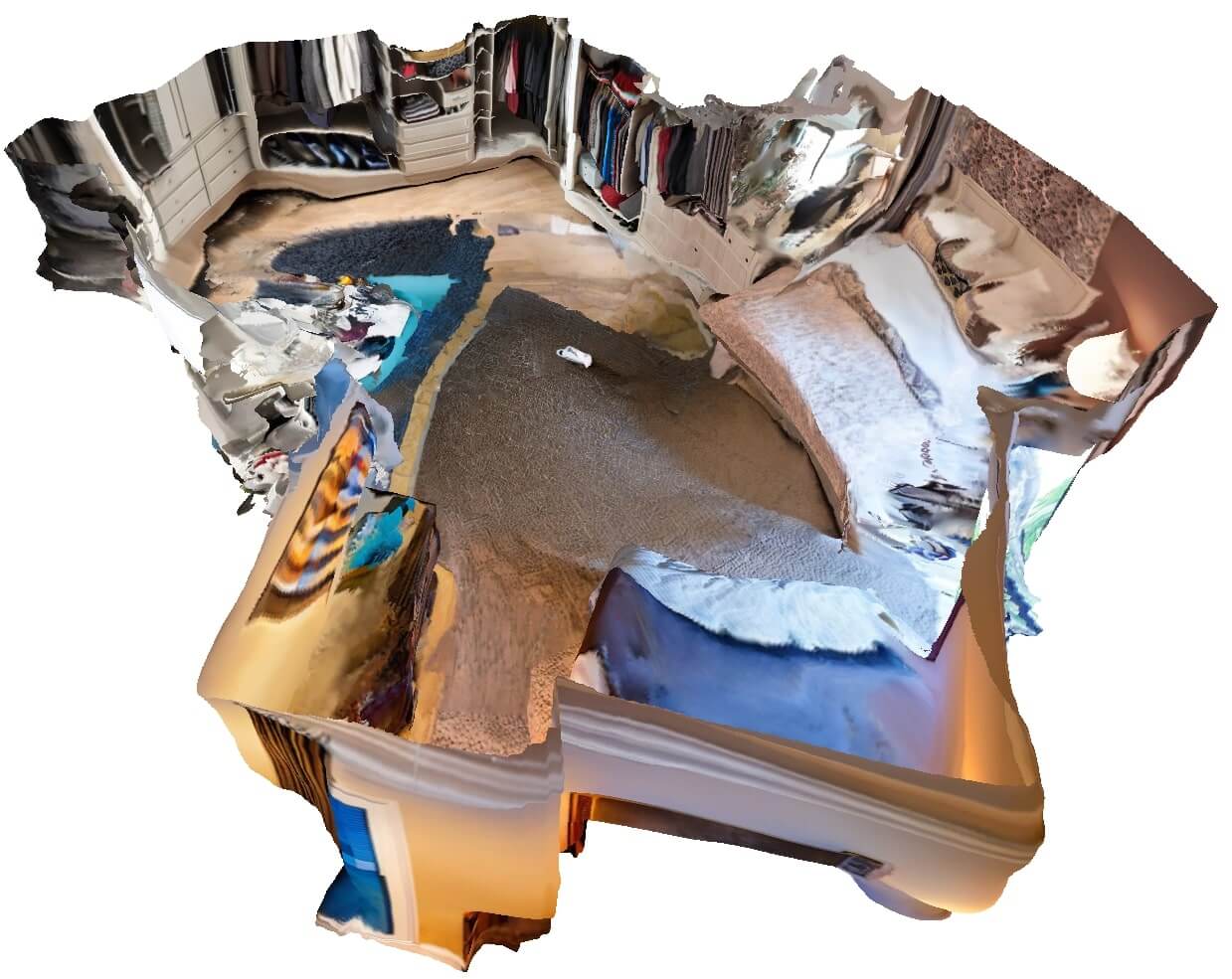} \\ 
\includegraphics[width=0.194\textwidth]{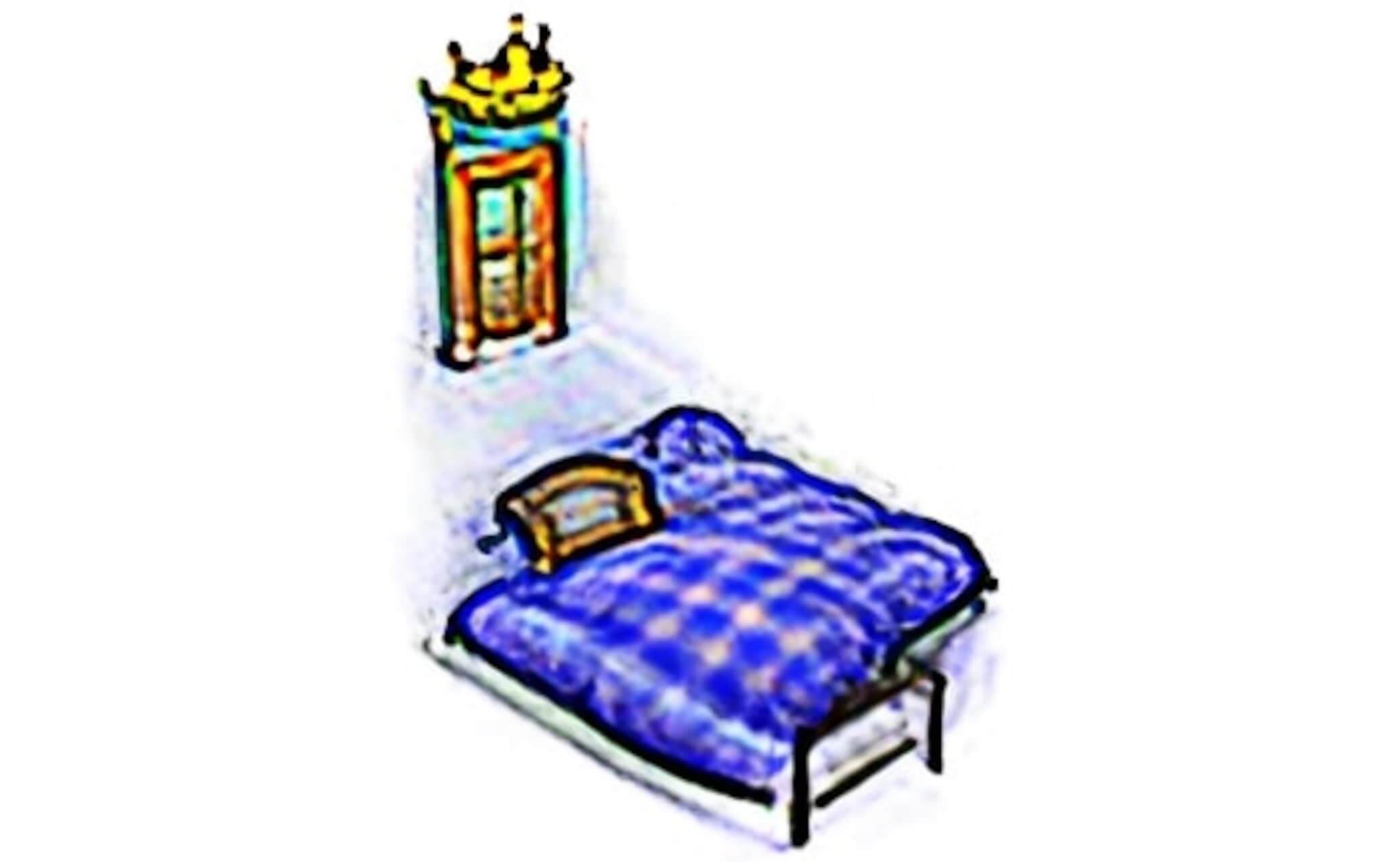} &
\includegraphics[width=0.194\textwidth]{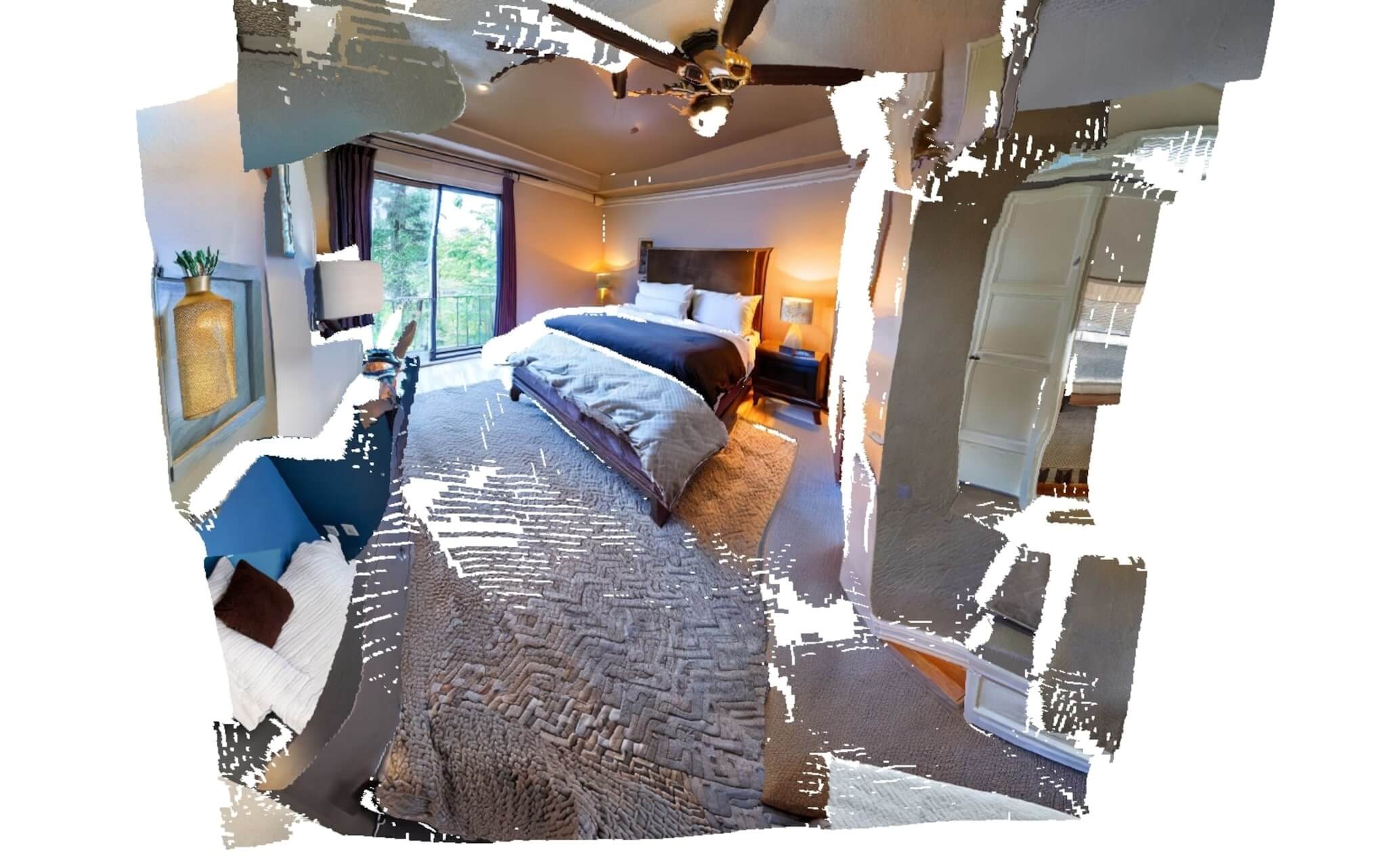} &
\includegraphics[width=0.194\textwidth]{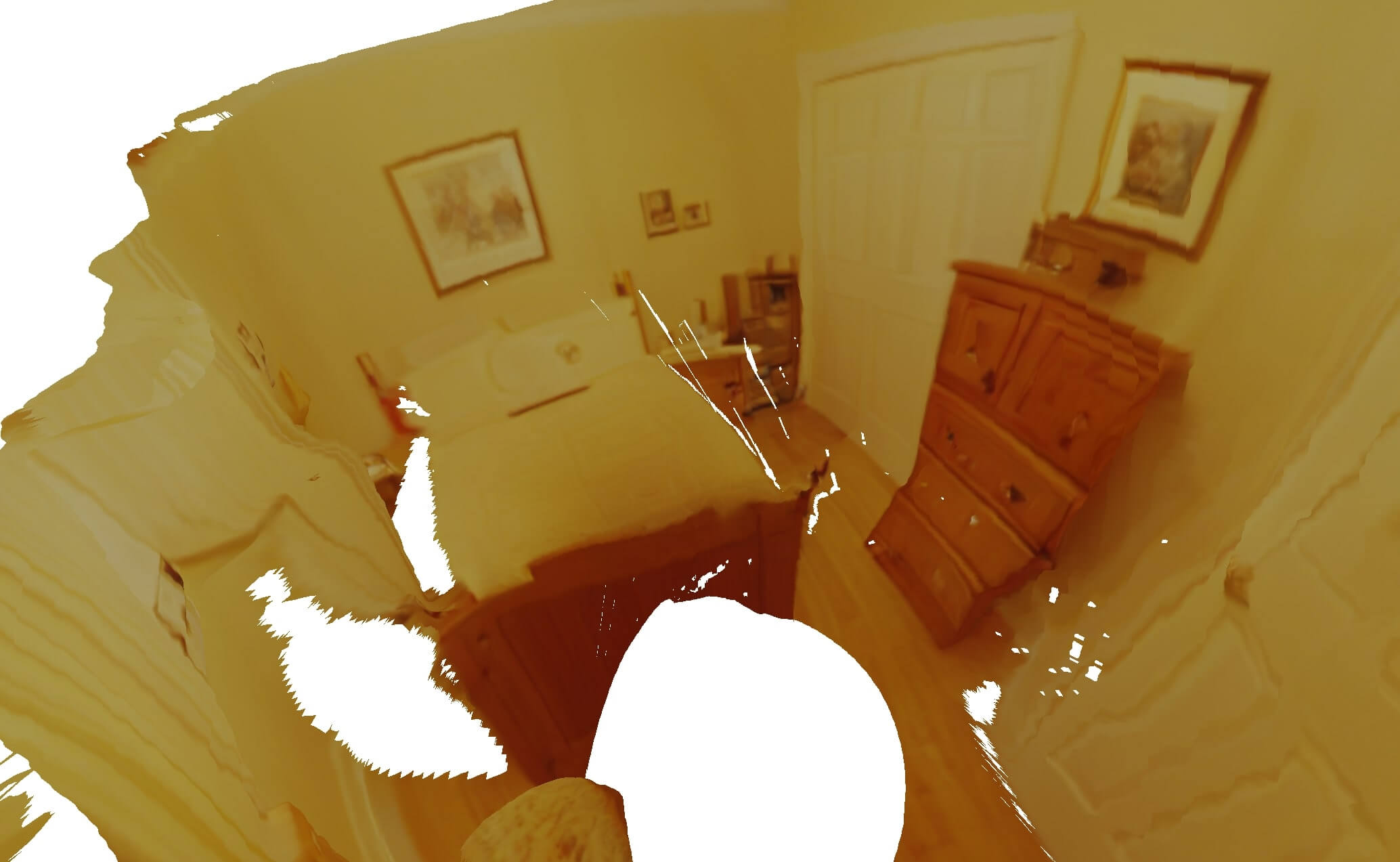} &
\includegraphics[width=0.194\textwidth]{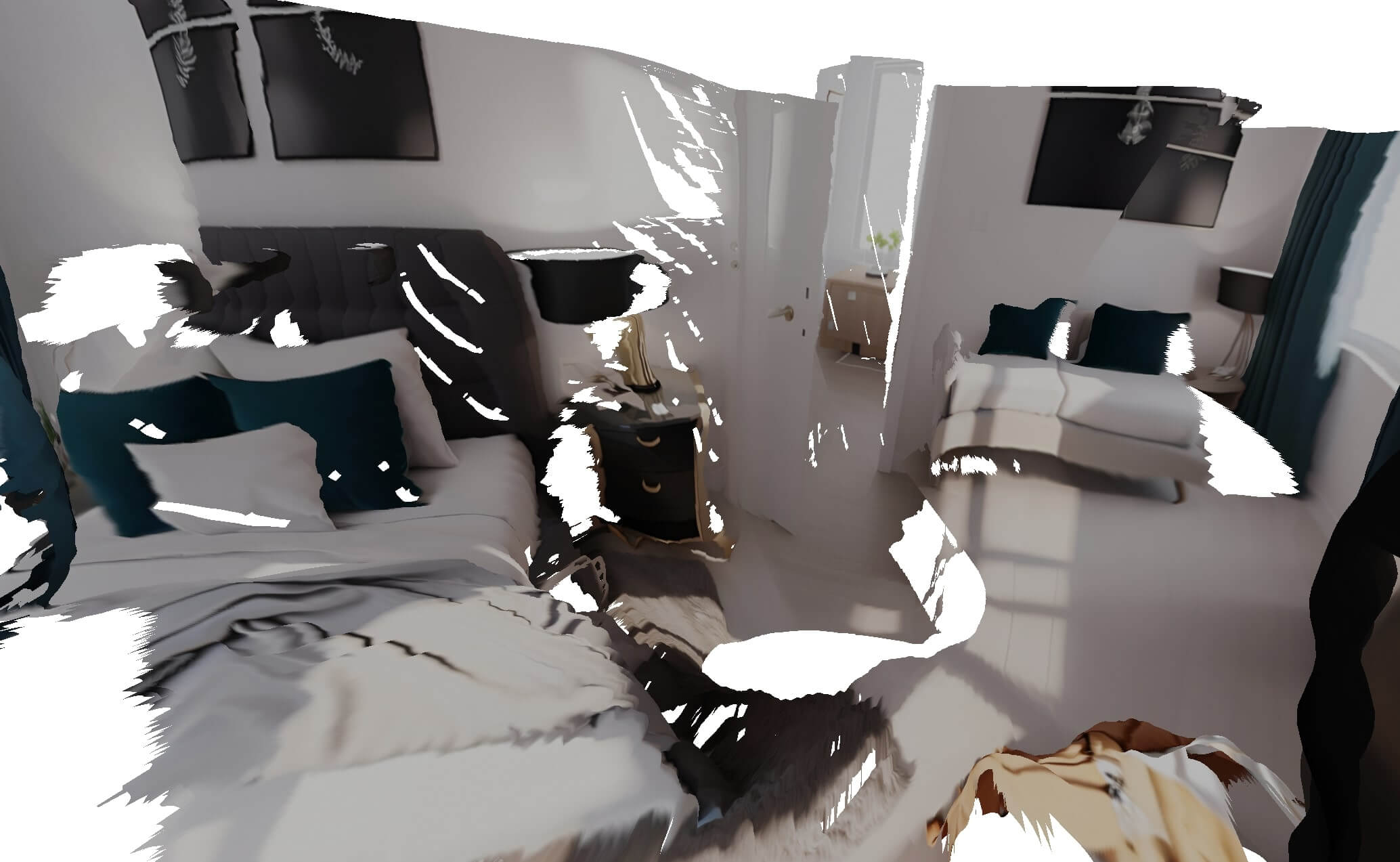} &
\includegraphics[width=0.194\textwidth]{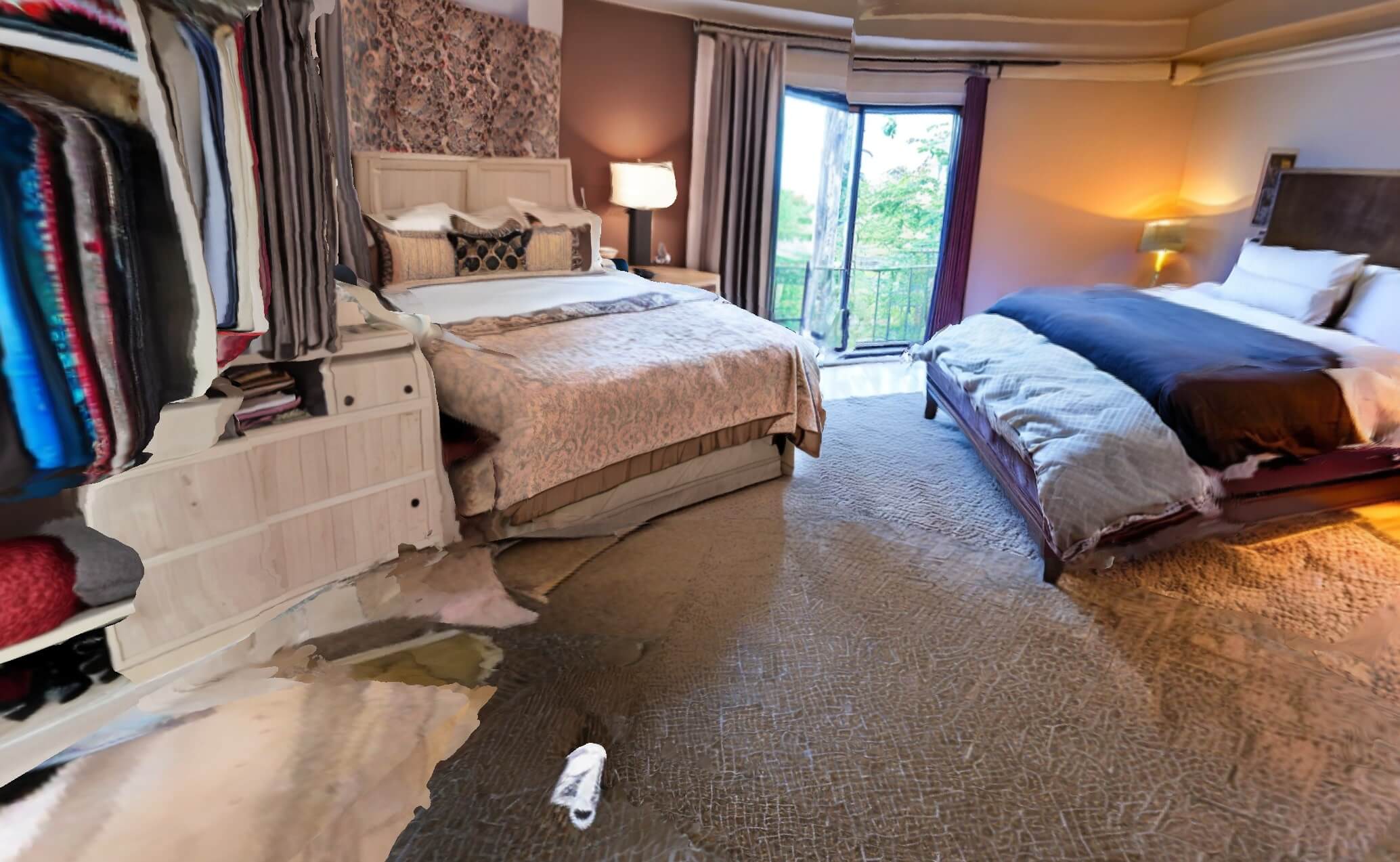} \\ 
\multicolumn{5}{c}{\textit{a bedroom with a king-size bed and a large wardrobe}} \\

PureClipNeRF~\cite{lee2022understanding} & Outpainting~\cite{Ramesh2022HierarchicalTI, OpenAI2022} & Text2Light~\cite{chen2022text2light}${+}$Ours & Blockade~\cite{blockade}${+}$Ours & Ours
\end{tabular}
\vspace{0.5mm}
\caption{
\textbf{
Qualitative comparison of our method and baselines.} \emph{PureClipNeRF}~\cite{lee2022understanding} cannot produce immersive scenes with floors and walls. \emph{Outpainting}~\cite{Ramesh2022HierarchicalTI, OpenAI2022} does not produce 3D consistent scenes. \emph{Text2Light}~\cite{chen2022text2light} and \emph{Blockade}~\cite{blockade} both have holes due to occlusions. In contrast, our method creates complete meshes without holes and high details. We remove the ceiling in the top-down view for better visualization of the scene layout.}
\label{fig:ours-baselines}
\end{figure*}

We show top-down views into the scene and RGB renderings from within for our method and baselines in Figure~\ref{fig:ours-baselines}.
We show additional results of our method in Figure~\ref{fig:ours-only}.
\emph{PureClipNeRF}~\cite{lee2022understanding} creates the key objects of the given text prompt, but does not create a complete 3D structure with floor, walls and ceilings.
\emph{Outpainting}~\cite{Ramesh2022HierarchicalTI, OpenAI2022} creates high-detail textures, but projection from a single viewpoint creates holes due to occlusion and hinders the creation of complete 3D geometry.
\emph{Text2Light}~\cite{chen2022text2light} and \emph{Blockade}~\cite{blockade} both create a high-detail $360^\circ$ view of a complete scene, but occlusions that cannot be resolved from a single panoramic viewpoint lead to holes in the extracted 3D geometry.

In contrast, our approach creates high-detail textures and geometry, that are fused into a complete 3D scene mesh without holes.
The resulting scenes contain flat floors, walls and ceilings, as well as 3D object geometry distributed throughout the scene.
When specifying text prompts with a huge variety, the resulting scene contains a diverse set of objects.
Please see the supplemental material for more scenes, animated results, intermediate outputs of our baselines (such as the panoramic images) as well as top-down views of meshes, that contain the reconstructed ceilings.

\subsection{Quantitative Results}

\begin{table}
  \centering
  \resizebox{0.48\textwidth}{!}{
  \begin{tabular}{l cc cc}
    \toprule
        \multirow{2}{*}{Method} & \multicolumn{2}{c}{2D Metrics} & \multicolumn{2}{c}{User Study}\\
                        \cmidrule(l{2pt}r{2pt}){2-3} \cmidrule(l{2pt}r{2pt}){4-5}
    & CS $\uparrow$ & IS $\uparrow$ & PQ $\uparrow$ & 3DS $\uparrow$\\
    \midrule
    PureClipNeRF~\cite{lee2022understanding} & 24.06 & 1.26 & 2.34 & 2.38\\
    Outpainting~\cite{Ramesh2022HierarchicalTI, OpenAI2022} & 23.10 & 1.60 & 2.90 & 2.58\\
    Text2Light~\cite{chen2022text2light}${+}$Ours & 25.99 & 2.21 & 2.82 & 2.97\\
    Blockade~\cite{blockade}${+}$Ours & 26.29 & 2.13 & 3.35 & 3.36\\
    \midrule
    Ours w/o alignment & 26.73 & 1.78 & 3.12 & 2.96\\
    Ours w/o stretch removal & 27.72 & 1.86 & 3.28 & 3.75\\
    Ours w/o completion & 27.97 & 2.18 & 3.72 & 3.87\\
    Ours & \textbf{28.02} & \textbf{2.31} & \textbf{4.01} & \textbf{4.19}\\
    \bottomrule
  \end{tabular}
  }
  \vspace{2mm}
  \caption{
  \textbf{Quantitative comparison.}
  We report 2D metrics and user study results, including: Clip Score~(\emph{CS}), Inception Score~(\emph{IS}), Perceptual Quality (\emph{PQ}), and 3D Structure Completeness ~(\emph{3DS}).
  Our method creates scenes with the highest quality.}\label{tab:ours-baseline}
\end{table}

We show quantitative results averaged over multiple scenes in Table~\ref{tab:ours-baseline}.
We render 60 images from novel viewpoints for each scene to calculate the 2D metrics.
We present users with multiple top-down views and renderings for each scene and let them rate each method individually (no side-by-side comparison).
Stretched-out geometry and holes in the 3D geometry lead to lower scores for the baselines in all image-based metrics.
Our approach achieves the highest scores, because the renderings are complete from arbitrary novel poses, satisfy the given text-prompt and contain high-resolution image features.
Users prefer our method, which highlights the quality of our accurate and complete geometry, as well as the RGB texture.

\subsection{Ablations}

\begin{figure*}
\centering
\setlength\tabcolsep{1pt}
\begin{tabular}{cccc}
\includegraphics[width=0.24\textwidth]{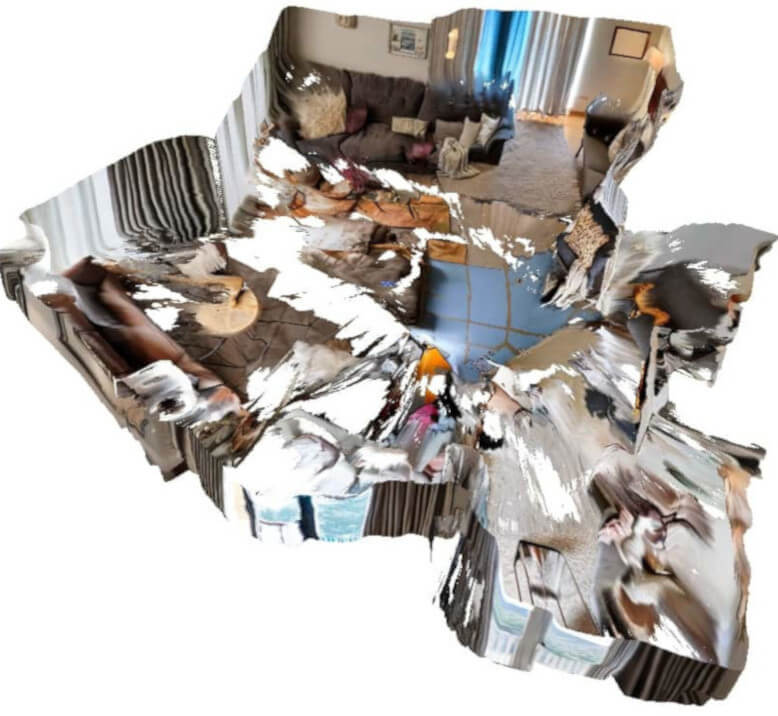} &
\includegraphics[width=0.24\textwidth]{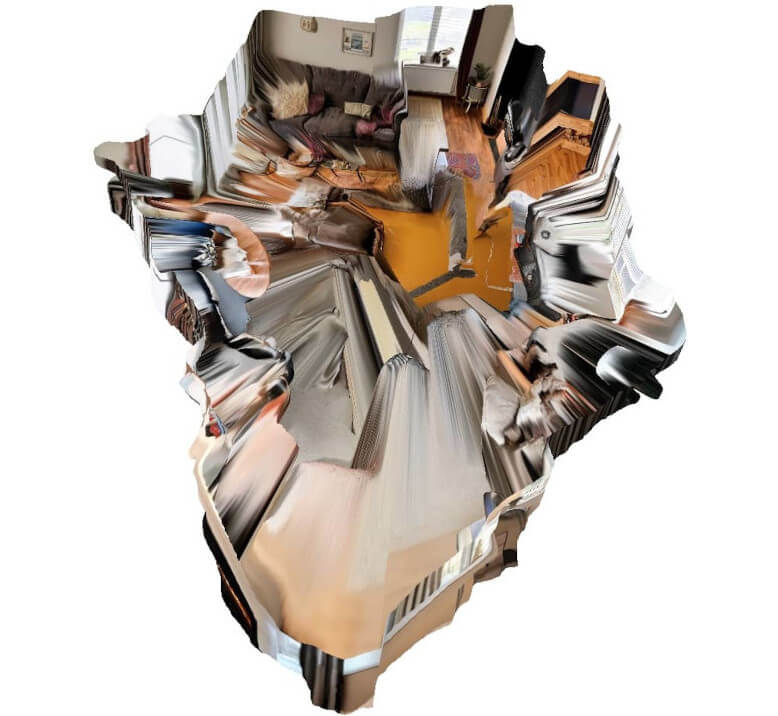} &
\includegraphics[width=0.24\textwidth]{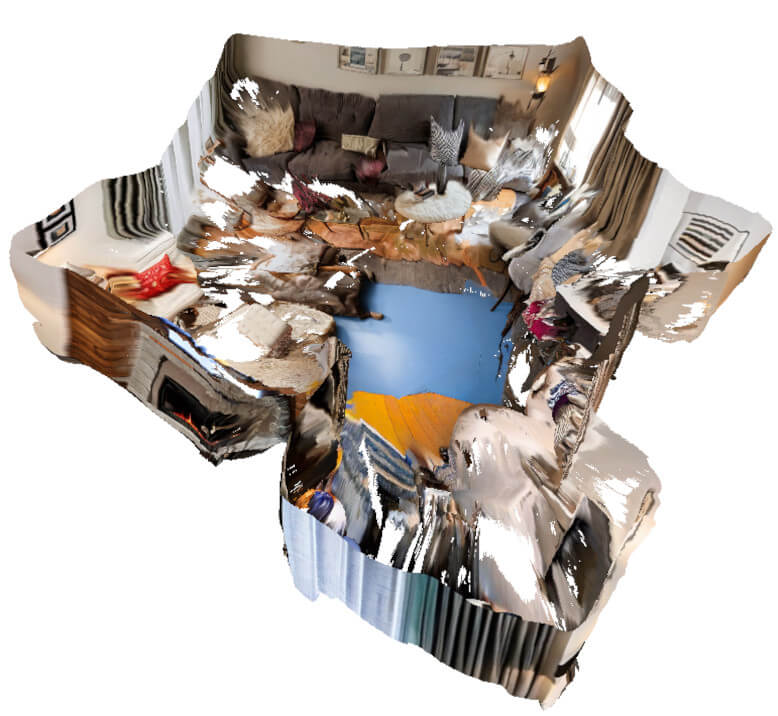} &
\includegraphics[width=0.24\textwidth]{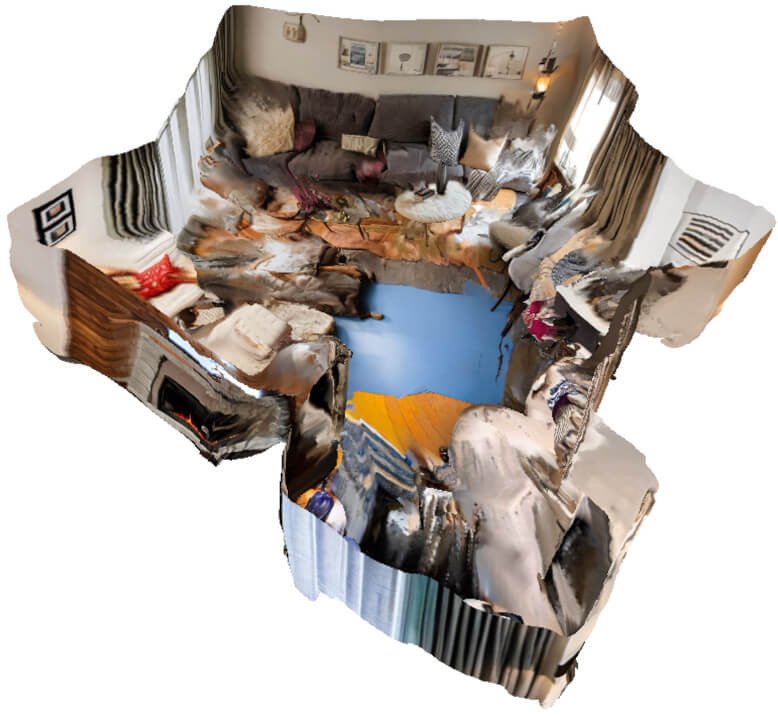} \\
\includegraphics[width=0.24\textwidth]{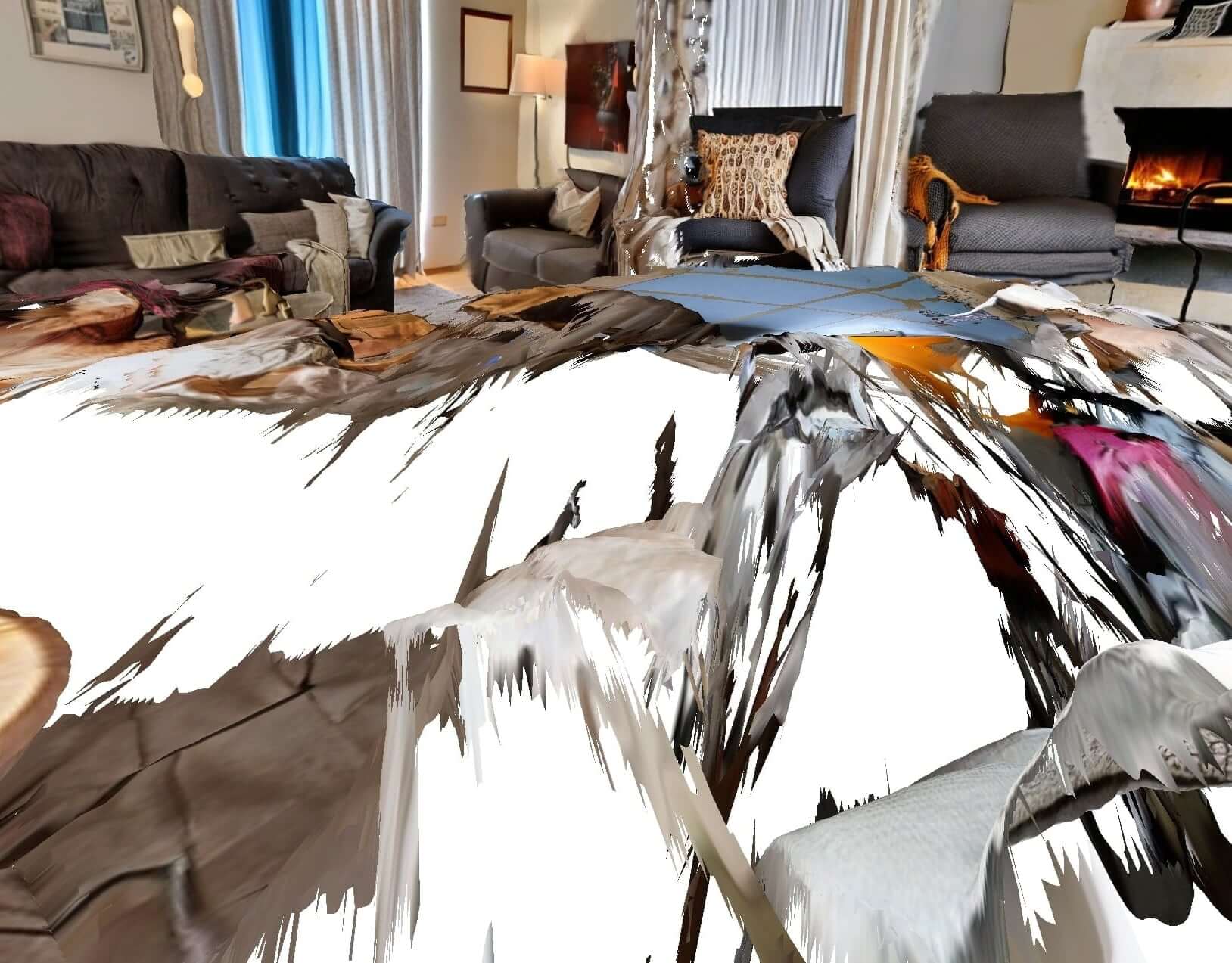} &
\includegraphics[width=0.24\textwidth]{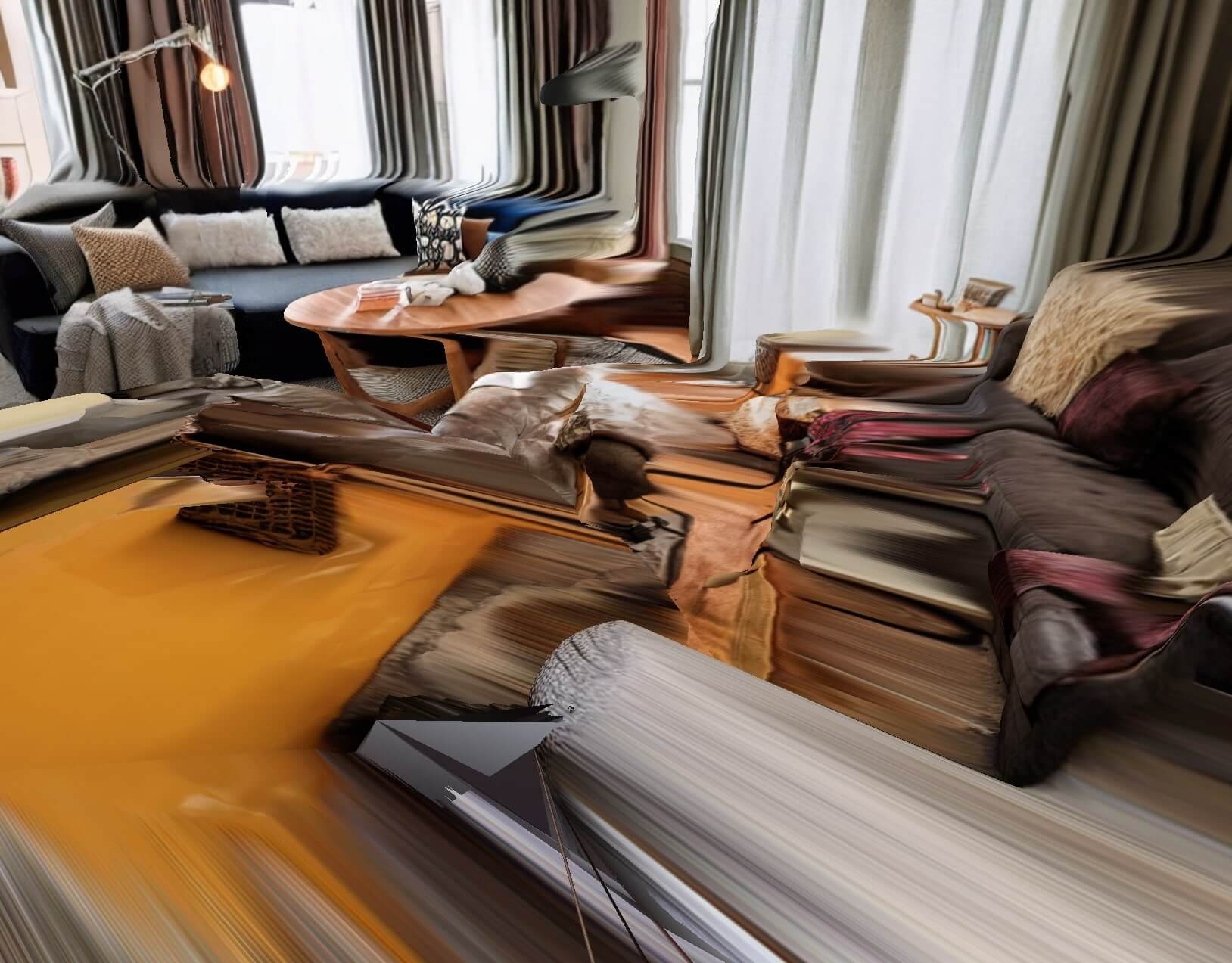} &
\includegraphics[width=0.24\textwidth]{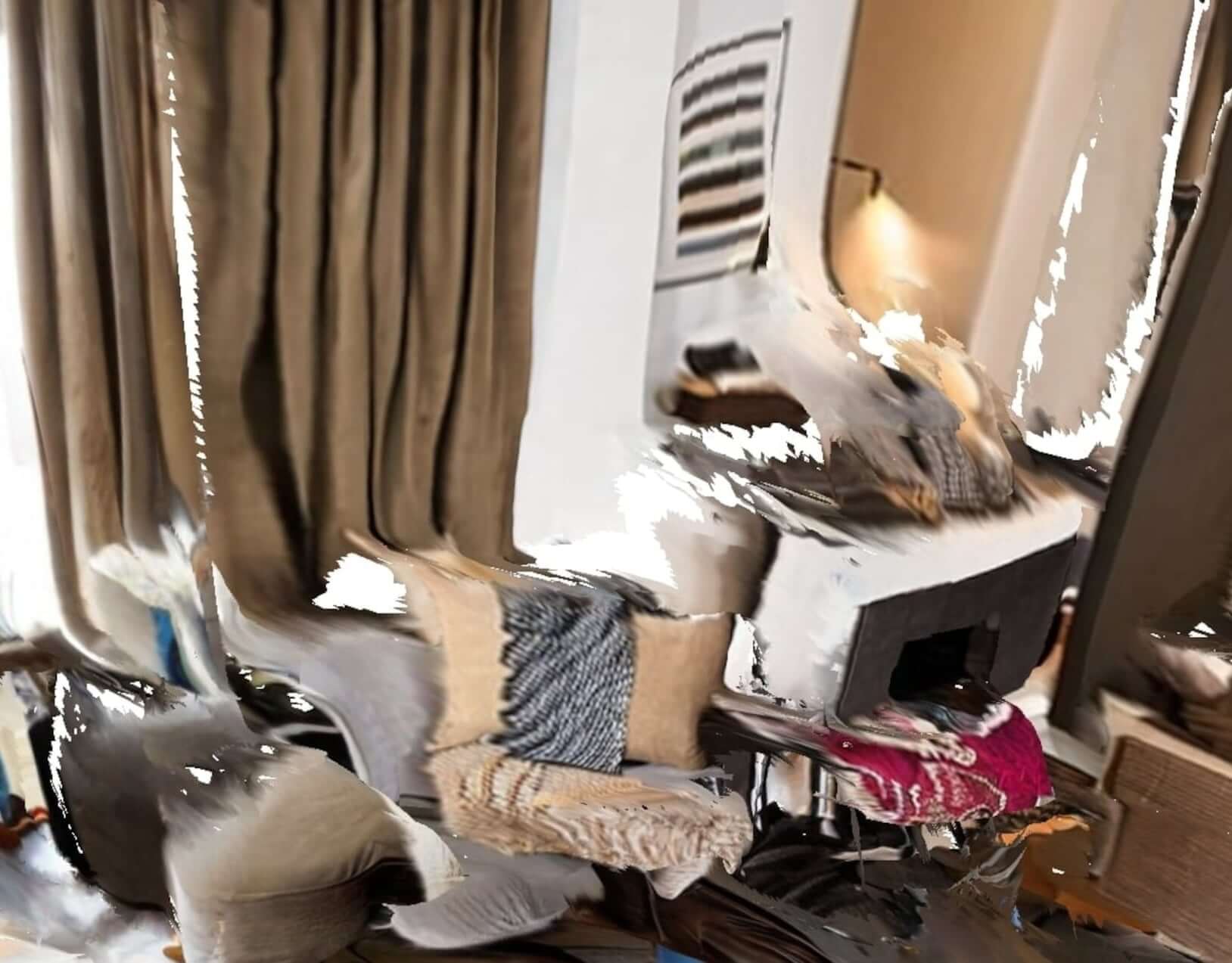} &
\includegraphics[width=0.24\textwidth]{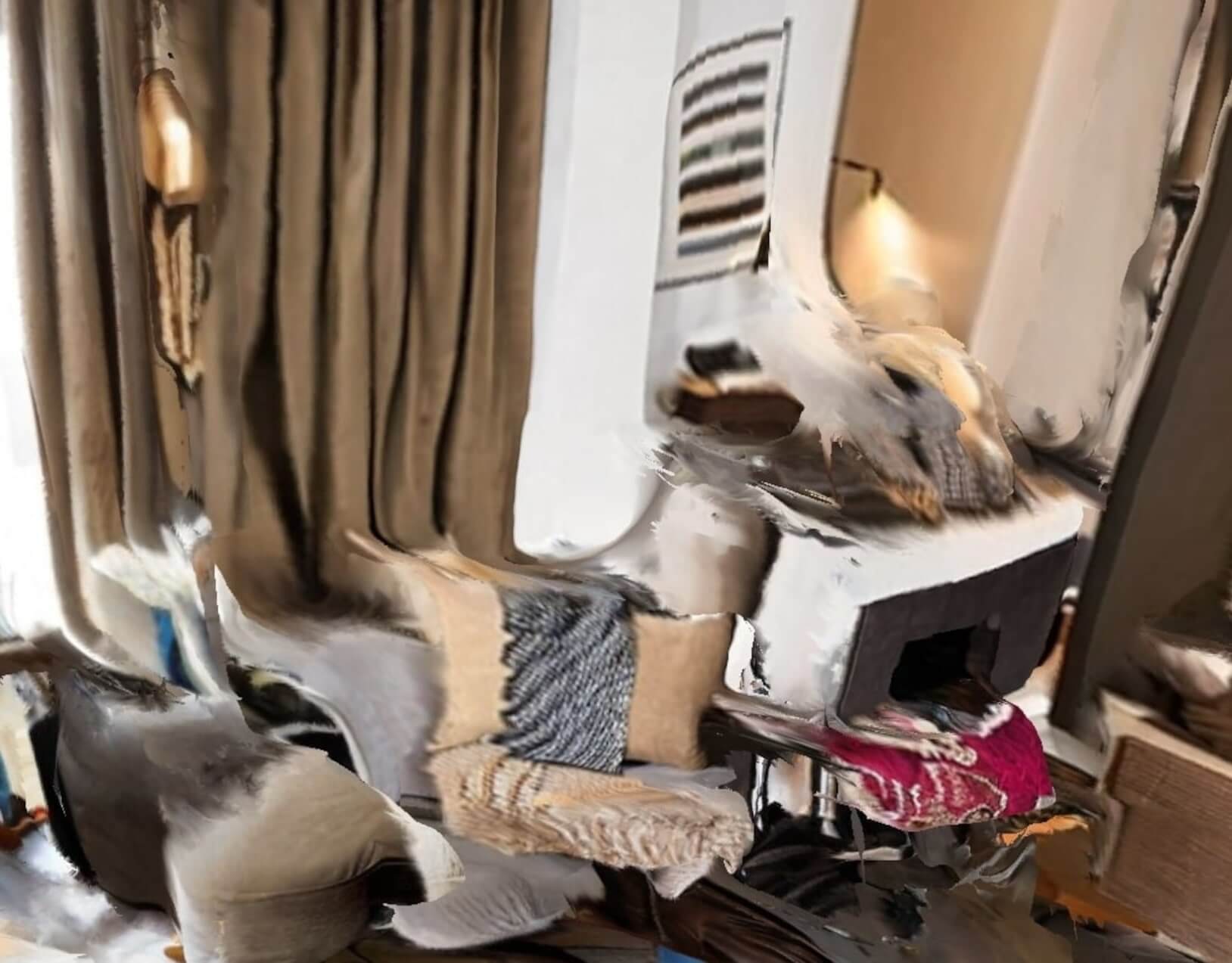} \\
\multicolumn{4}{c}{\textit{living room, couches, curtains, lit furnace, lamps}} \\
(a) Ours w/o alignment & (b) Ours w/o stretch removal & (c) Ours w/o completion & (d) Ours~(full) \\

\end{tabular}
\vspace{.5mm}
\caption{
\textbf{Ablation study on the key components of our method.}
Without depth alignment (see Section \ref{subsec:Depth Alignment}), different parts of the scene are disconnected and do not fuse into a seamless mesh. Without edge and surface normal thresholds (see Section \ref{subsec:3D Mesh Generation}), many faces are stretched out unnaturally. Without completion (see Section \ref{subsec:Trajectory Generation}), the mesh has holes in remaining unobserved regions. Our full pipeline creates complete, high-resolution scenes. We remove the ceiling in the top-down view for better visualization of the scene layout.}
\label{fig:ours-ablation}
\end{figure*}

The key ingredients of our method are depth alignment (Section~\ref{subsec:Depth Alignment}), mesh fusion (Section~\ref{subsec:3D Mesh Generation}) and the two-stage viewpoint selection (Section~\ref{subsec:Trajectory Generation}).
We demonstrate the importance of each component in Figure~\ref{fig:ours-ablation} and Table~\ref{tab:ours-baseline}.

\mypar{Depth alignment creates seamless scenes}
Monocular depth predictions from subsequent frames can be inconsistent in scale.
This leads to disconnected components in the mesh that are backprojected from multiple viewpoints (see Figure~\ref{fig:ours-ablation}a).
Our depth alignment strategy allows fusing multiple frames into a seamless mesh, eventually creating a complete scene with flat floors, walls, ceilings and no holes.

\mypar{Stretch removal creates undistorted scene geometry}
During mesh fusion, we update the scene geometry with the contents of the next frame.
Due to noisy depth prediction, the objects become stretched out, if they are observed from small grazing angles.
Thus, we propose two filters (edge length and surface normal thresholds) that alleviate this issue.
Instead of baking in stretched-out geometry (see Figure~\ref{fig:ours-ablation}b), we disregard the corresponding faces and let the object be completed from a more suitable, later viewpoint.

\mypar{Two-stage generation creates complete scenes}
Our approach chooses camera poses in two stages to create a complete scene without holes.
After generating the scene from predefined trajectories, the scene still contains some holes (see Figure~\ref{fig:ours-ablation}c).
Because the scene is built-up over time, it is impossible to choose camera poses \emph{a-priori}, that view all unobserved regions.
To this end, our completion stage samples poses \emph{a-posteriori} to refine those regions.
The resulting mesh is watertight and contains no holes (see Figure~\ref{fig:ours-ablation}d).

\subsection{Spatially Varying Scene Generation}
\label{subsec:cont-gen}
\begin{figure}
\centering
\setlength\tabcolsep{1pt}
\begin{tabular}{cc}
\includegraphics[height=24mm]{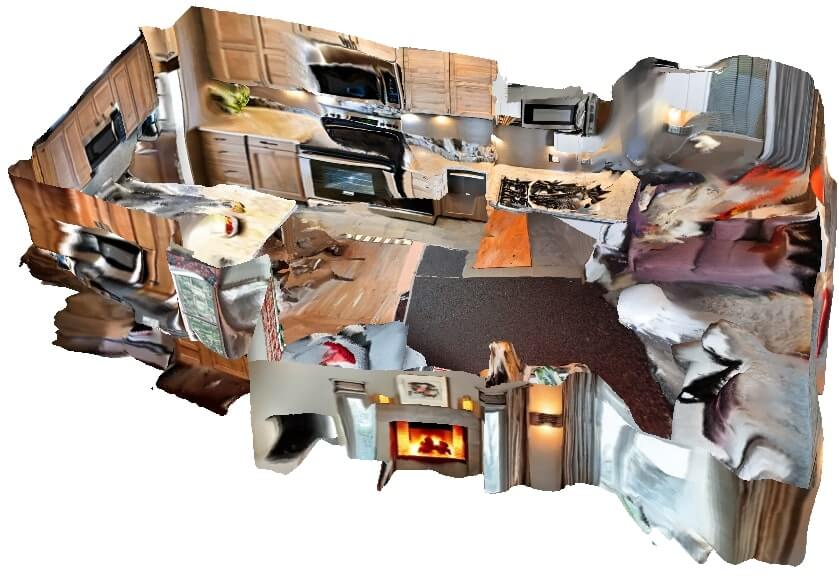} &
\includegraphics[height=24mm]{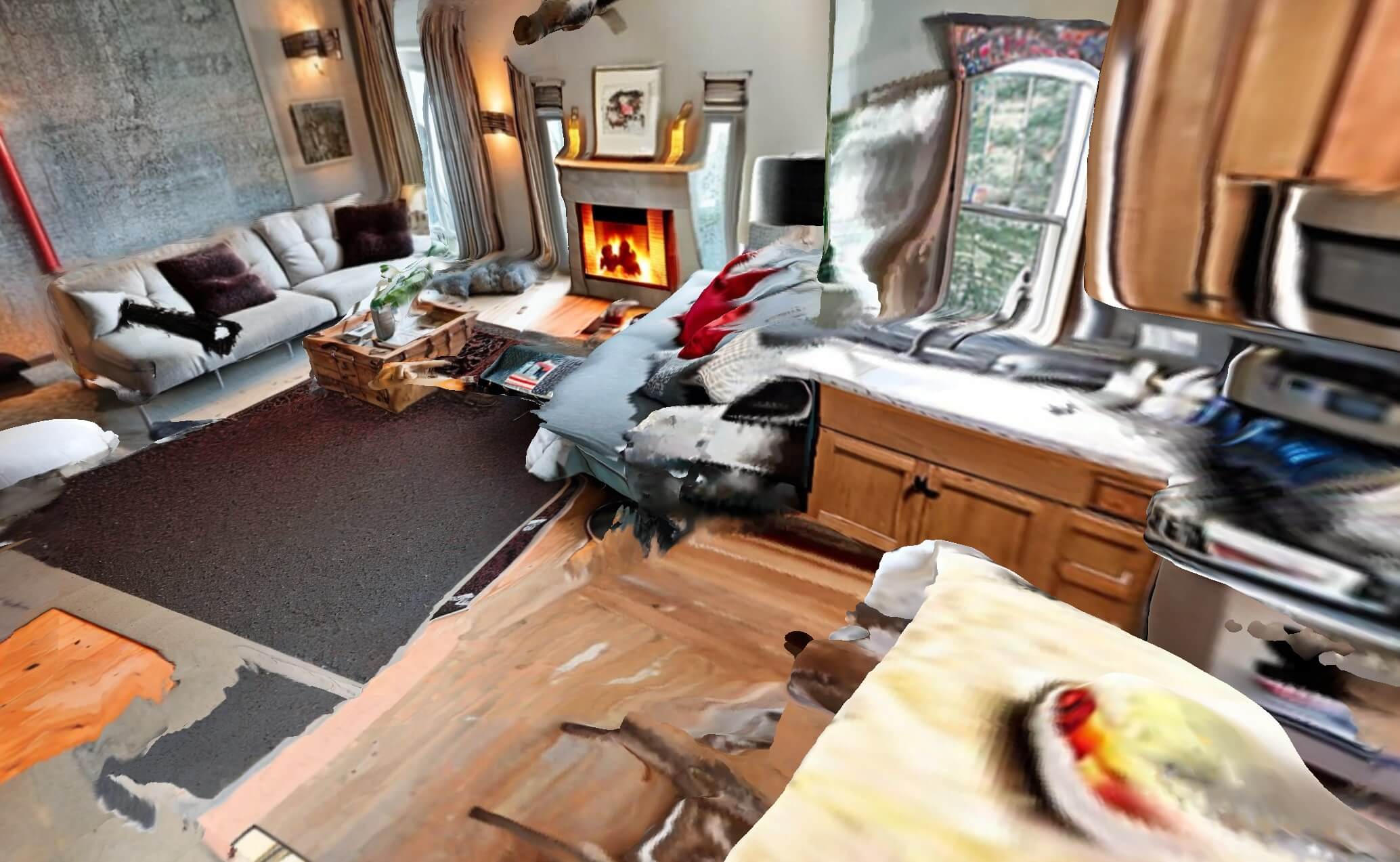} \\ 
\multicolumn{2}{c}{\textit{\textcolor{Mycolor1}{kitchen, dinner table, dishwashers, ovens, countertops}}} \\
\multicolumn{2}{c}{\textit{\textcolor{Mycolor2}{living room, lit furnace, couch, curtains}}} \\

\includegraphics[height=24mm]{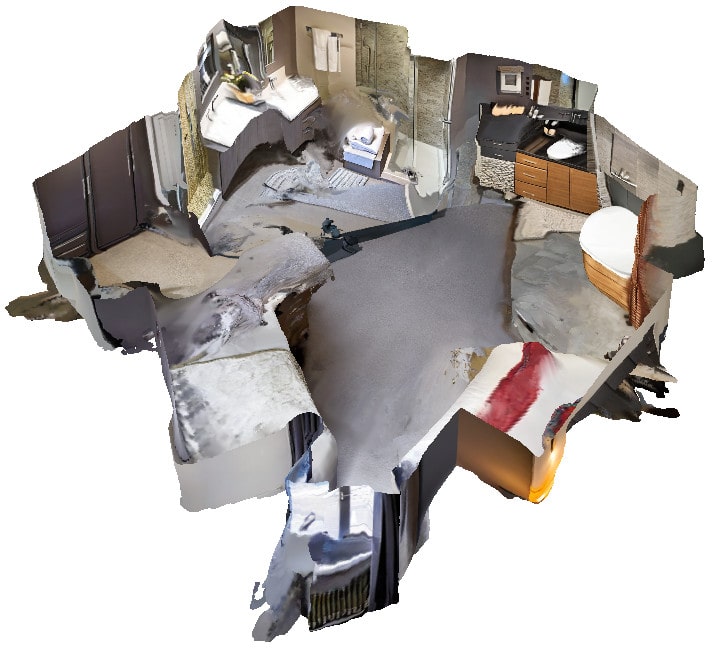} &
\includegraphics[height=24mm]{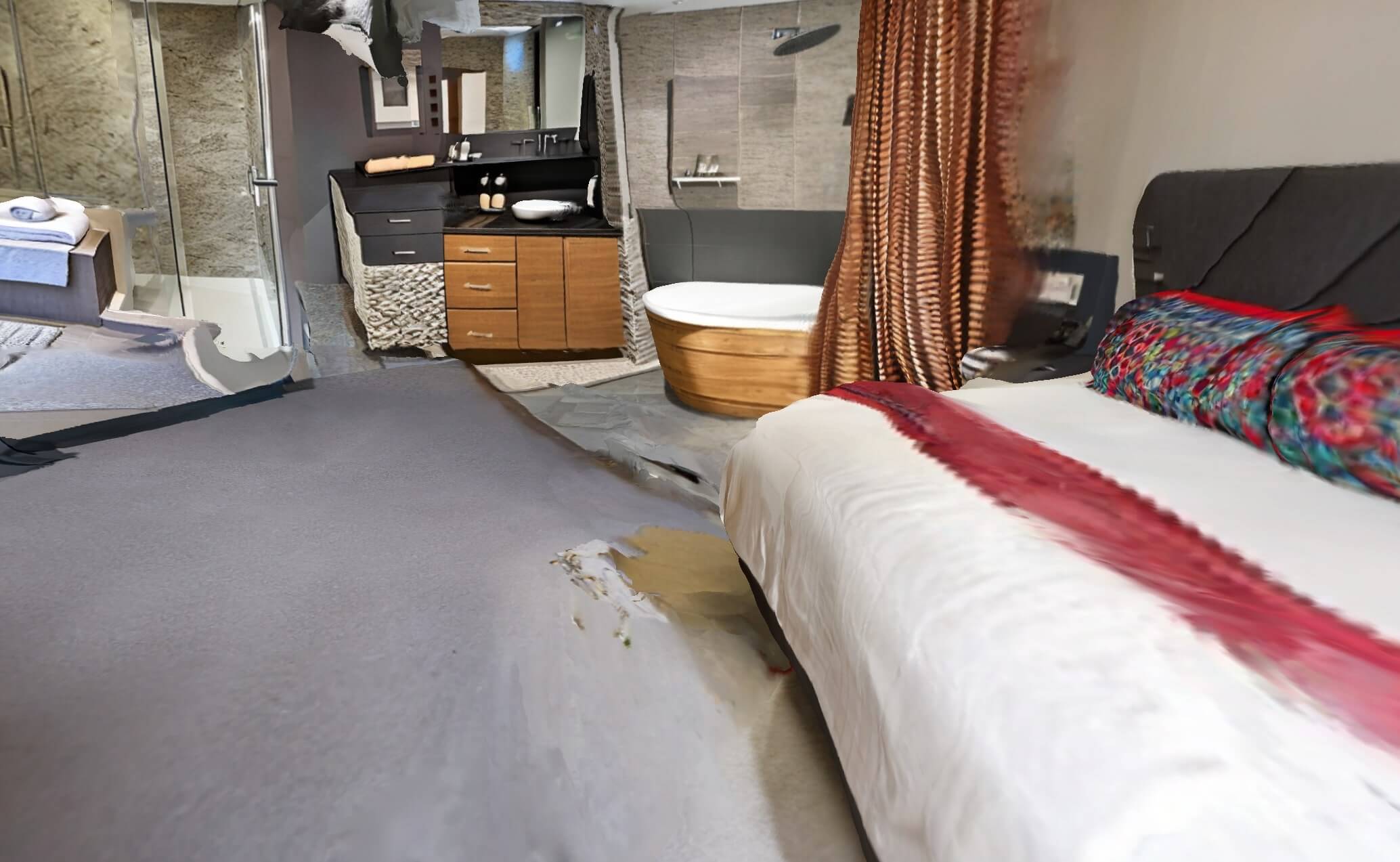} \\ 
\multicolumn{2}{c}{\textit{\textcolor{Mycolor1}{bathroom, shower, bathtub}}} \\
\multicolumn{2}{c}{\textit{\textcolor{Mycolor2}{bedroom, king-size bed, wardrobes}}} \\

\end{tabular}
\caption{
\textbf{Spatially varying scene generation.}
Our method can create rooms with multiple parts by prompt mixing.
We use separate prompts for cameras viewing different parts of the scene.
This is a controllable way to create rooms from multiple descriptions.
}
\label{fig:ours-mix-appl}
\end{figure}
\begin{figure}
    \centering
    \includegraphics[height=2.3cm]{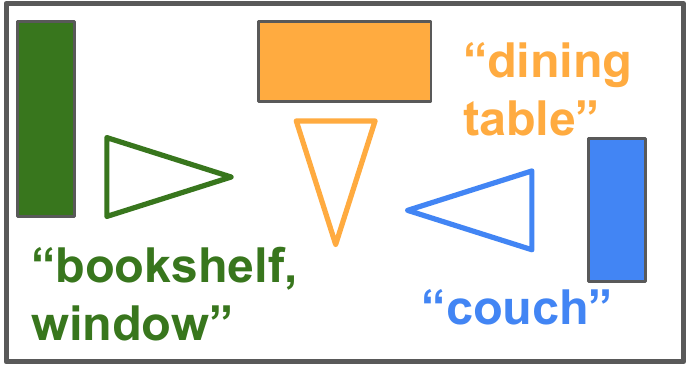}
    \includegraphics[height=2.3cm]{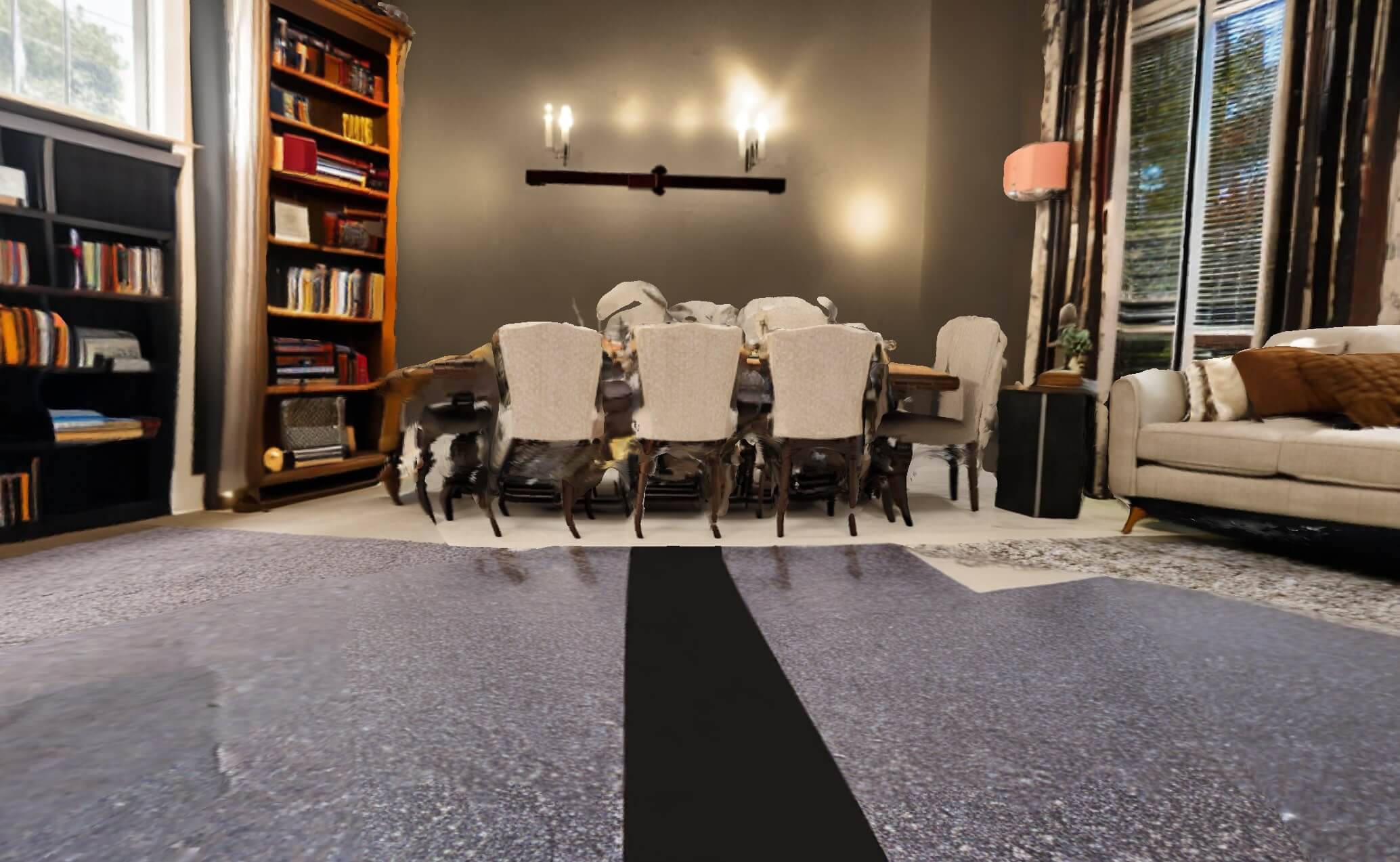}
    \caption{\textbf{Scene generation with layout guidance.}
    Our method can generate scenes from layout guidance. 
    Left: we describe objects with different prompts for cameras facing at different directions.
    Right: the generated part of the room.
    }
    \label{fig:layout-control}
\end{figure}

Our method can be applied to generate a scene as the combination of multiple text prompts.
Specifically, we use separate text prompts for different poses, crafting a set of trajectories that spatially combines scene descriptions.
This can be desired to avoid repeating elements in a complete scene (e.g., multiple couches could be spread out over the whole room when using the same prompt for every camera).
Instead, users can specify different object positions through different camera poses and text prompts.
It can also be used to design a house comprised of multiple rooms, each with a different type (e.g., a living room that leads to a kitchen).
We show results that combine multiple text prompts in Figure~\ref{fig:ours-mix-appl} and Figure~\ref{fig:layout-control}.

We note that the layout can only be partially controlled by the camera poses, since scene generation can create chunks with larger or smaller extent.
We believe this demonstrates an exciting application of our method, that can be further explored in future work.

\subsection{Limitations}

Our approach allows to generate 3D room geometry from arbitrary text prompts that are highly detailed and contain consistent geometry.
Nevertheless, our method can still fail under certain conditions (see supplemental material).
First, our thresholding scheme (see Section ~\ref{subsec:3D Mesh Generation}) may not detect all stretched-out regions, which may lead to remaining distortions.
Additionally, some holes may still not be completed fully after the second stage (see Section ~\ref{subsec:Trajectory Generation}), which results in over-smoothed regions after applying poisson reconstruction.
Our scene representation does not decompose material from lighting, which bakes in shadows or bright lamps, that are generated from the diffusion model.

\section{Conclusion}
We have shown a method to generate textured 3D meshes from only text input.
We use text-to-image 2D generators to create a sequence of images.
The core insight of our method is a tailored viewpoint selection, that allows to create a 3D mesh with seamless geometry and compelling textures.
Specifically, we lift the images into a 3D scene, by employing our alignment strategy that iteratively fuses all images into the mesh.
Our output meshes represent arbitrary indoor scenes that can be rendered with classical rasterization pipelines.
We believe our approach demonstrates an exciting application of large-scale 3D asset creation, that only requires text as input.

\section*{Acknowledgements}
This work was funded in part by Cisco Systems.
It was also supported by the ERC Starting Grant Scan2CAD (804724) as well as the German Research Foundation (DFG) Research Unit ``Learning and Simulation in Visual Computing.''
We also thank Angela Dai for the video voice over.

{\small
\bibliographystyle{ieee_fullname}
\bibliography{bib}
}
\clearpage
\appendix
\section{Supplemental Video}
Please watch our attached video~\footnote[1]{\url{https://youtu.be/fjRnFL91EZc}} for a comprehensive evaluation of the proposed method.
We include rendered videos of multiple generated scenes from novel trajectories, that showcase the quality of both generated texture and geometry (and also show the generated ceilings).
We also show an animation how the mesh is built up over time, that illustrates the usage of our two-stage pose sampling scheme (generation and completion).
We compare against baselines and ablations of our method by showing rendered videos.

\section{Societal Impact}

Our method leverages text-to-image models to generate a sequence of images from text, specifically we use the Stable Diffusion model~\cite{Rombach2021HighResolutionIS}.
Thus it inherits possible drawbacks of these 2D models.
First, our method could be exploited to generate harmful content, by forcing the text-to-image model to generate respective images.
Furthermore, our method is biased towards the cultural or stereotypical data distribution, that was used to train the text-to-image model.
Lastly, we note that text-to-image models are trained on large-scale text-image datasets~\cite{Schuhmann2022LAION5BAO}.
Thus, the model learns to reproduce and combine the style of artists, whose works are contained in these datasets.
This raises questions regarding the correct way to credit these artists or if it is ethical to benefit from their works in this way at all.

Our method can be used to generate meshes, that depict entire scenes, from only text as input.
This significantly reduces the required expertise to model and design such 3D assets.
Thus, we believe our work proposes a promising step towards the democratization of large-scale 3D content creation.

\section{Limitations}
\begin{figure*}
\centering
\setlength\tabcolsep{1pt}
\begin{tabular}{ccc}
\includegraphics[width=0.33\textwidth]{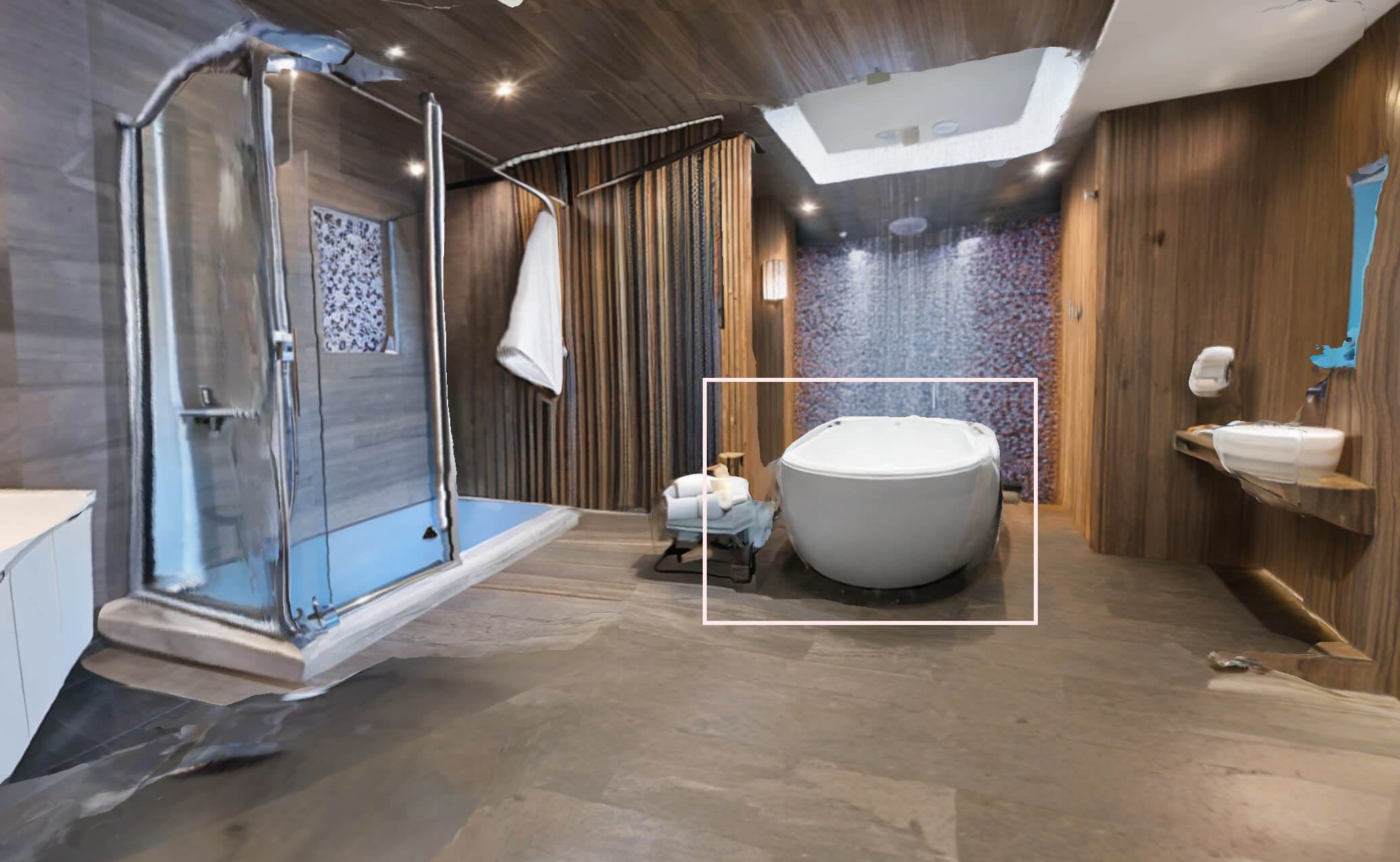} &
\includegraphics[width=0.33\textwidth]{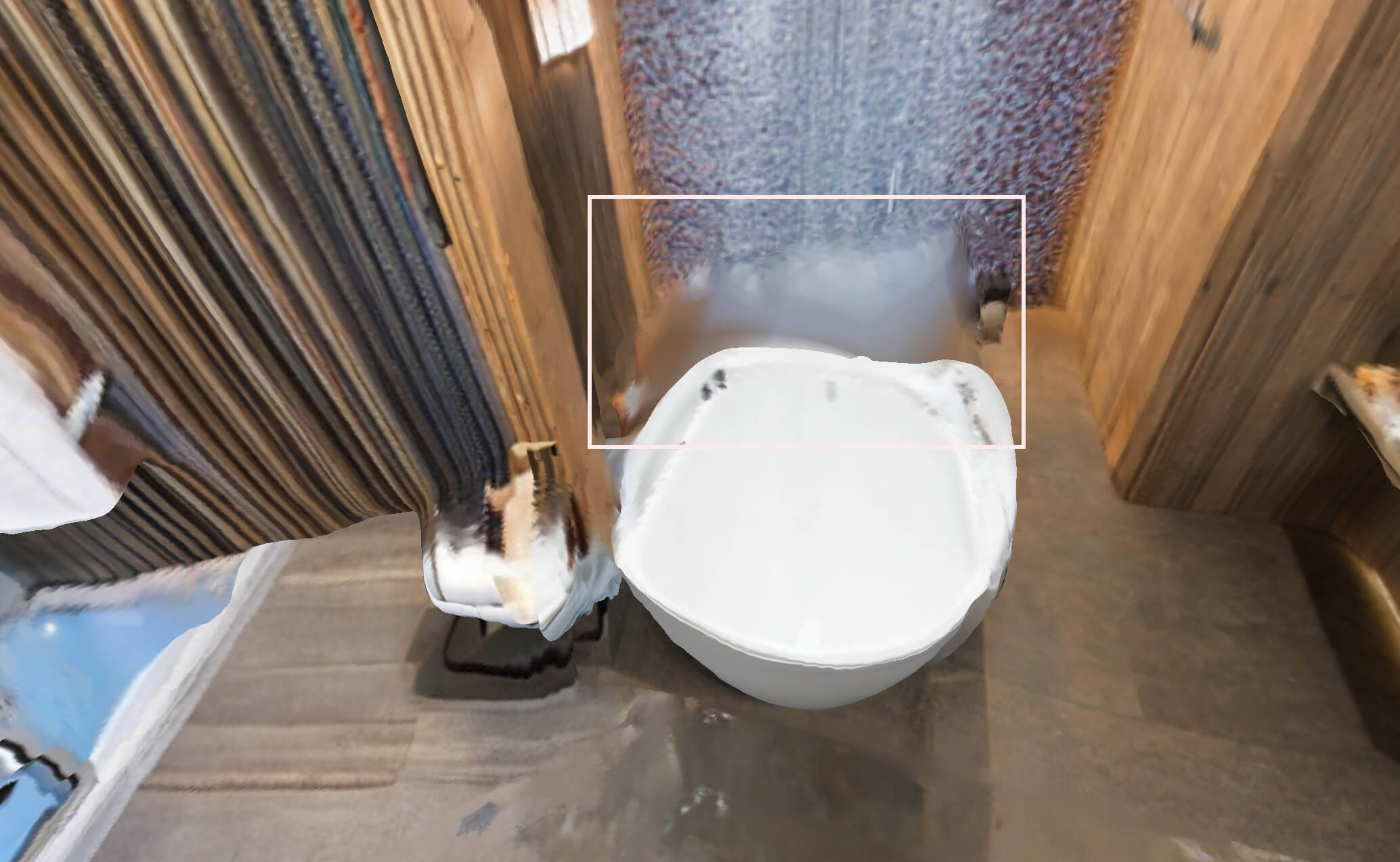} &
\includegraphics[width=0.33\textwidth]{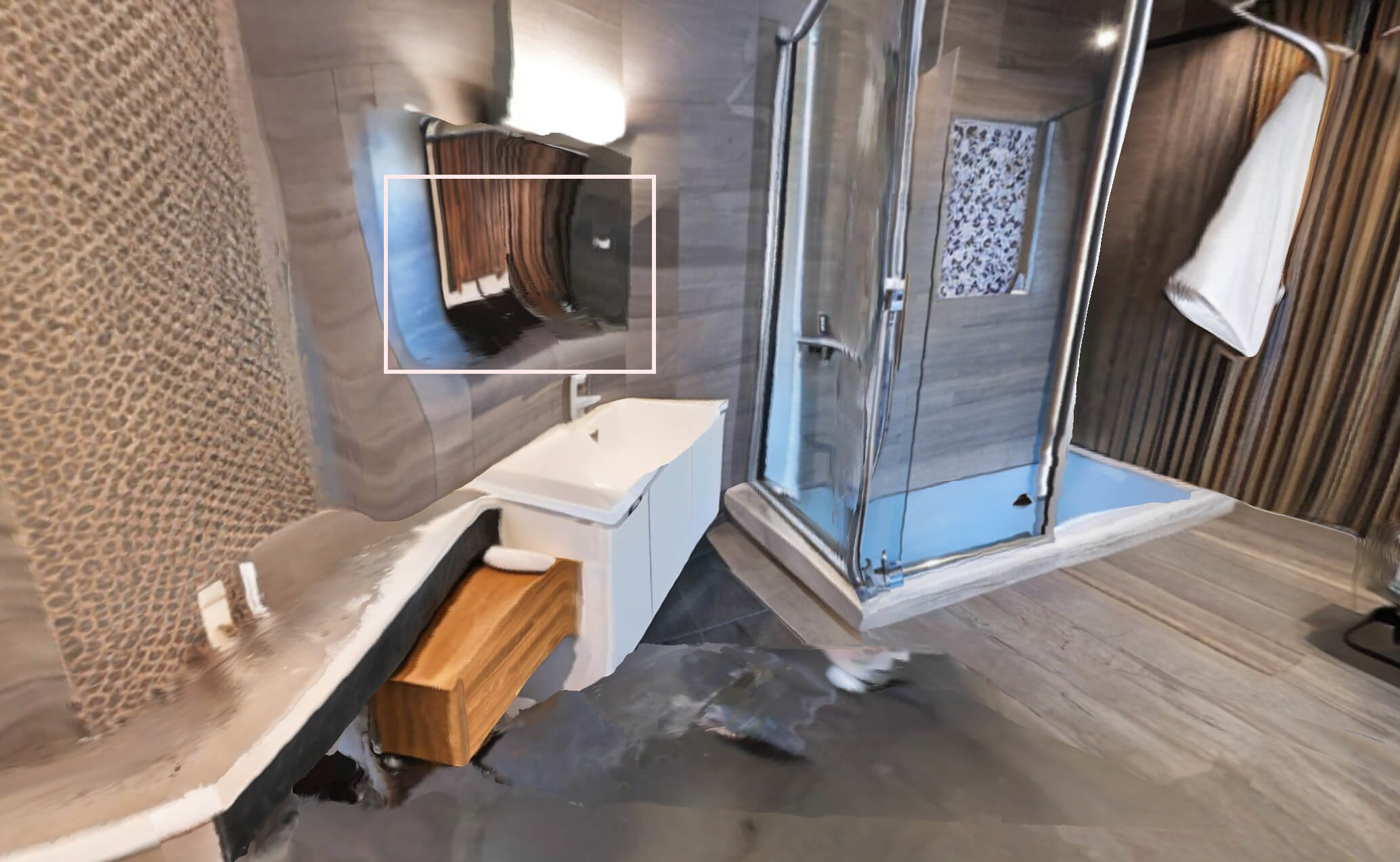} \\
(a) created scene & (b) overly smoothed geometry & (c) stretched geometry
\end{tabular}
\caption{
\textbf{Limitations of our method.}
(a) Our approach creates scenes with compelling textures and complete structure like walls, floor and ceiling.
(b) Our completion stage
(see Section~3.4)
might not be able to inpaint all holes, if no suitable camera pose could be sampled (e.g. small areas behind an object that are close to a wall). The hole is still closed through Poisson reconstruction~\cite{kazhdan2006poisson}, but the geometry may become smoothed.
(c) Our fusion stage
(see Section~3.3)
might not remove all stretched-out faces, because we use fixed thresholds.
}
\label{fig:supp-limitation}
\end{figure*}

Given a text prompt, our approach allows to generate 3D room geometry that is highly detailed and contains consistent 3D geometry.
Nevertheless, our method can still fail under certain conditions (see Figure~\ref{fig:supp-limitation}).

First, our completion stage
(see Section~3.4)
might not be able to inpaint all holes (Figure~\ref{fig:supp-limitation}b).
For example this can happen, if an object contains holes that are close to a wall.
These angles are hard to see from additional cameras and thus might remain untouched.
We still close these holes by applying Poisson surface reconstruction~\cite{kazhdan2006poisson}.
However, this can results in overly smoothed geometry.

Second, our mesh fusion stage
(see Section~3.3)
might not remove all stretched-out faces.
Faces can appear stretched-out because of imperfect depth estimation and alignment.
Over time this can yield unusual room shapes such as the curved wall in Figure~\ref{fig:supp-limitation}c.
We apply two filtering schemes to remove stretched-out faces before fusing them with the existing geometry.
Both use thresholds $\delta_{sn}{=}0.1, \delta_{edge}{=}0.1$, that we fix during all our experiments.
It can happen that some faces are not removed by the filtering schemes, but are still stretched-out unnaturally.
However, we find that lowering the thresholds would also remove unstretched geometry.
This would make creating a complete scene harder, because more holes need to be inpainted in the completion stage.

\section{Details on User Study}
\begin{figure}
    \centering
    \includegraphics[width=0.5\textwidth]{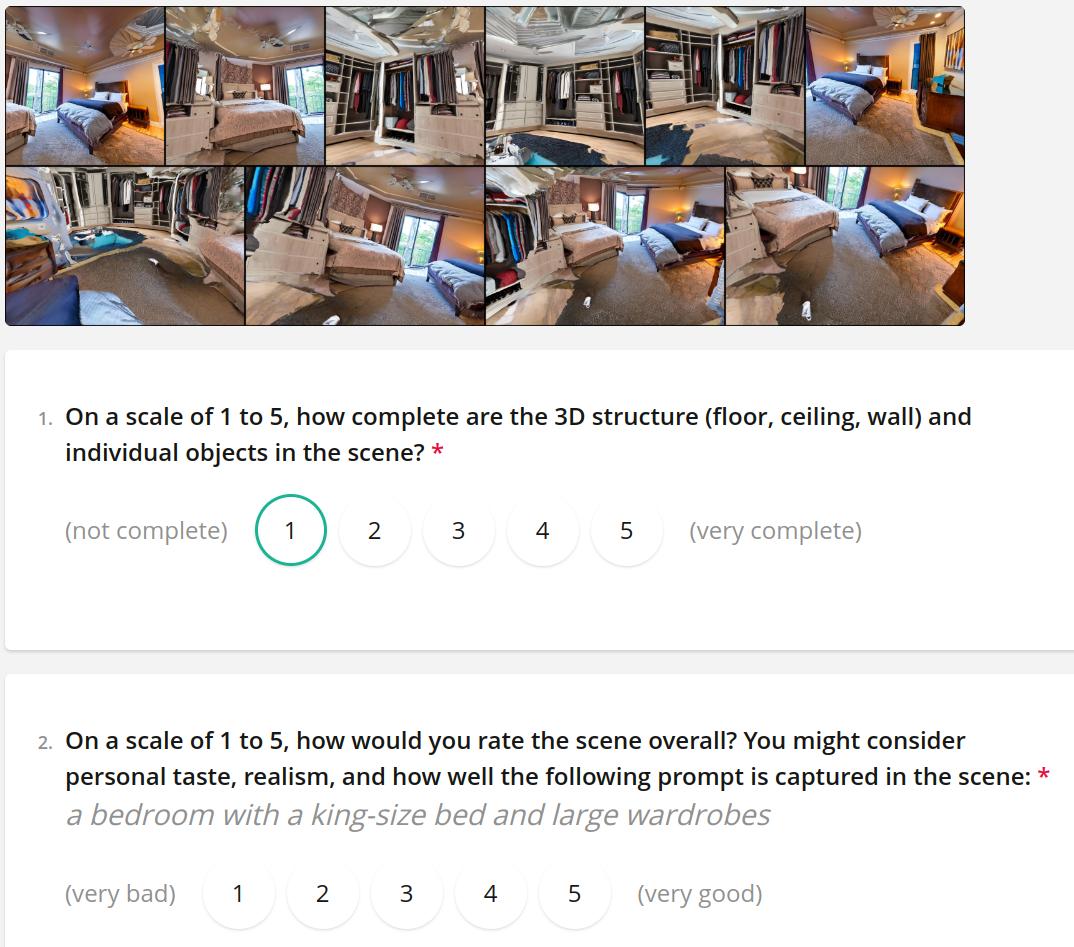}
    \caption{
    \textbf{User study interface.} (Top) We present users with multiple images from each scene, that show it from multiple angles. (Bottom) We ask users to rate the scene on a scale from $1{-}5$ by asking them about the 3D structure completeness (question 1) and the overall perceptual quality (question 2).}
    \label{fig:sup_user_study}
\end{figure}

We conduct a user study and ask $n{=}61$ users to score Perceptual Quality (\emph{PQ}) and 3D Structure Completeness (\emph{3DS}) of the whole scene on a scale of $1{-}5$.
We show an example of how we asked the users to score these two metrics in Figure~\ref{fig:sup_user_study}.
We present users with multiple images from each scene, that show it from multiple angles.
Then we ask them to rate the scene on a scale from $1{-}5$ by asking them about the 3D structure completeness and the overall perceptual quality.
In total, we received $1098$ datapoints from multiple scenes and report averaged results per method.

\section{Additional Implementation Details}
We give additional implementation details in the following subsections.

\subsection{Importance of Predefined Trajectories}
\begin{figure*}
\centering
\setlength\tabcolsep{1pt}
\begin{tabular}{cccc}
\includegraphics[width=0.24\textwidth]{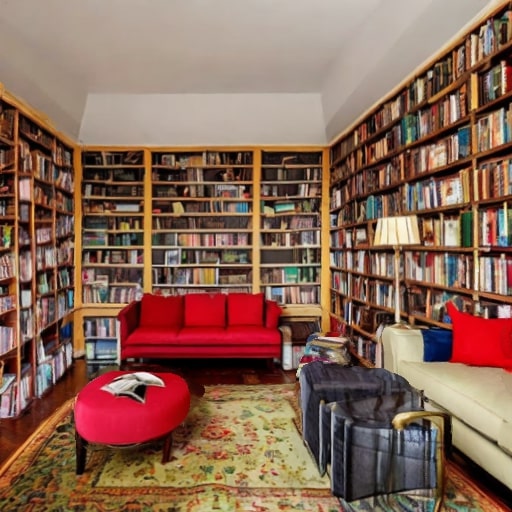} &
\includegraphics[width=0.24\textwidth]{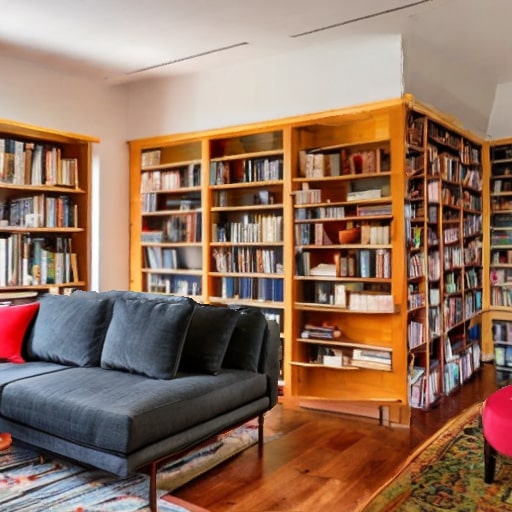} &
\includegraphics[width=0.24\textwidth]{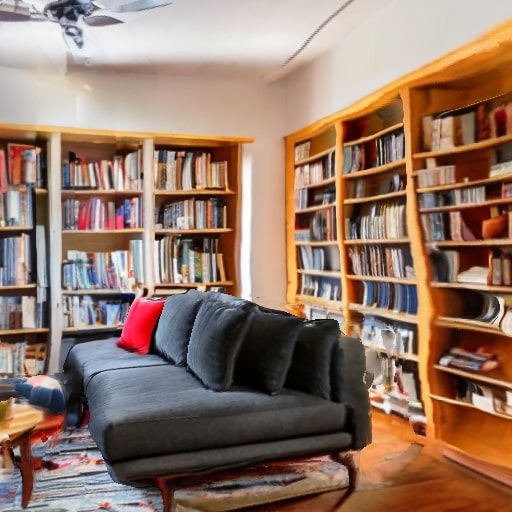} &
\includegraphics[width=0.24\textwidth]{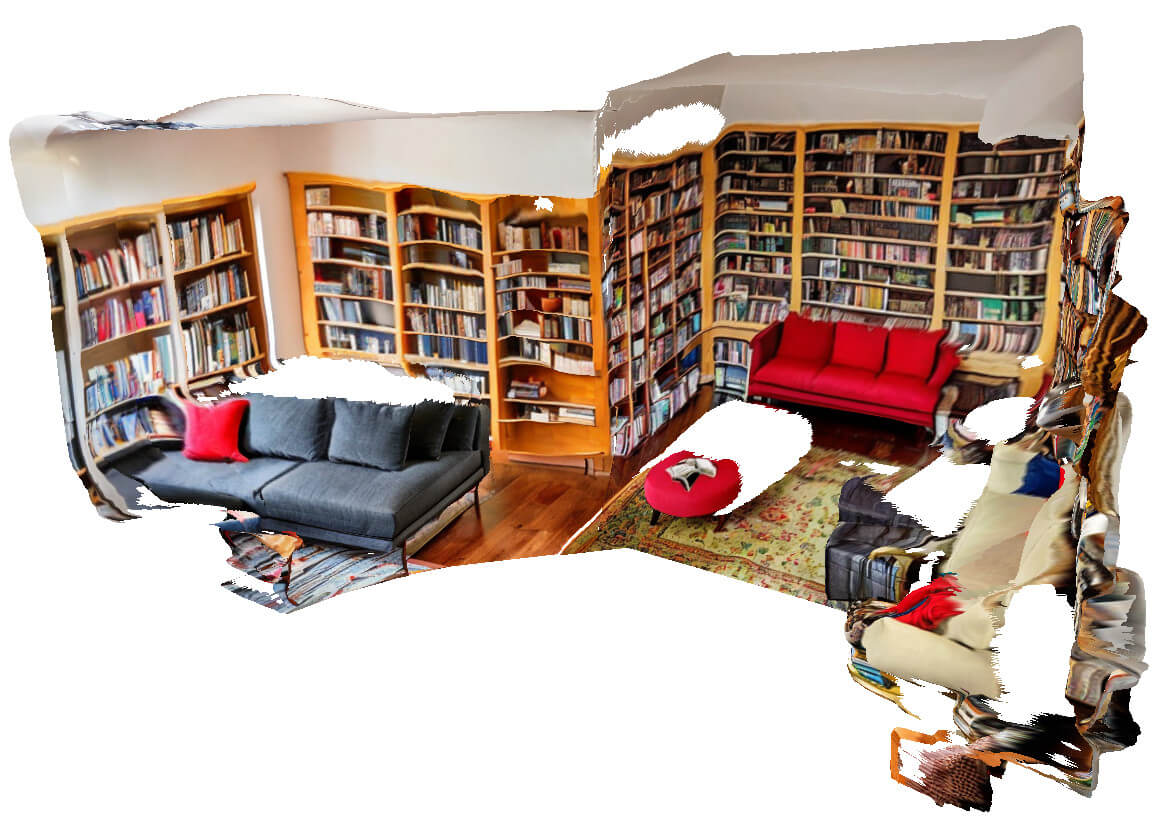} \\
(a) start image & (b) ours: next chunk & (c) ours: refine chunk & (d) ours: after trajectory \\
\includegraphics[width=0.24\textwidth]{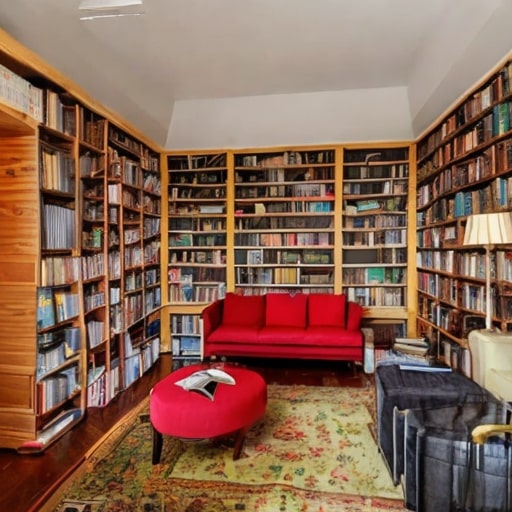} &
\includegraphics[width=0.24\textwidth]{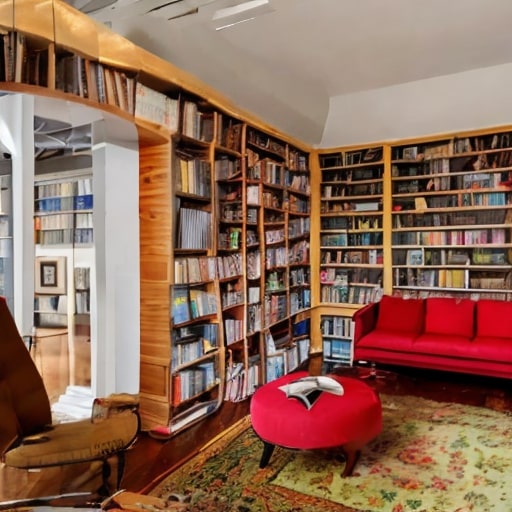} &
\includegraphics[width=0.24\textwidth]{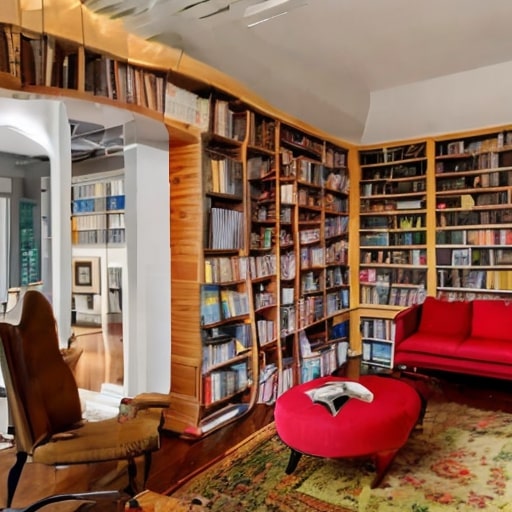} &
\includegraphics[width=0.24\textwidth]{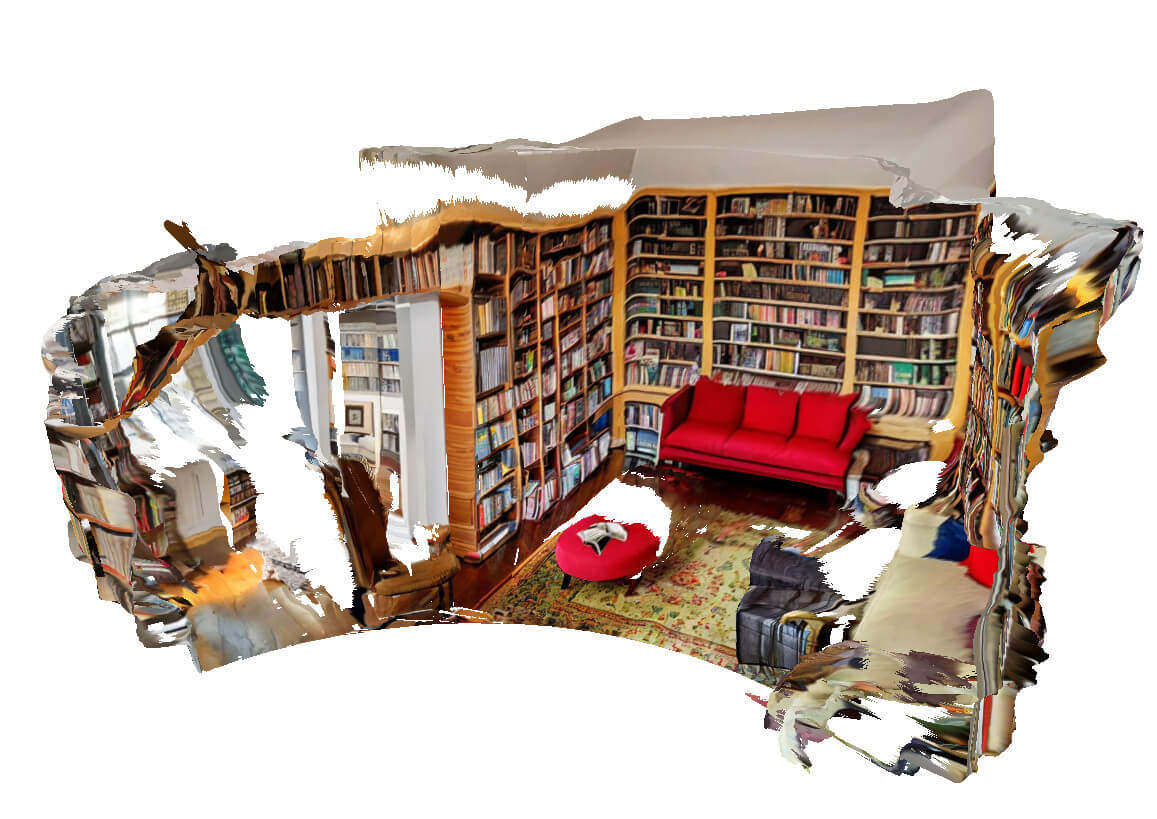} \\
(e) small change I & (f) small change II & (g) small change III & (h) small change: after trajectory \\
\end{tabular}
\caption{
\textbf{Importance of predefined trajectories.}
We sample predefined trajectories in the first stage of our tailored two-stage viewpoint selection scheme
(see Section~3.4).
First, we create the outline of the next scene chunk (b).
Then, we sample additional poses that translate and rotate into the new scene chunk to complete its 3D structure (c).
This results in a 3D consistent next mesh patch, that we fuse with existing content (d).
In contrast, results degenerate (h), if we sample sub-optimal poses (e.g. small viewpoint changes in e-g).
}
\label{fig:supp-predefined_traj}
\end{figure*}

We create the complete scene layout and furniture in the first stage of our tailored two-stage viewpoint selection scheme
(see Section~3.4).
To this end, we sample multiple \emph{predefined} trajectories from which we iteratively generate the scene.
We fix the trajectories for our main results, as we found it already creates rooms with a variety of different layouts.
Users can modify them according to our guidelines as demonstrated in
Section~4.4 in the main paper.
Each trajectory consists of a start pose and an end pose and we linearly interpolate between both.
We found generation works best, if each trajectory starts off from a viewpoint with mostly unobserved regions.
This gives the text-to-image model enough freedom to create novel content with reasonable global structure.

Thus, we construct each trajectory with the following principle.
First, we select a start pose that views mostly unobserved content and generate the outline of the next scene chunk from it (Figure~\ref{fig:supp-predefined_traj}b).
Then, we subsequently translate and rotate into the chunk to refine its 3D structure until the end of the trajectory (Figure~\ref{fig:supp-predefined_traj}c).
This creates mesh patches with convincing 3D structure (Figure~\ref{fig:supp-predefined_traj}d).
In contrast, if we design trajectories that do not follow this principle, results can degenerate.
For example, if the viewpoint change is small, the text-to-image model creates novel content only for small portions of the image (Figure~\ref{fig:supp-predefined_traj}e-g).
Thus, locally the generated content looks reasonable, but it accumulates into inconsistent global structure (Figure~\ref{fig:supp-predefined_traj}h).

\subsection{Effect of Depth Smoothing in Alignment}
\begin{figure*}
\centering
\setlength\tabcolsep{1pt}
\begin{tabular}{cccc}
\includegraphics[width=0.24\textwidth]{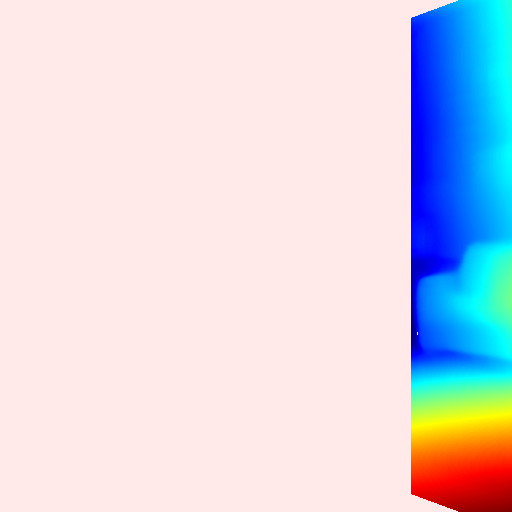} &
\includegraphics[width=0.24\textwidth]{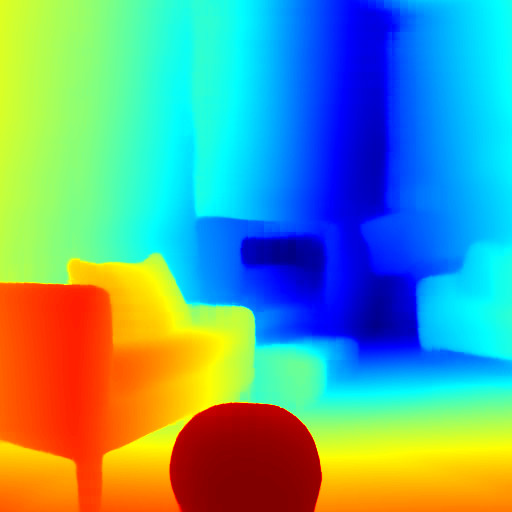} &
\includegraphics[width=0.24\textwidth]{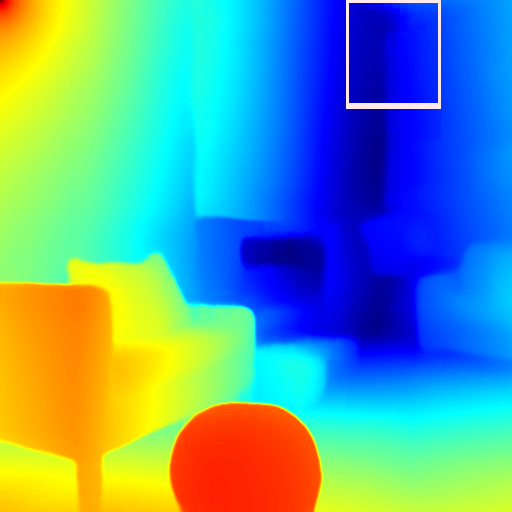} &
\includegraphics[width=0.24\textwidth]{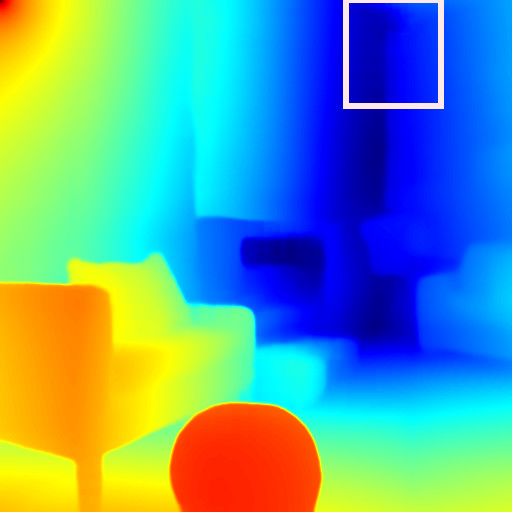} \\
(a) rendered depth & (b) inpainted depth & (c) aligned depth & (d) aligned + smoothed depth \\
\includegraphics[width=0.24\textwidth]{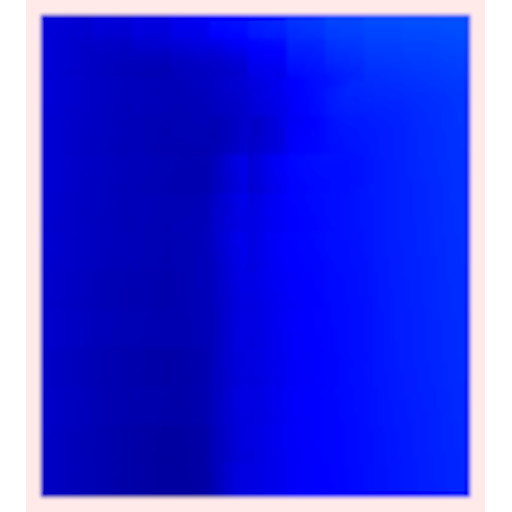} &
\includegraphics[width=0.24\textwidth]{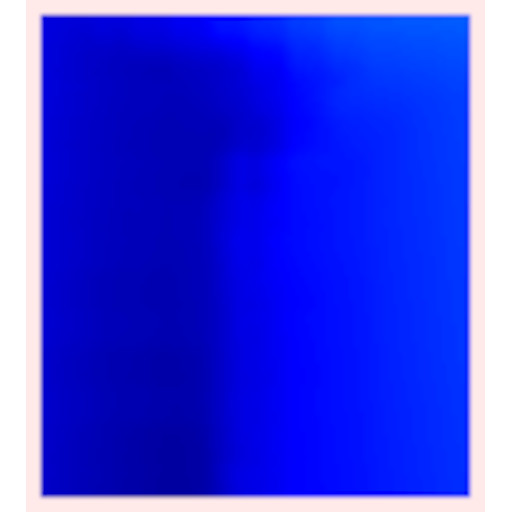} &
\includegraphics[width=0.24\textwidth]{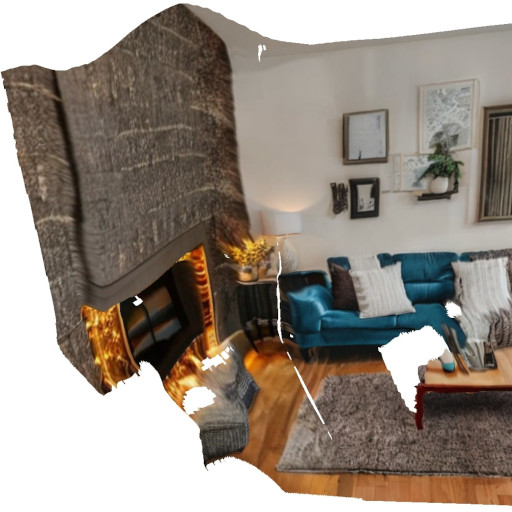} &
\includegraphics[width=0.24\textwidth]{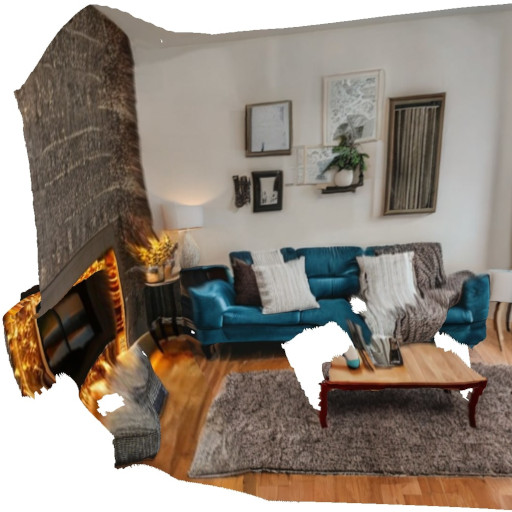} \\
(e) zoom-in of (c) & (f) zoom-in of (d) & (g) fused mesh from (b) & (h) fused mesh from (d) \\
\end{tabular}
\caption{
\textbf{Details on the depth alignment step.}
For each novel pose, we predict the depth for the newly generated image content
(see Section~3.2).
First we inpaint the depth using a monocular depth prediction network (b).
Then, we align inpainted depth (b) and rendered depth (a) in the least squares sense to obtain an aligned depth (c).
Finally, we smooth the result to remove remaining sharp borders between old and new content (d).
This results in smoother, less blocky depth (e and f).
Our depth alignment is necessary to create transitions without holes between mesh patches (g and h).
}
\label{fig:supp-depth-smoothing}
\end{figure*}

For each camera pose in both stages, we follow an iterative scene generation scheme
(see Section~3.1).
After generating novel content, we predict its depth in our depth alignment stage
(see Section~3.2).
First, we predict the depth using a monocular depth inpainting network (Figure~\ref{fig:supp-depth-smoothing}b).
However, directly using this depth for mesh fusion results in unaligned mesh patches (Figure~\ref{fig:supp-depth-smoothing}g).
Thus, we improve the result by aligning rendered depth and inpainted depth in the least squares sense (Figure~\ref{fig:supp-depth-smoothing}c).
Finally, we smooth the aligned depth by applying a $5\times5$ gaussian blur kernel at the image edges between rendered and predicted depth (Figure~\ref{fig:supp-depth-smoothing}d).
This smoothens out remaining discontinuity artifacts between old and new content (Figure~\ref{fig:supp-depth-smoothing}e and f).
In practice, we found this can further reduce sharp borders between objects, leading to overall better alignment (Figure~\ref{fig:supp-depth-smoothing}h).

\subsection{Importance of Mask Dilation in Completion}
\begin{figure*}
\centering
\setlength\tabcolsep{1pt}
\begin{tabular}{ccccc}
\includegraphics[width=0.19\textwidth]{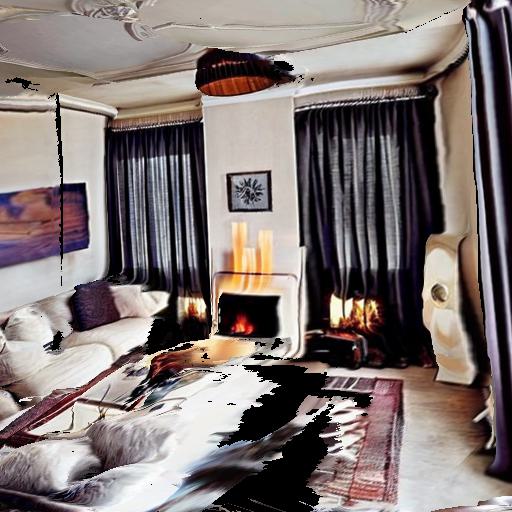} &
\includegraphics[width=0.19\textwidth]{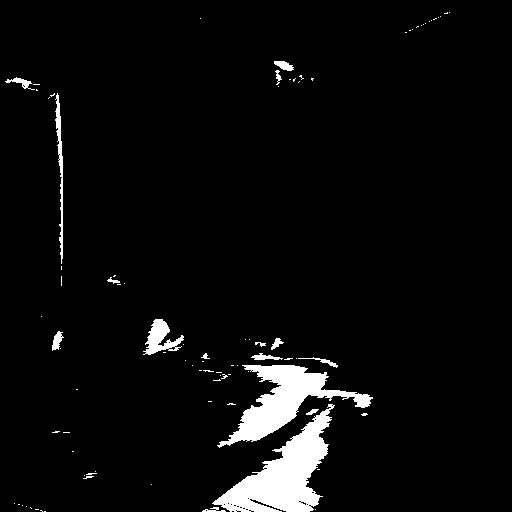} &
\includegraphics[width=0.19\textwidth]{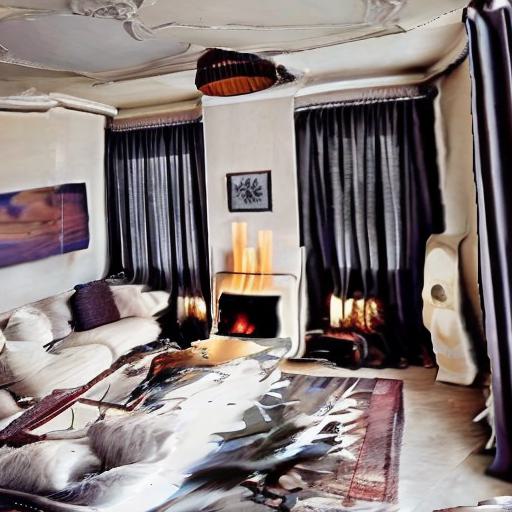} &
\includegraphics[width=0.19\textwidth]{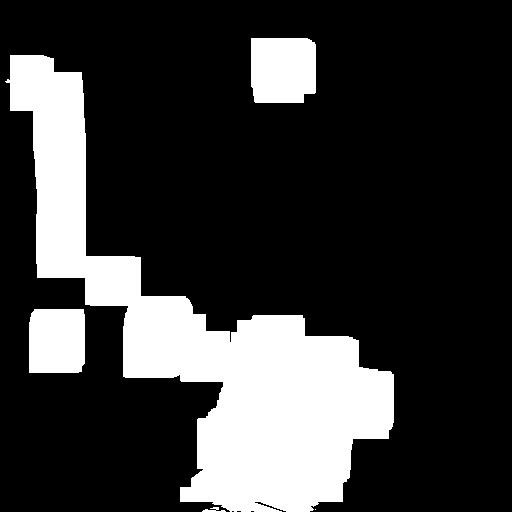} &
\includegraphics[width=0.19\textwidth]{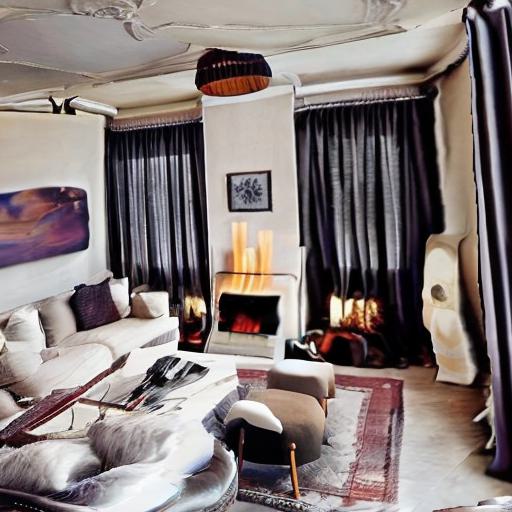} \\
(a) rendered image & (b) rendered mask & (c) inpaint na\"ive & (d) dilated mask & (e) inpaint dilated
\end{tabular}
\caption{
\textbf{Importance of mask dilation during completion.}
In our second stage, we complete the scene mesh by filling in unobserved regions
(see Section~3.4).
First, we sample camera poses that view such unobserved regions (a).
The unobserved regions can have arbitrary size (b).
Directly inpainting only the masked regions from (b) gives distorted results, because the holes can be too small for reasonable inpainting results (c).
Instead, we inpaint small holes with a classical inpainting method~\cite{telea2004image} and dilate remaining holes to a larger size (d).
The resulting image after inpainting contains more reasonable structure (e).
}
\label{fig:supp-mask-dilation}
\end{figure*}

We complete the scene in the second stage of our tailored two-stage viewpoint selection scheme, by filling in remaining holes in the mesh
(see Section~3.4).
To this end, we first select suitable camera poses that look at these holes (Figure~\ref{fig:supp-mask-dilation}a).
We then follow the iterative scene generation scheme to fill in the holes in the mesh
(see Section~3.1).
The holes can have arbitrarily small or large sizes, depending on how the scene layout was generated in the first stage of our method (Figure~\ref{fig:supp-mask-dilation}b).
Similarly to Fridman~\etal~\cite{fridman2023scenescape}, we found that directly inpainting such holes can lead to sub-optimal results (Figure~\ref{fig:supp-mask-dilation}c).
This is because the text-to-image model needs to inpaint small regions and the direct neighborhood of the holes can be distorted.
To alleviate this issue, we inpaint small holes with a classical inpainting algorithm~\cite{telea2004image}.
We classify small holes by applying a morphological erosion operation with a $3\times3$ kernel on the inpainting mask.
Next, we increase the size of remaining holes, by repeating a morphological dilation operation with a $7\times7$ kernel on the eroded inpainting mask for five times (Figure~\ref{fig:supp-mask-dilation}d).
Finally, we inpaint the image using the dilated mask (Figure~\ref{fig:supp-mask-dilation}e).
This yields more convincing results because the text-to-image model can inpaint larger areas and create more meaningful global structure.
To combine the new content with the existing mesh, we apply our triangulation scheme
(see Section~3.3).
Additionally, we remove all faces that fall into the dilated region and are close to the rendered screen-space depth (since they are replaced by the novel content).

\section{Additional Discussion on Related Methods and Baselines}
\begin{figure}[!h]
\centering
\setlength\tabcolsep{1pt}
\begin{tabular}{ccc}
\includegraphics[height=2.1cm]{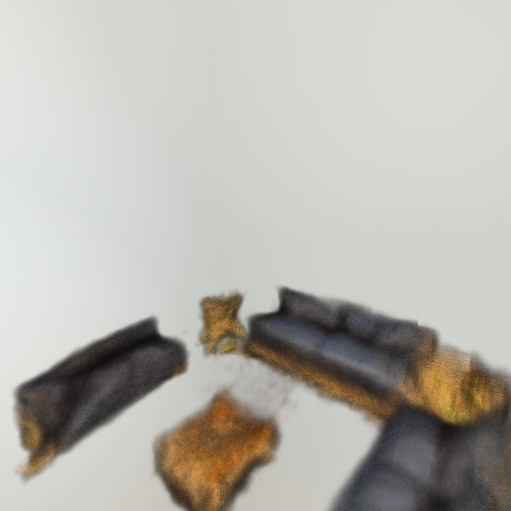} &
\includegraphics[height=2.1cm]{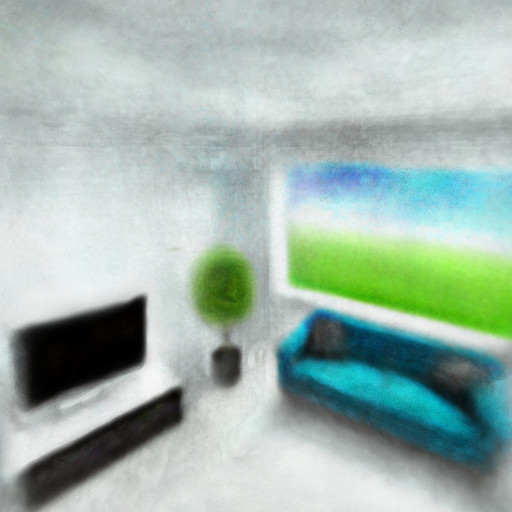} &
\includegraphics[height=2.1cm]{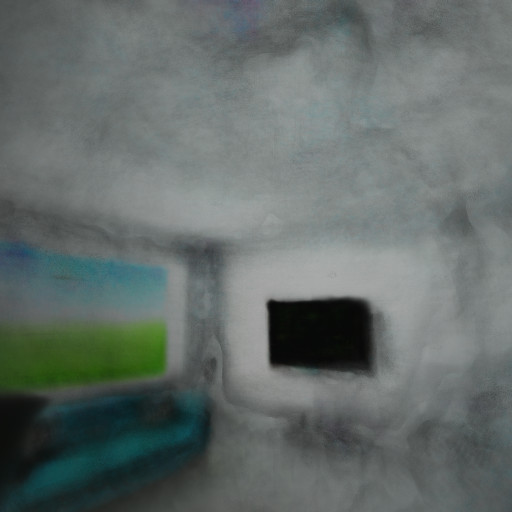}
\end{tabular}
\caption{Left: DreamFusion-Inward. Mid/Right: DreamFusion-Outward from in- and out-of-distribution viewpoints.}
\vspace{-0.30cm}
\label{fig:sup_fig_dreamfusion}
\end{figure}
\begin{figure*}
\centering
\setlength\tabcolsep{1pt}
\begin{tabular}{ccc}
\includegraphics[width=0.33\textwidth]{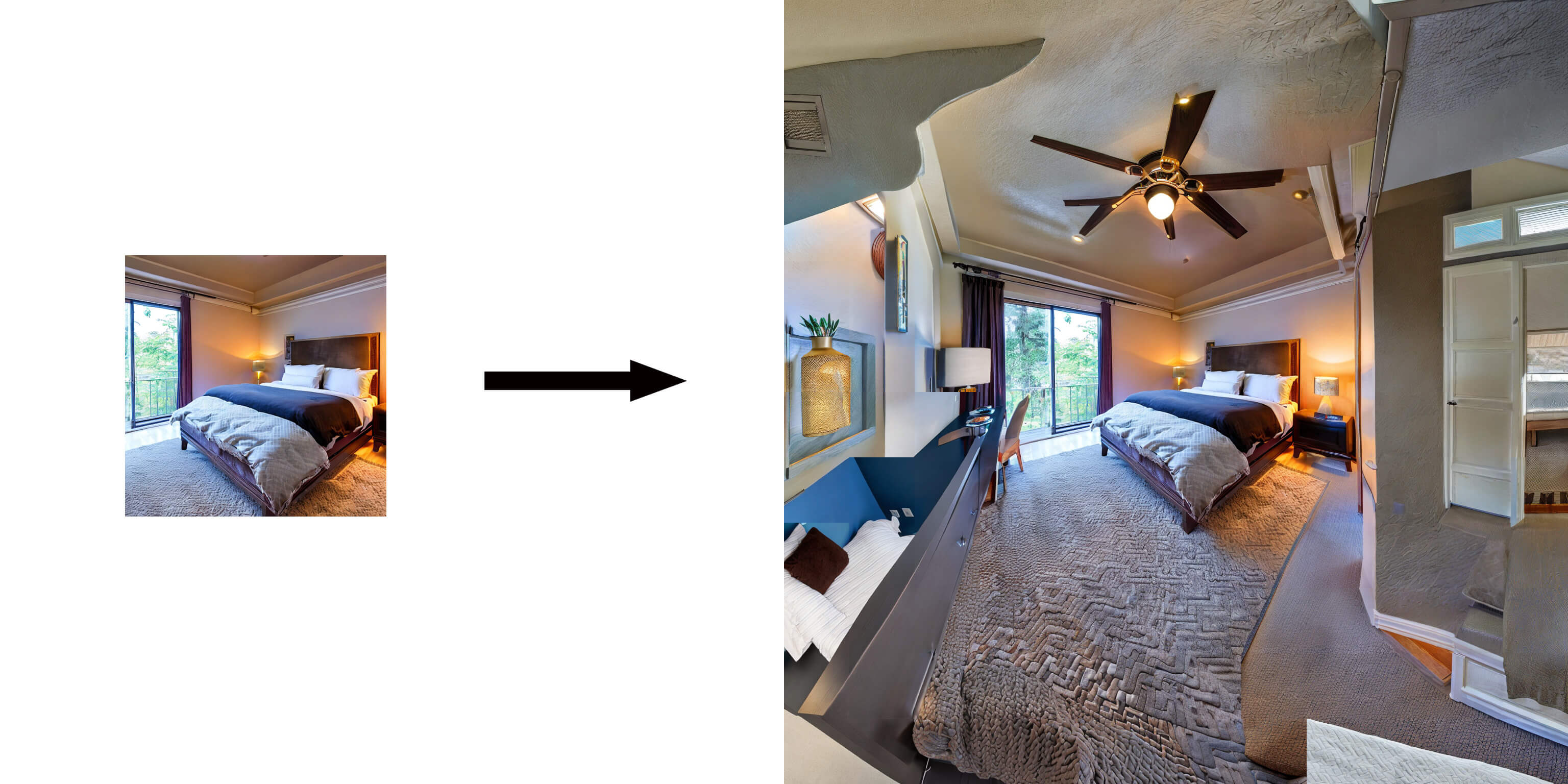} &
\includegraphics[width=0.33\textwidth]{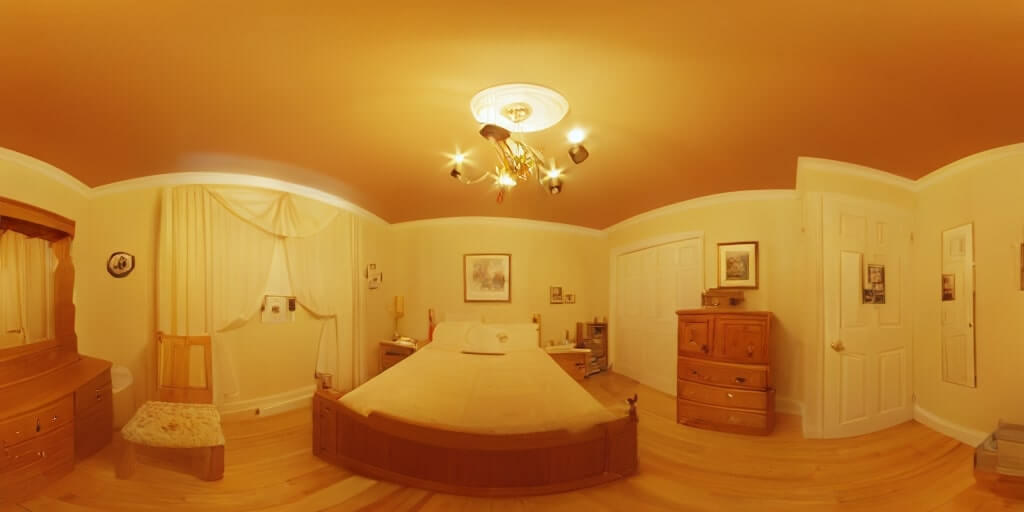} &
\includegraphics[width=0.33\textwidth]{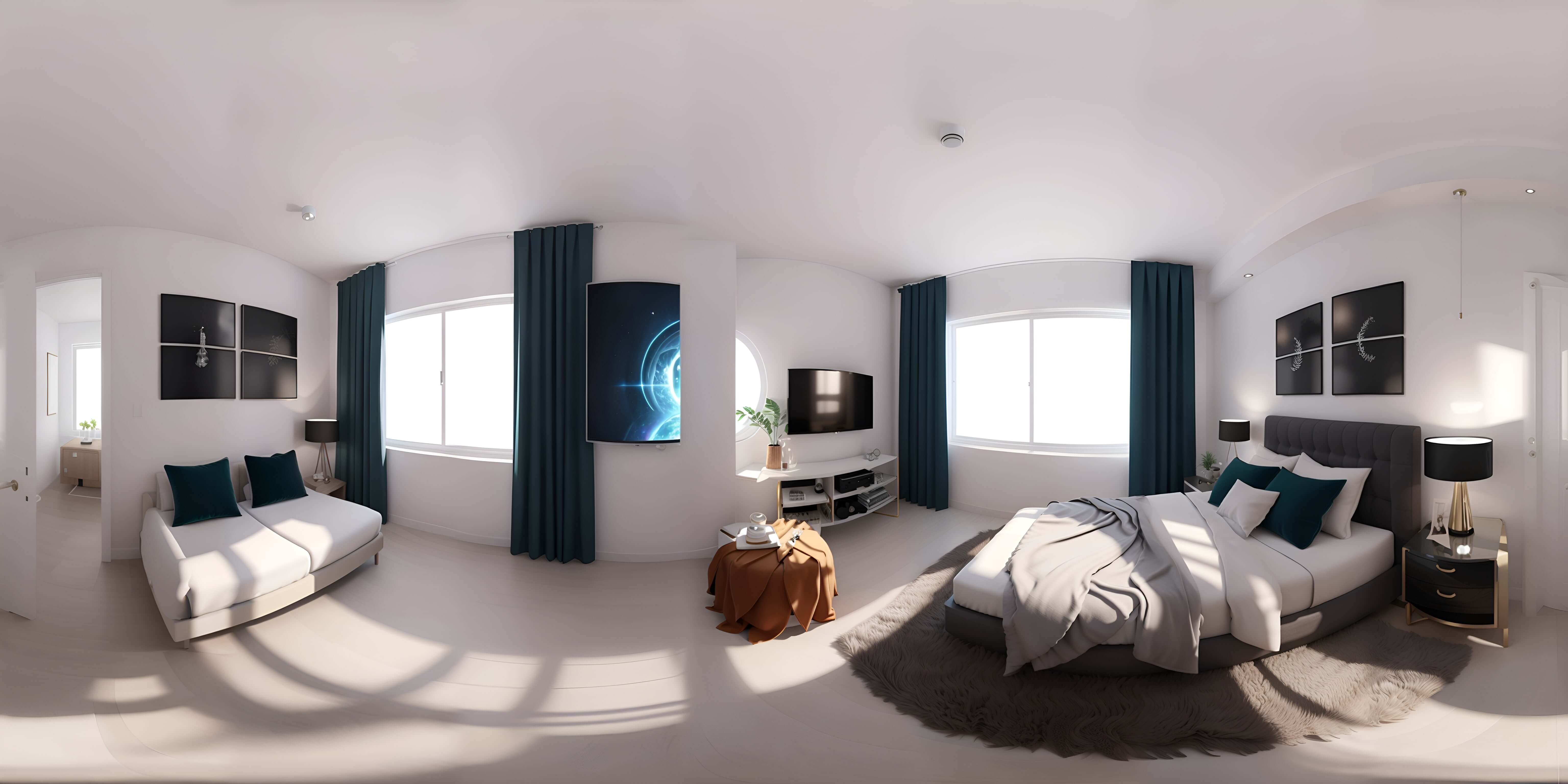} \\
\multicolumn{3}{c}{\textit{a bedroom with a king-size bed and a large wardrobe}} \\
\includegraphics[width=0.33\textwidth]{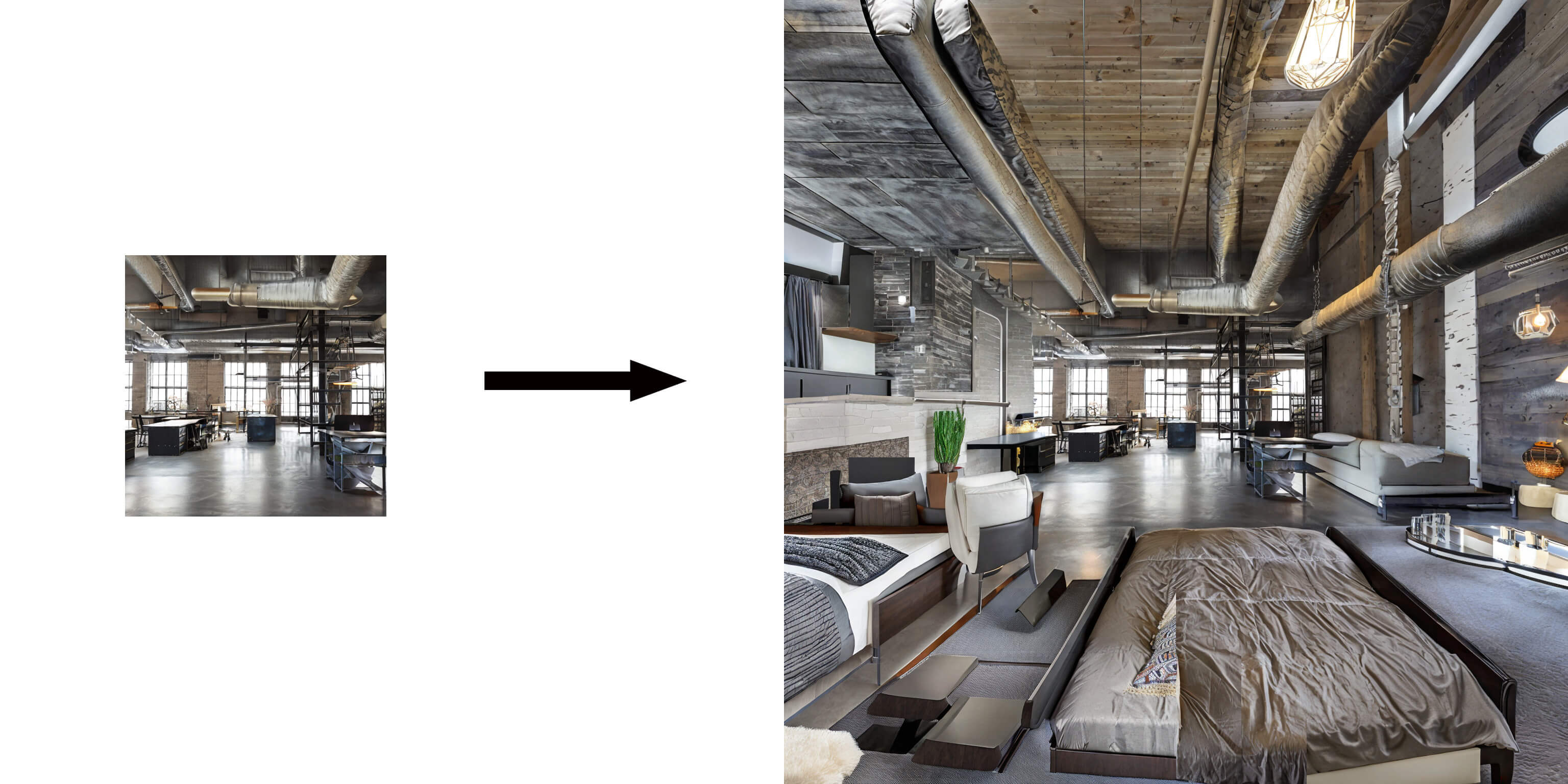} &
\includegraphics[width=0.33\textwidth]{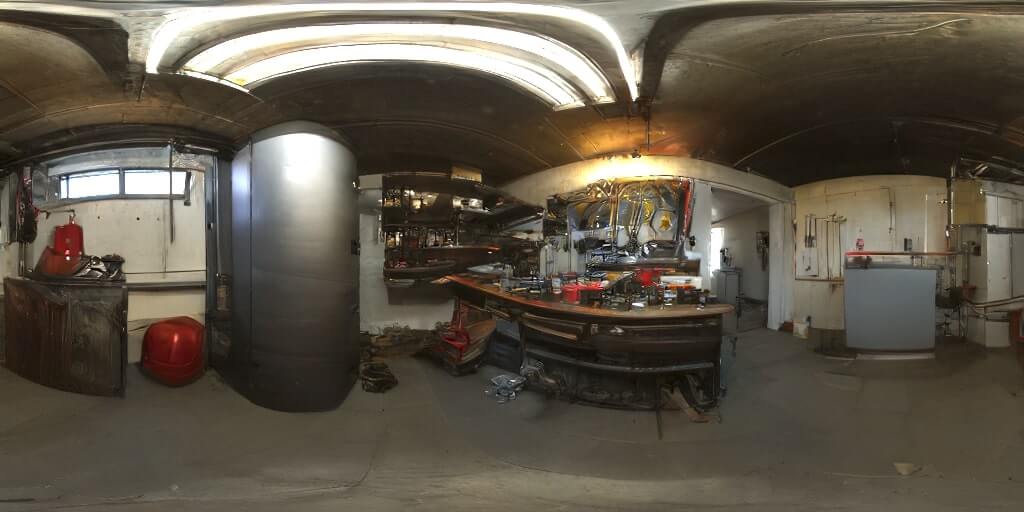} &
\includegraphics[width=0.33\textwidth]{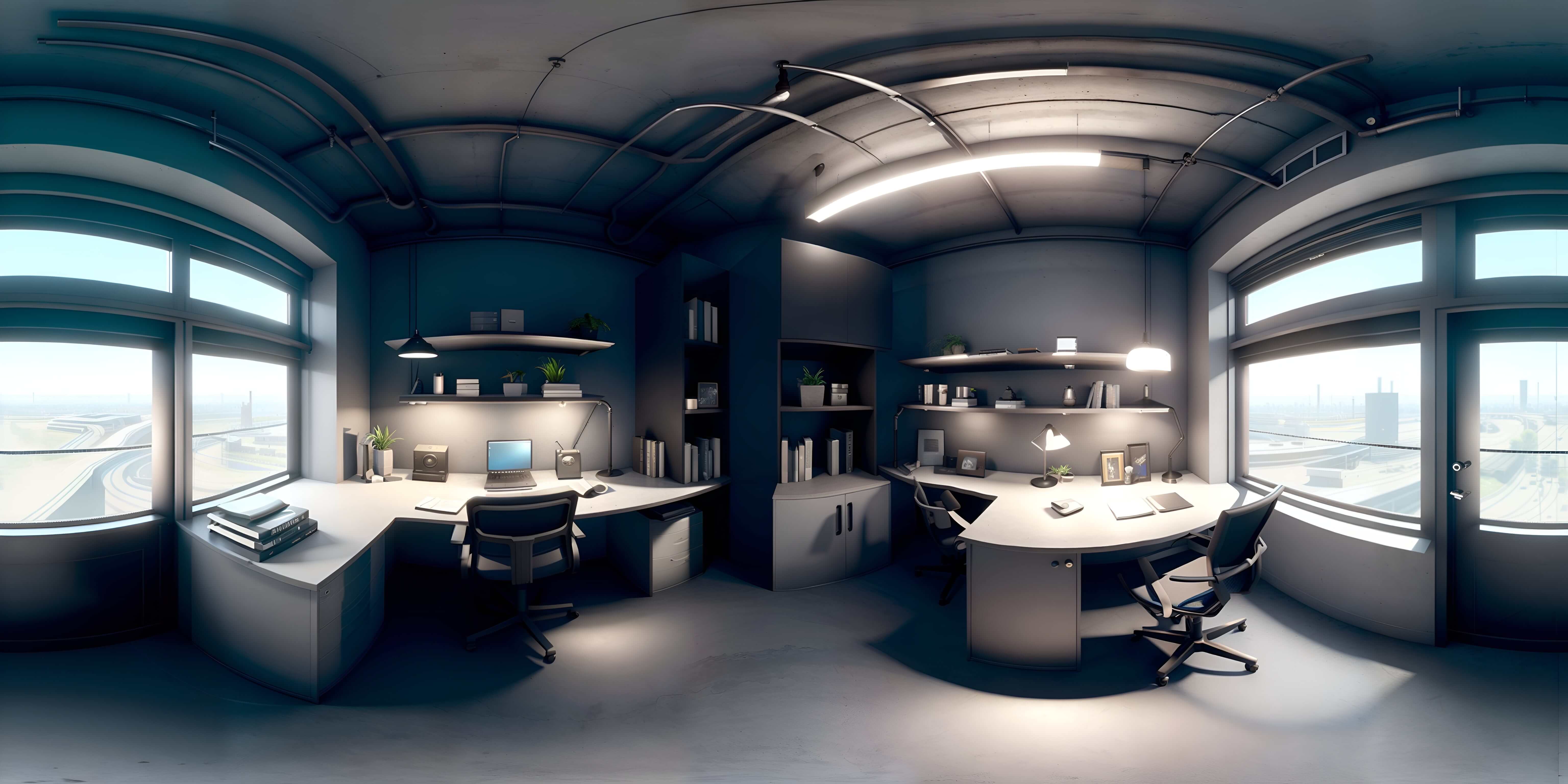} \\
\multicolumn{3}{c}{\textit{Editorial Style Photo, Industrial Home Office, Steel Shelves, Concrete, Metal, Edison Bulbs, Exposed Ductwork}} \\
(a) Outpainting~\cite{Ramesh2022HierarchicalTI, OpenAI2022} & (b) Text2Light~\cite{chen2022text2light} & (c) Blockade~\cite{blockade}
\end{tabular}
\caption{
\textbf{Intermediate results from baselines.}
We first produce these intermediate results, before unprojecting them into a 3D mesh.
(a) Outpainting~\cite{Ramesh2022HierarchicalTI, OpenAI2022} generates an enlarged scene from a single viewpoint.
(b) Text2Light~\cite{chen2022text2light} creates a panoramic image of a scene.
(c) Blockade~\cite{blockade} creates a panoramic image of a scene.
}
\label{fig:supp-itermediate-baseline}
\end{figure*}

To the best of our knowledge, there are no direct baselines that generate textured 3D room geometry from text.
We compare against four related methods, that do not require supervision from 3D datasets.
In the following we give additional discussion on related methods and our selected baselines.

\rwpar{PureClipNeRF}~\cite{lee2022understanding}: We compare against text-to-3D methods for generating objects~\cite{poole2022dreamfusion, lin2022magic3d, jain2021dreamfields, lee2022understanding, wang2022score} and choose Lee~\etal~\cite{lee2022understanding} as open-source representative.
A common pattern in these text-to-3D methods is to sample inward-facing poses on a hemisphere, from which the object is iteratively optimized.
While the method of Lee~\etal~\cite{lee2022understanding} does not use a diffusion model to create high-fidelity images, it still uses the same pose sampling pattern.
This allows us to compare against these methods in general, by analyzing how well this pose sampling pattern can produce complete 3D scenes with structural elements like walls or floors.
We also run DreamFusion~\cite{poole2022dreamfusion} from the third-party implementation of Guo~\etal~\cite{threestudio2023}, see Figure~\ref{fig:sup_fig_dreamfusion}.
Similar to PureClipNeRF, object-centric cameras yield incomplete rooms.
Outward-facing cameras yield blurry $360^{\circ}$ surroundings, showing floaters when rendered out-of-distribution.

\rwpar{Outpainting}~\cite{Ramesh2022HierarchicalTI, OpenAI2022}: We compare against image outpainting. 
We combine outpainting from a Stable Diffusion~\cite{Rombach2021HighResolutionIS} model with depth estimation and triangulation to create a mesh from an enlarged viewpoint.
Starting off from a single generated image, we can synthesize novel content around it to create a complete scene in a single image plane (Figure~\ref{fig:supp-itermediate-baseline}a).
After creating the image, we then perform depth estimation and triangulation to lift the image into a 3D mesh.

\rwpar{Text2Light}~\cite{chen2022text2light}: We generate RGB panoramas from text using Chen~\etal~\cite{chen2022text2light}. 
We show example outputs in Figure~\ref{fig:supp-itermediate-baseline}b.
One can create immersive experiences by rendering a panorama onto a sphere, allowing to view the scene from arbitrary $360^\circ$ viewpoints.
However, it is not possible to simulate a true 3D environment directly (e.g., translating or rotating around objects), because the panorama only captures a single viewpoint.
Thus, related approaches estimate room layout~\cite{xu2021layout}, perform view synthesis ~\cite{kulkarni2022360fusionnerf, hsu2021moving, hara2022enhancement, huang2022360roam} or predict $360^\circ$ depth~\cite{area2022360monodepth, jin2020geometric} from one or multiple panoramas.
To compare to our method, we reconstruct the 3D mesh structure that can be obtained from a single panoramic image.
To this end, we perform depth prediction and subsequently apply our mesh fusion step.

\rwpar{Blockade}~\cite{blockade}: We compare against \emph{Blockade}~\cite{blockade}, which uses a text-to-image diffusion model to produce expressive RGB panoramas. We then extract the mesh similarly.

\rwpar{GAUDI}~\cite{Bautista2022GAUDIAN}: Bautista and Guo~\etal~\cite{Bautista2022GAUDIAN} present a method to generate large-scale 3D scenes encoded into a NeRF~\cite{Mildenhall2020NeRFRS} representation.
Their generative model can be conditioned to produce 3D indoor scenes from text as input.
In general, each scene allows for a different distribution of camera poses.
Walls and objects are placed at different positions in each scene, thus it depends on the scene to determine valid camera poses.
They model this joint latent distribution of scenes and cameras.
This allows to synthesize scenes that can be rendered from corresponding camera trajectories (e.g., a scene is rendered in a forward motion).
However, it requires training supervision from 3D datasets that contain ground-truth camera trajectories.
This restricts the method to the domain of a specific dataset of (synthetic, low-resolution) 3D scenes, which is limited in size and diversity.

In contrast, we choose another approach to represent the joint distribution of scenes and camera trajectories.
Our two-stage tailored viewpoint selection
(see Section~3.4)
first creates the general scene layout and furniture from predefined trajectories.
We choose these trajectories such that the camera poses do not intersect with generated geometry
(see Section~3.4 for more details).
Then we inpaint remaining holes by sampling additional poses.
This allows us to generate complete scenes with varying layouts.
Our resulting mesh can be rendered from arbitrary viewpoints, i.e., it is not bound to the specific trajectory used during generation.
Furthermore, our method can directly lift the generated images of a 2D text-to-image model into 3D, without requiring supervised training from 3D datasets.
This allows us to generate meshes, that can represent a much larger and more diverse set of indoor scenes with higher visual quality.

\section{Additional Qualitative Results}
\begin{figure*}
\centering
\setlength\tabcolsep{1pt}
\begin{tabular}{ccc}

\multirow{2}{*}[0.8in]{\includegraphics[height=48mm]{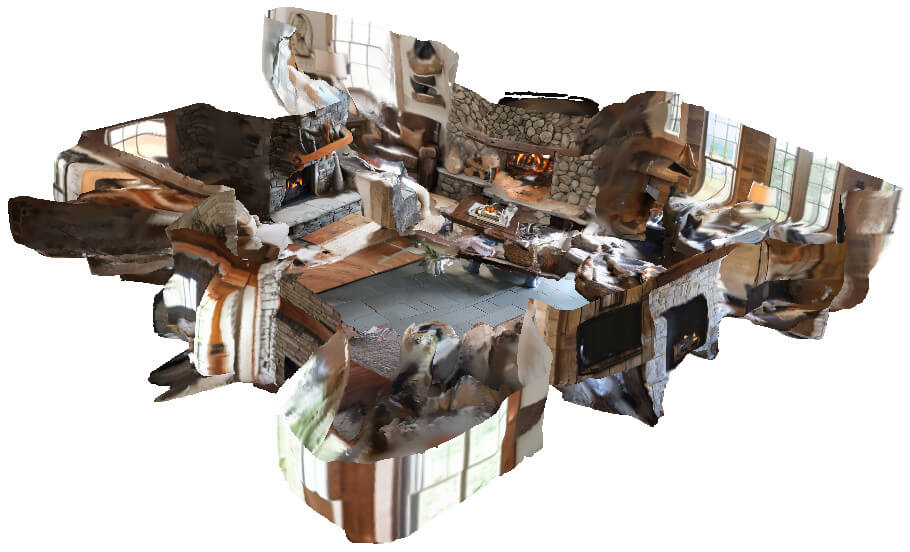}} & 
\includegraphics[height=24mm]{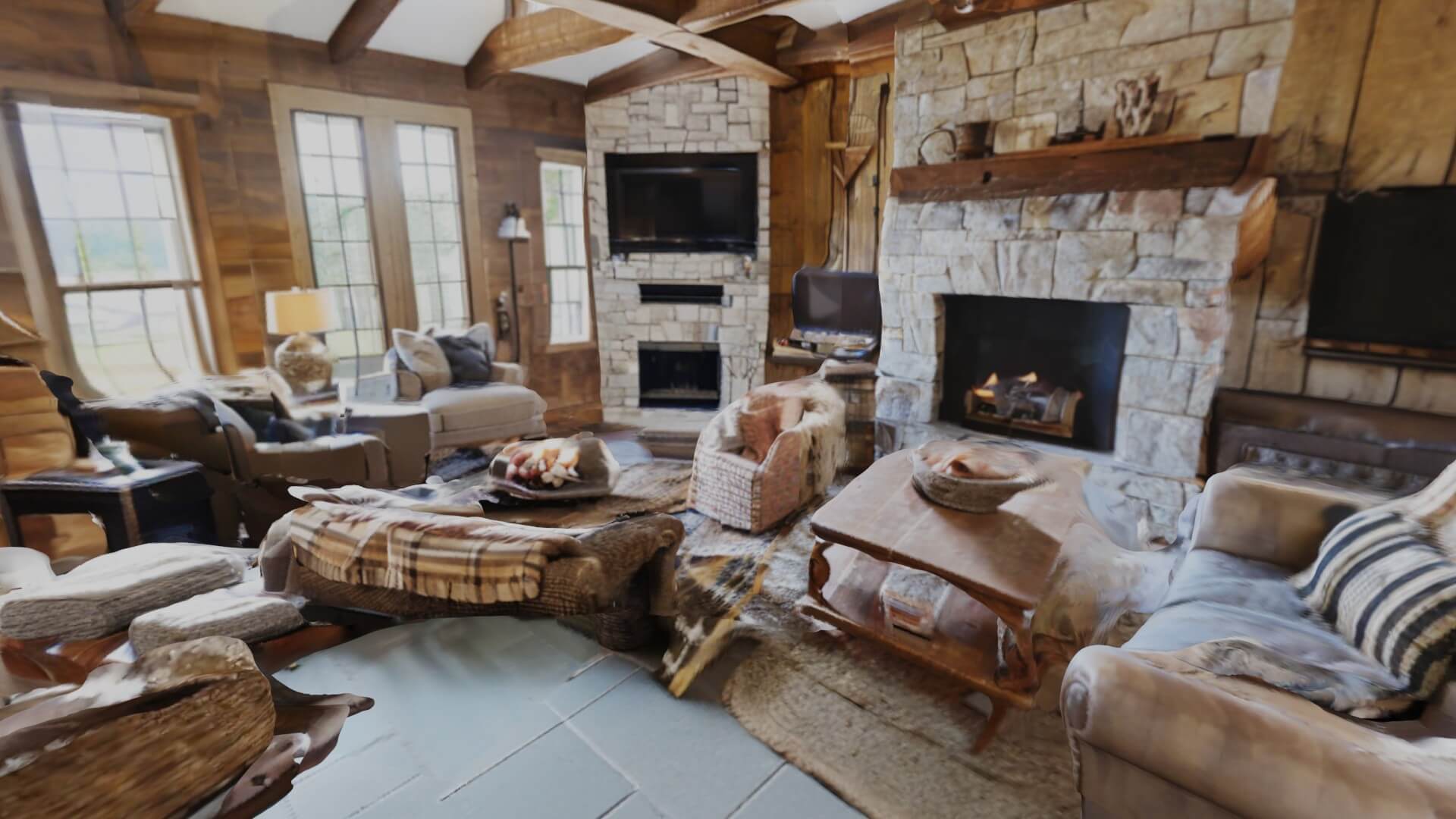} & 
\includegraphics[height=24mm]{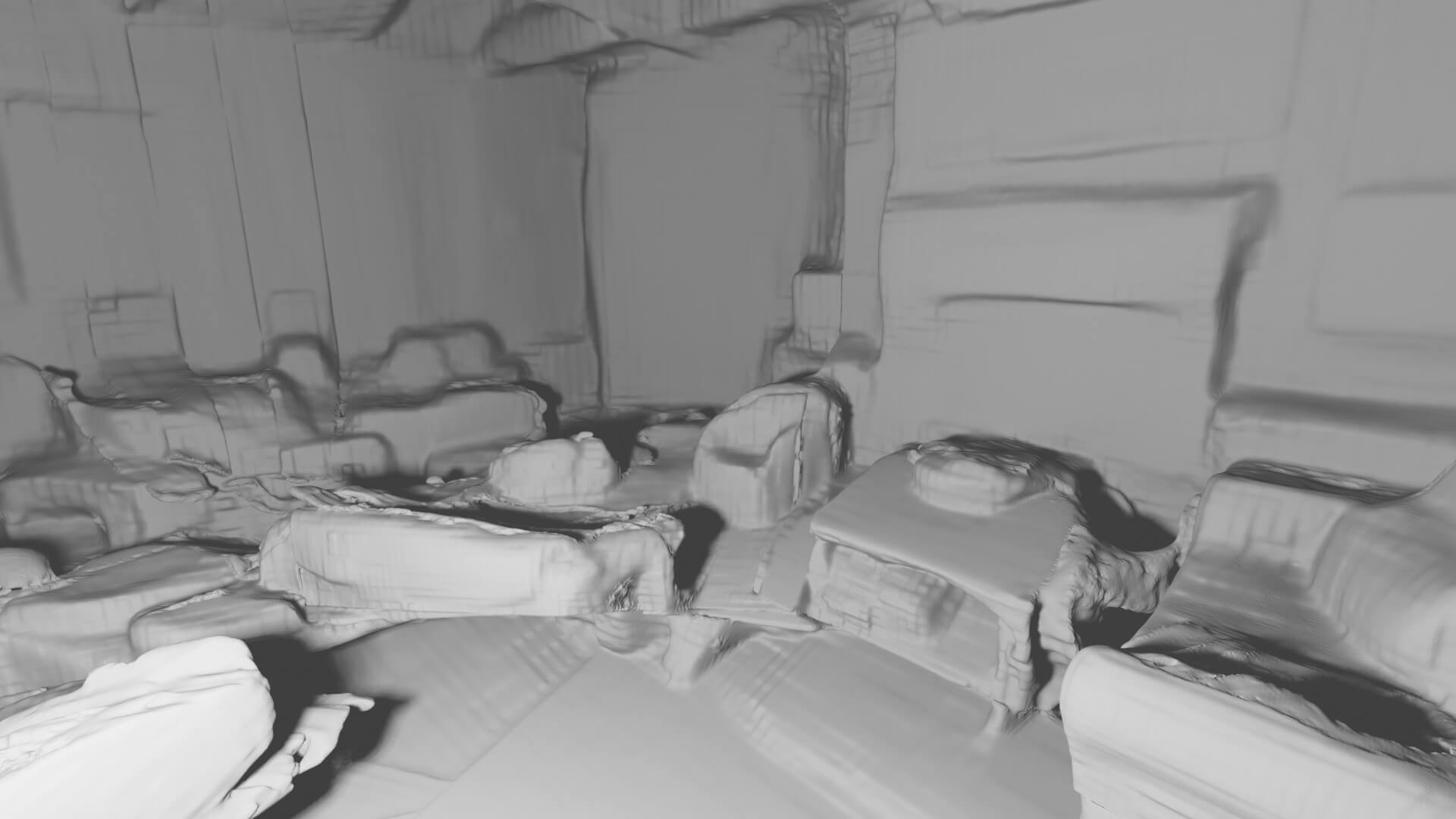} \\ 
& \includegraphics[height=24mm]{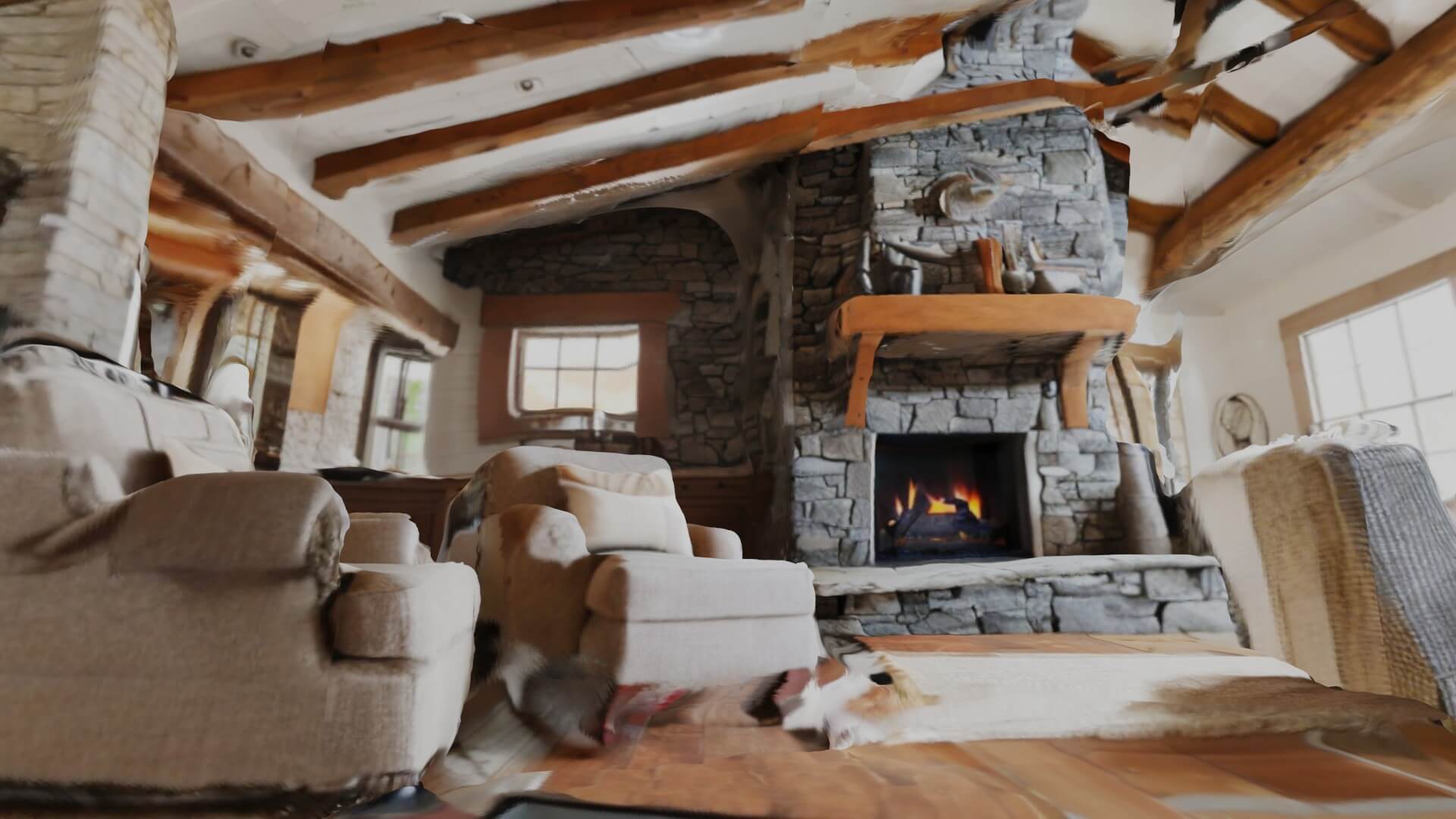} & 
\includegraphics[height=24mm]{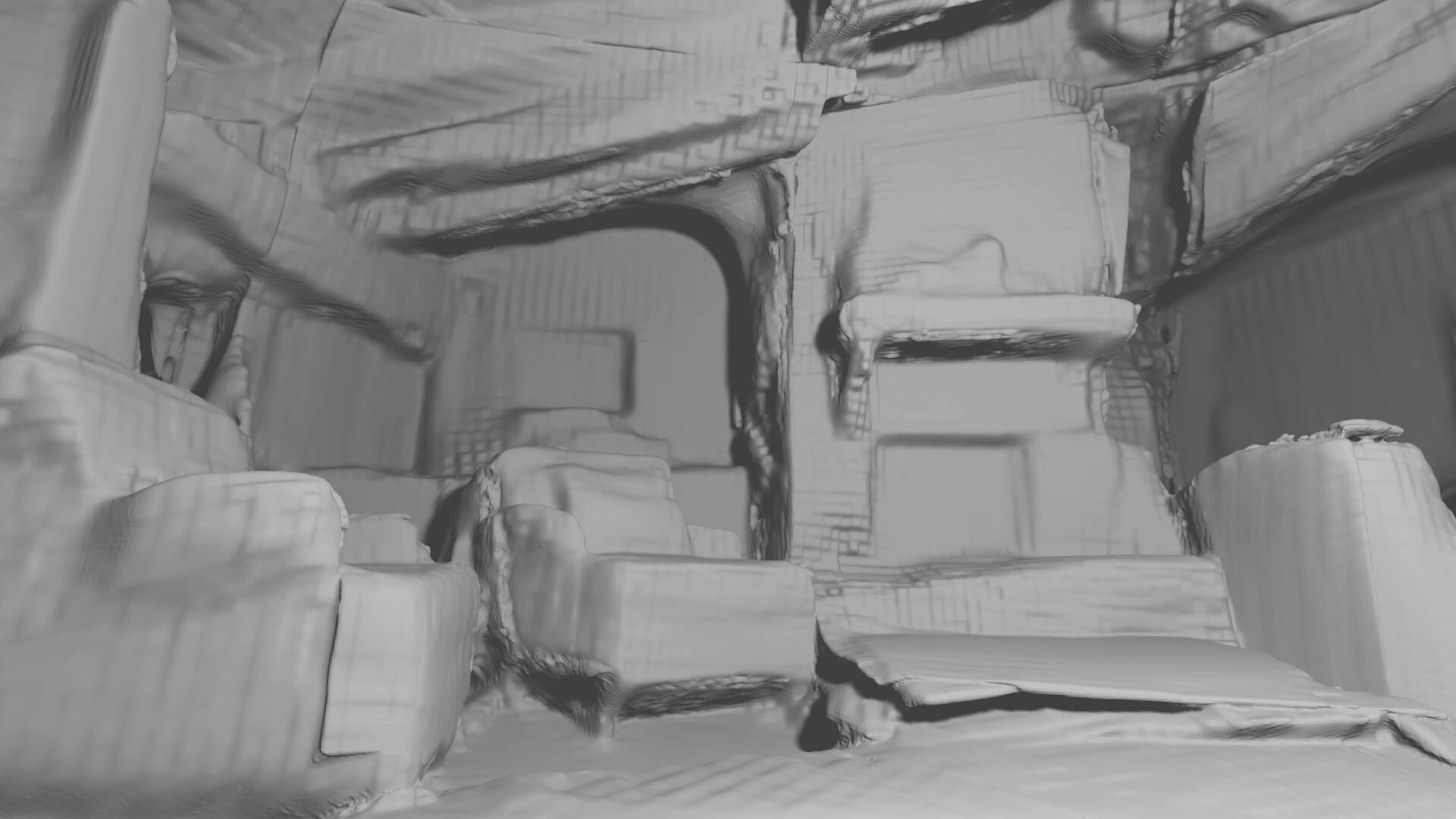} \\ 
\multicolumn{3}{c}{\textit{Editorial Style Photo, Rustic Farmhouse, Living Room, Stone Fireplace, Wood, Leather, Wool}} \\

\multirow{2}{*}[0.8in]{\includegraphics[height=48mm]{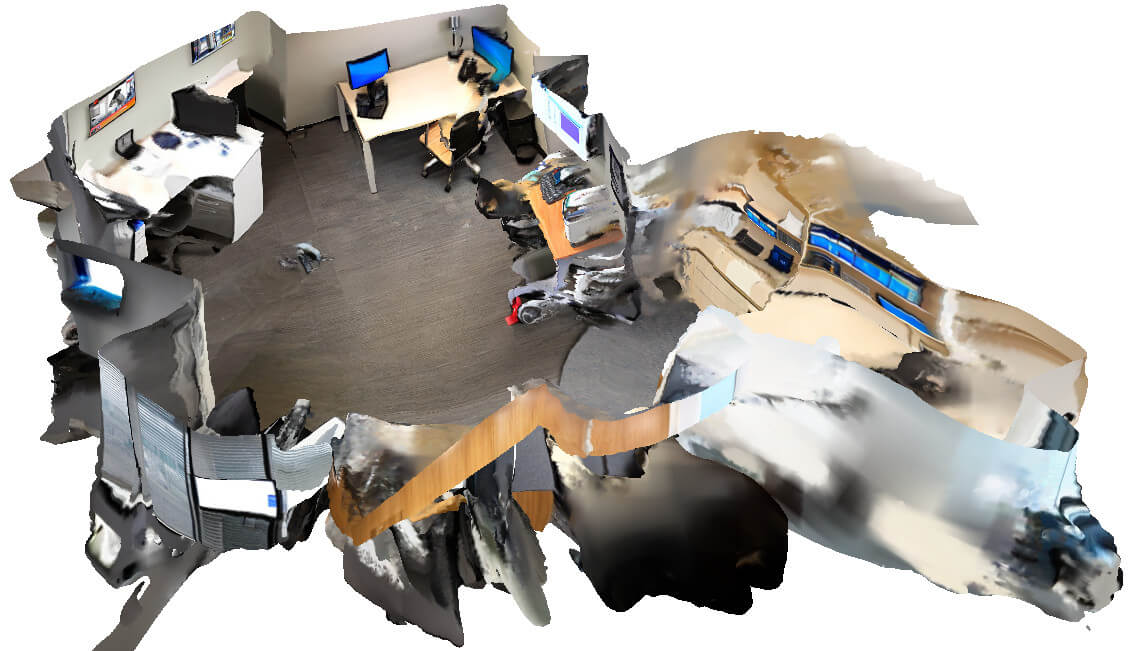}} & 
\includegraphics[height=24mm]{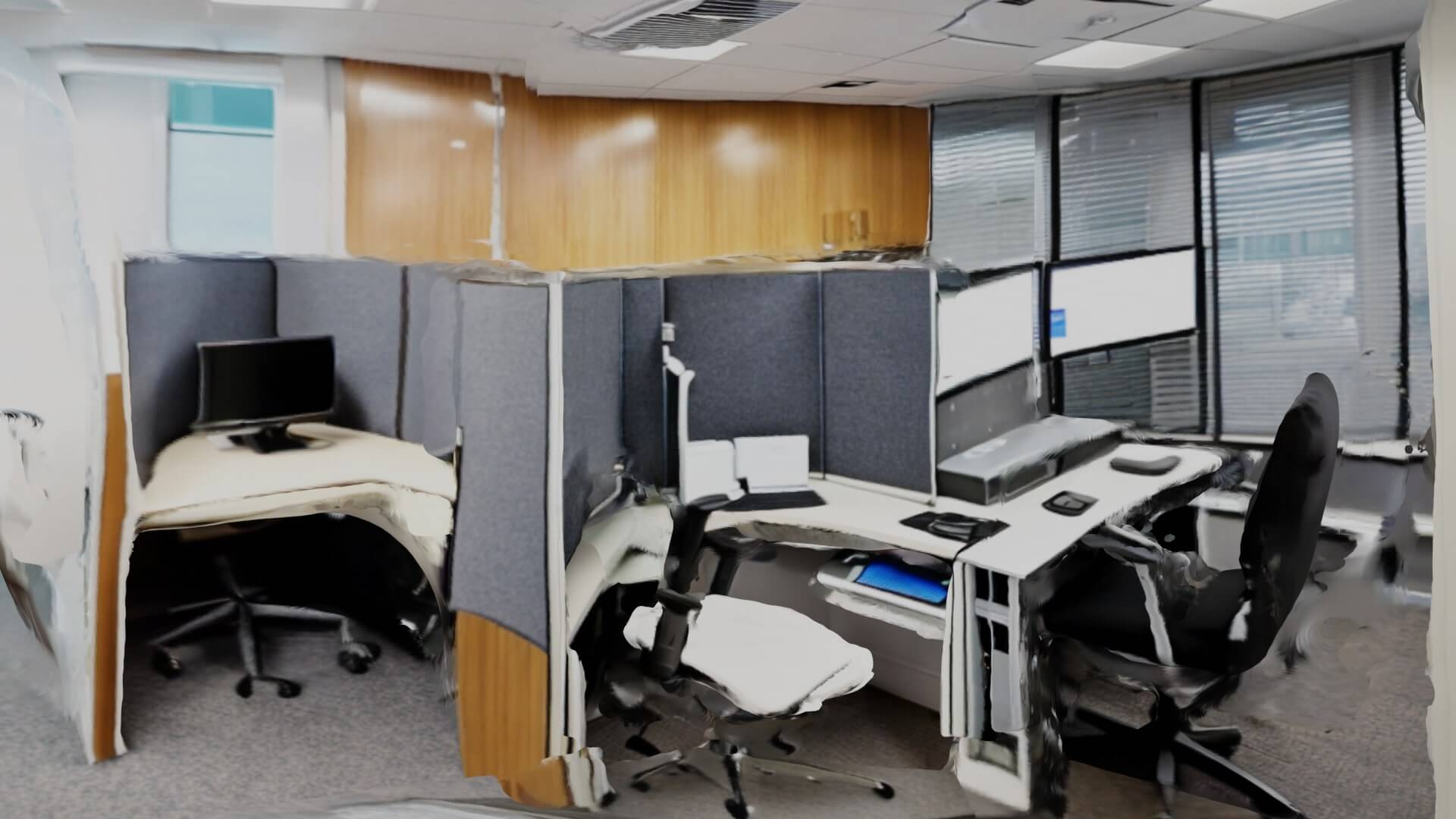} & 
\includegraphics[height=24mm]{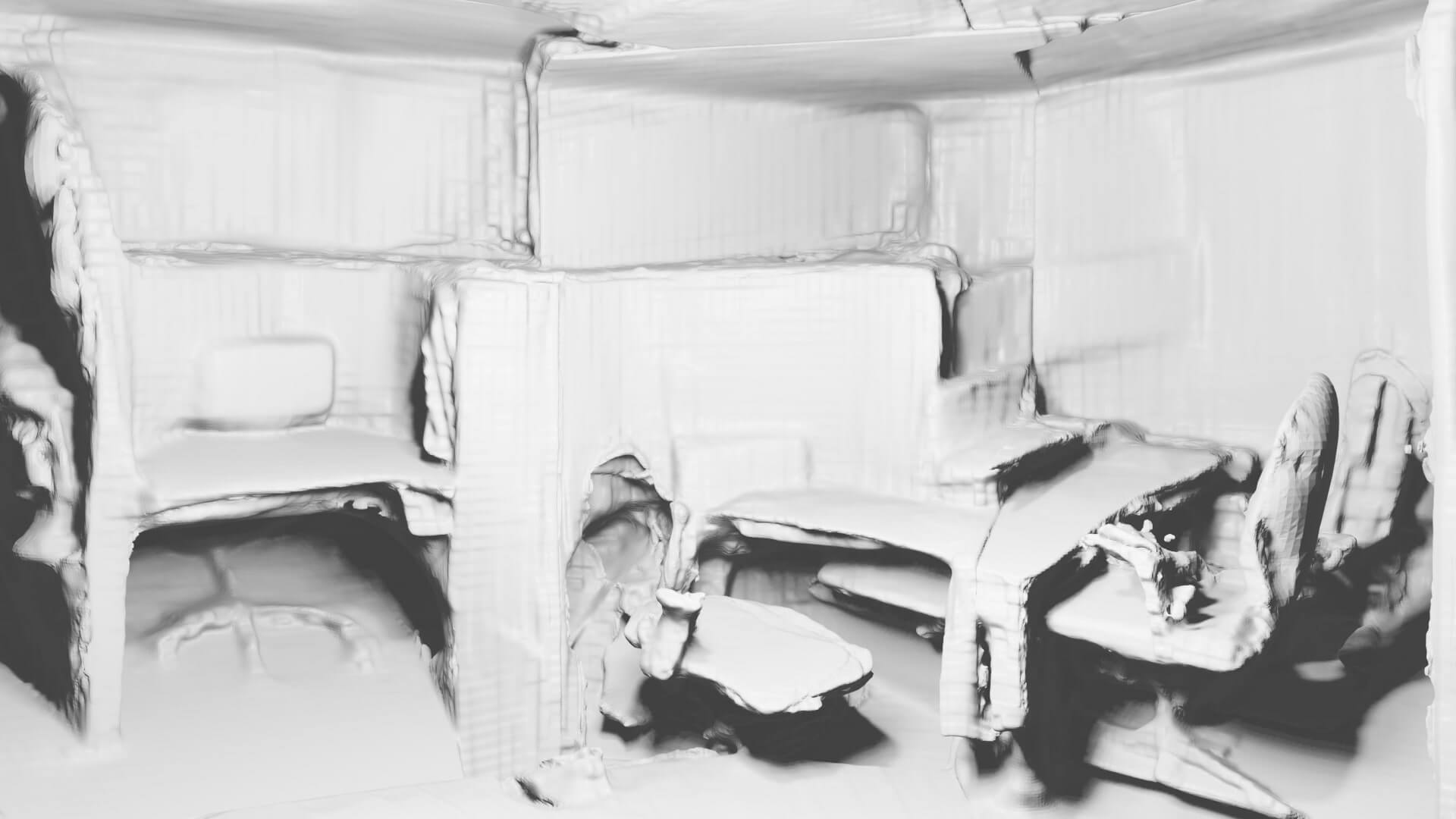} \\ 
& \includegraphics[height=24mm]{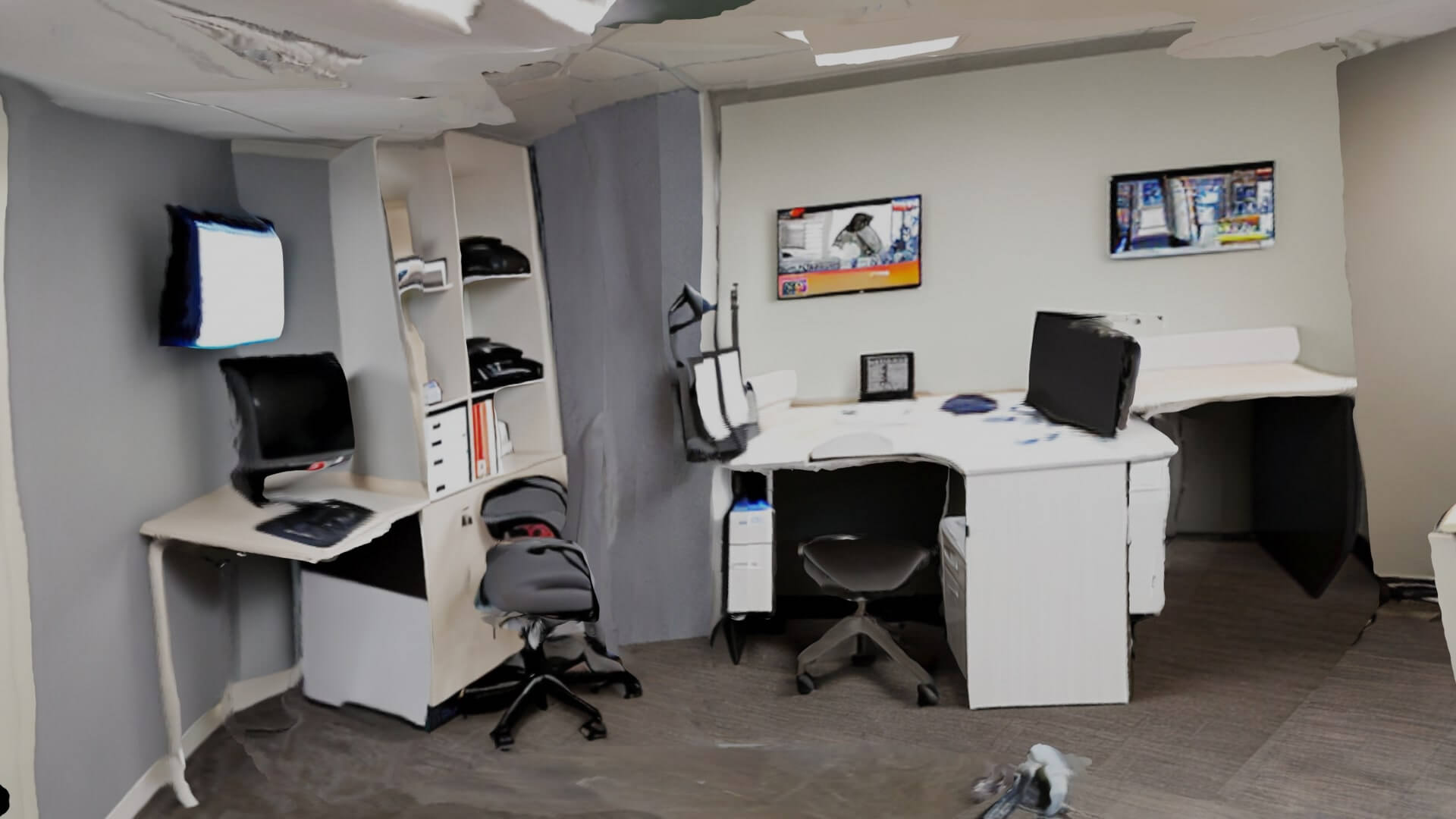} & 
\includegraphics[height=24mm]{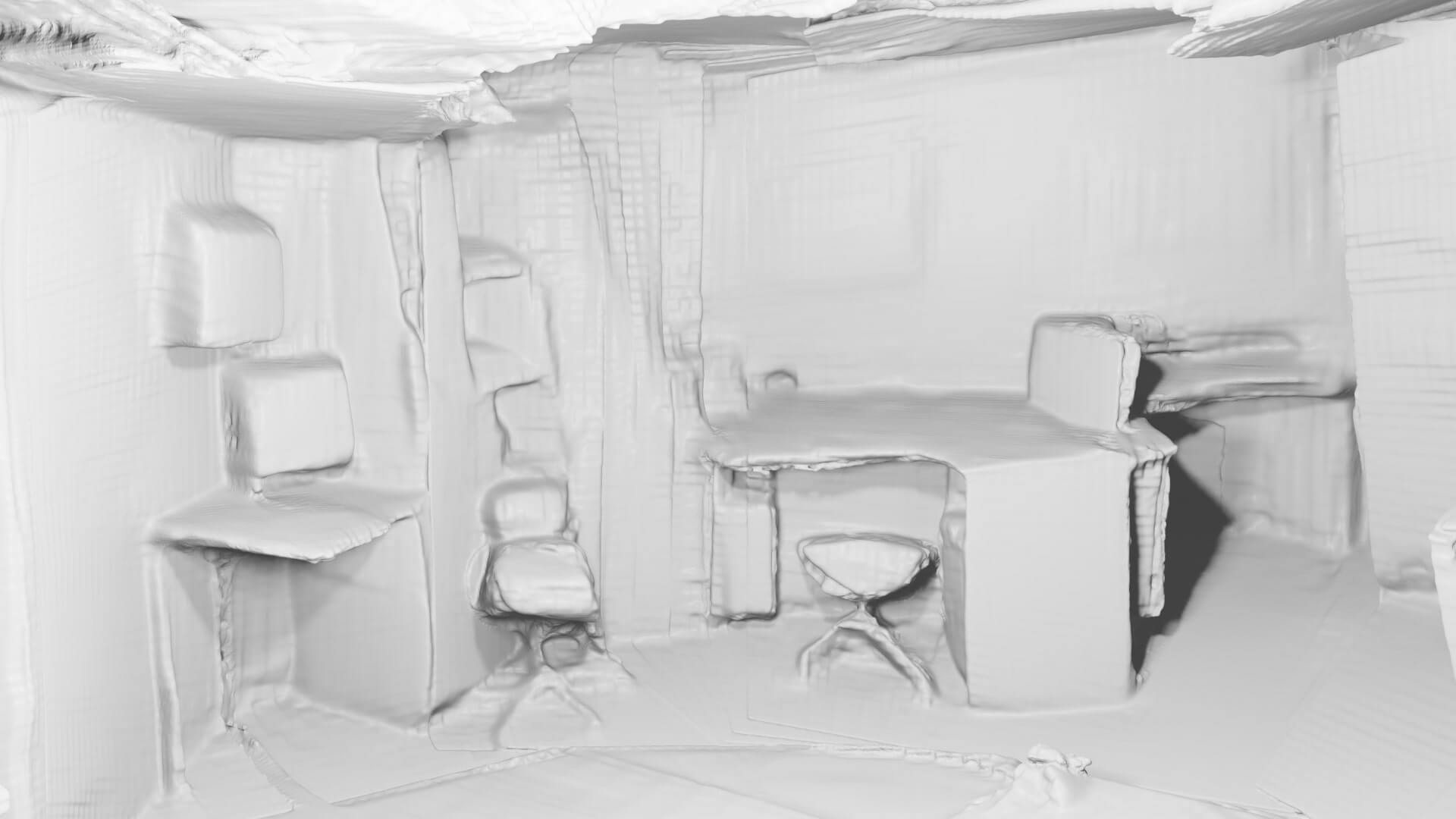} \\ 
\multicolumn{3}{c}{\textit{A small office with a chair, desk and monitors}} \\

\multirow{2}{*}[0.8in]{\includegraphics[height=48mm]{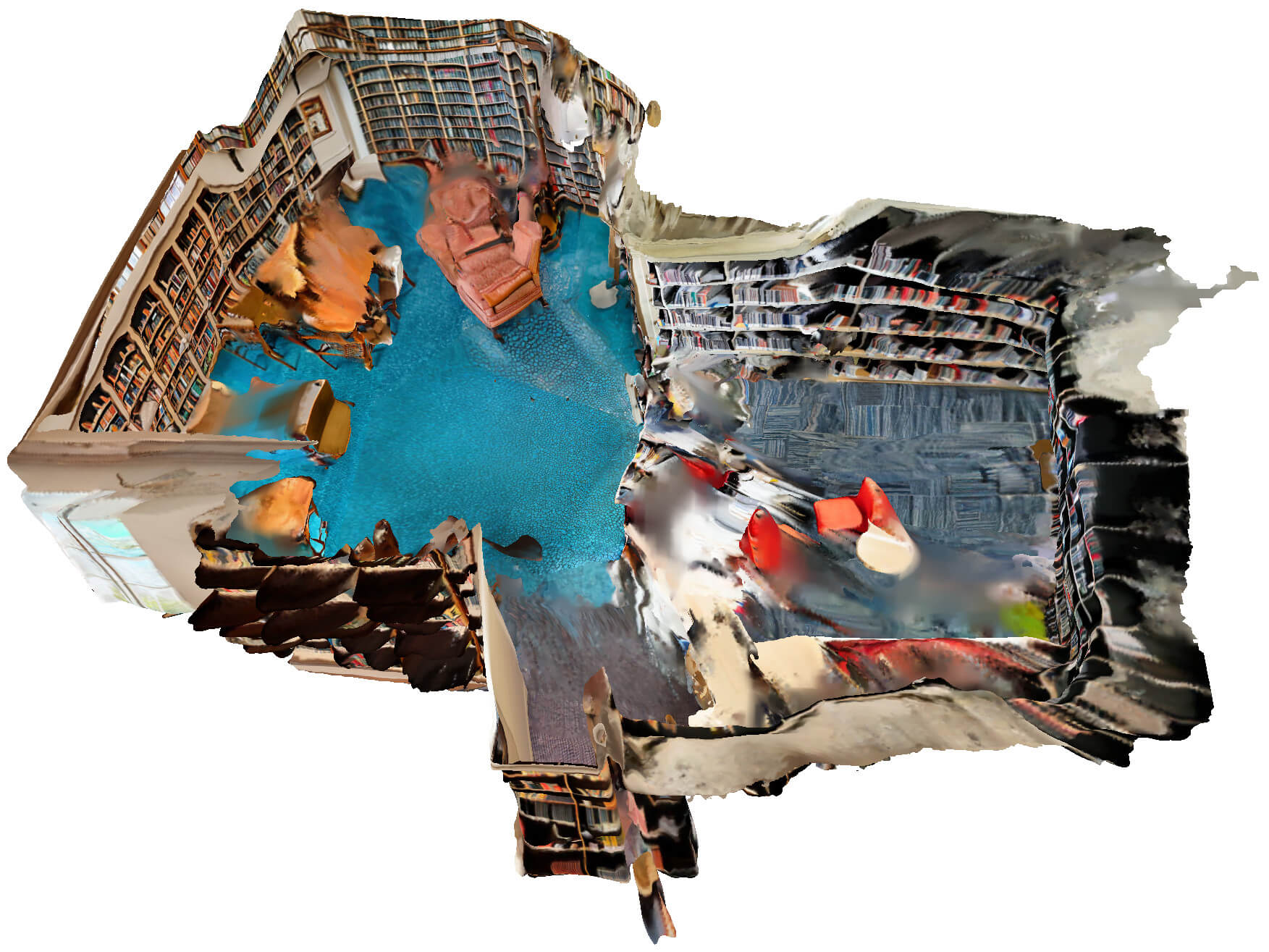}} & 
\includegraphics[height=24mm]{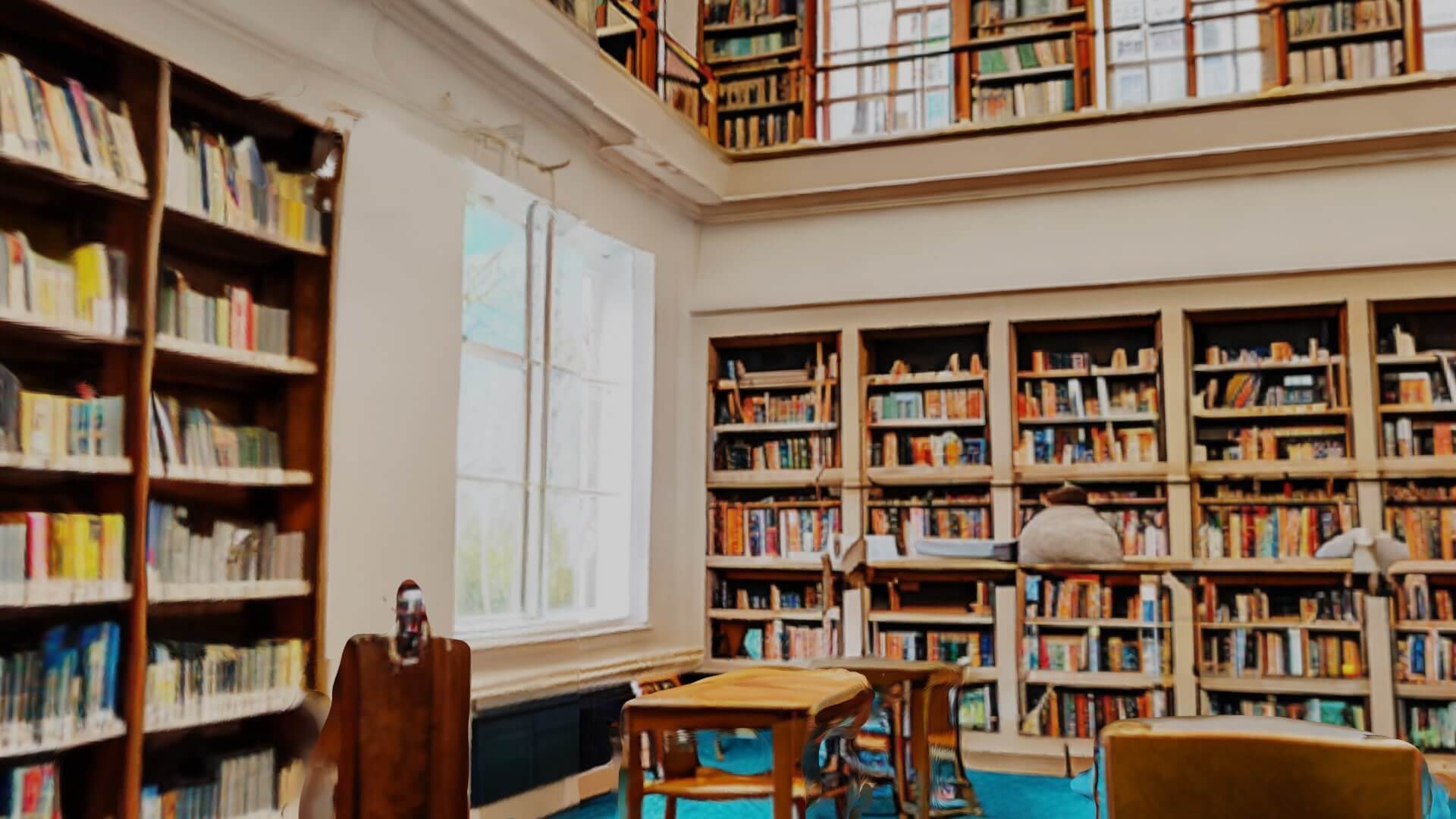} & 
\includegraphics[height=24mm]{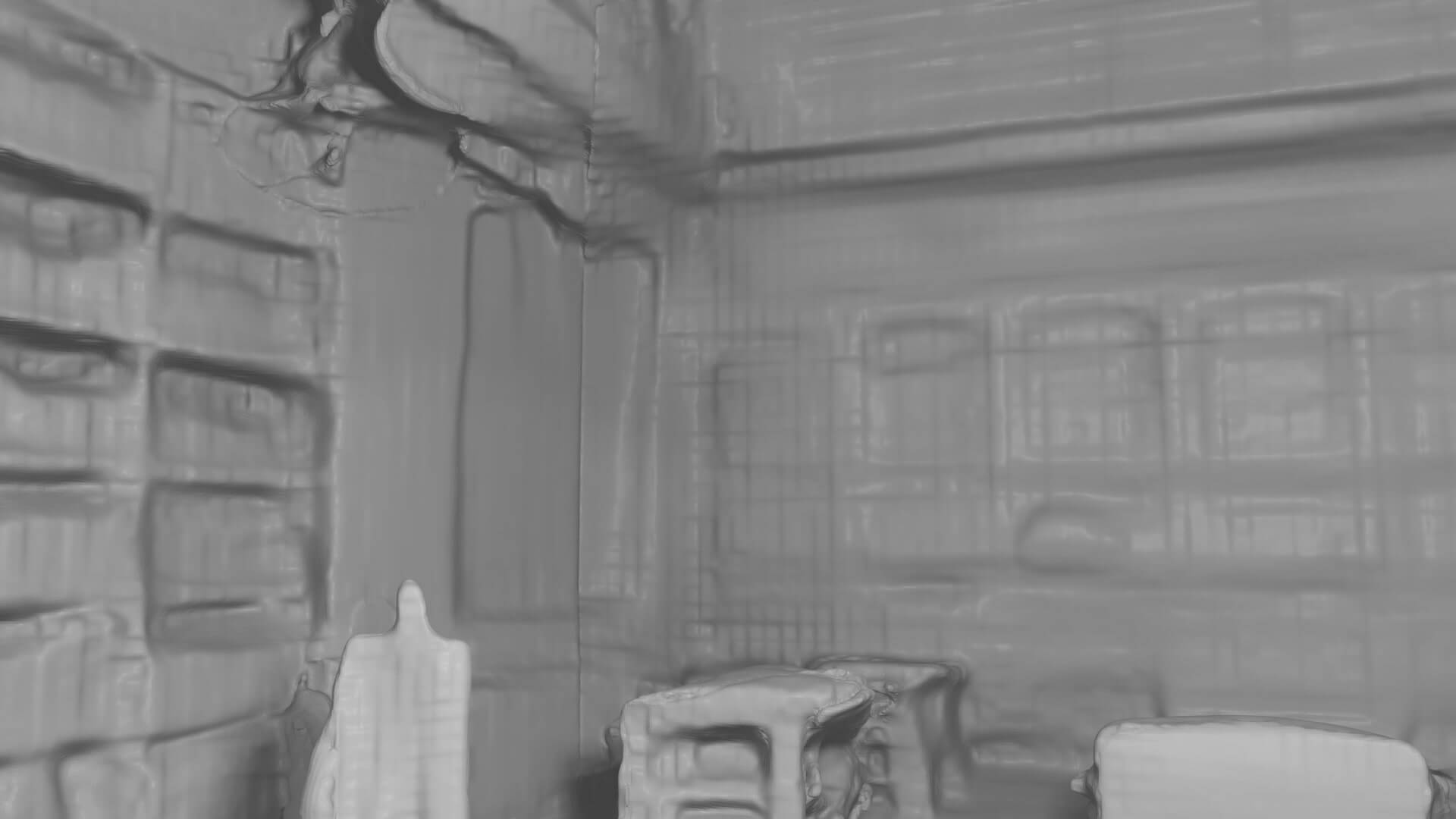} \\ 
& \includegraphics[height=24mm]{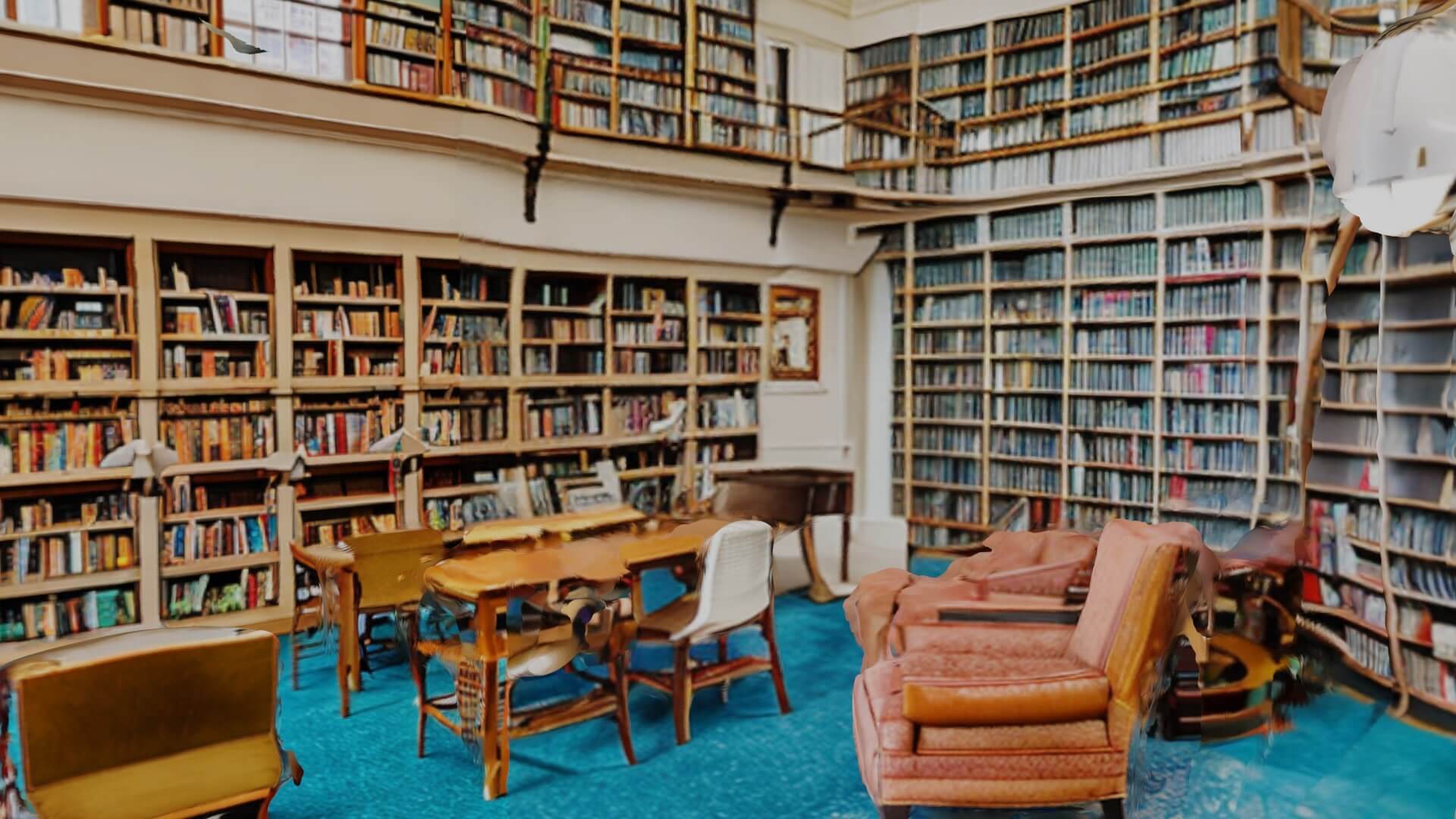} & 
\includegraphics[height=24mm]{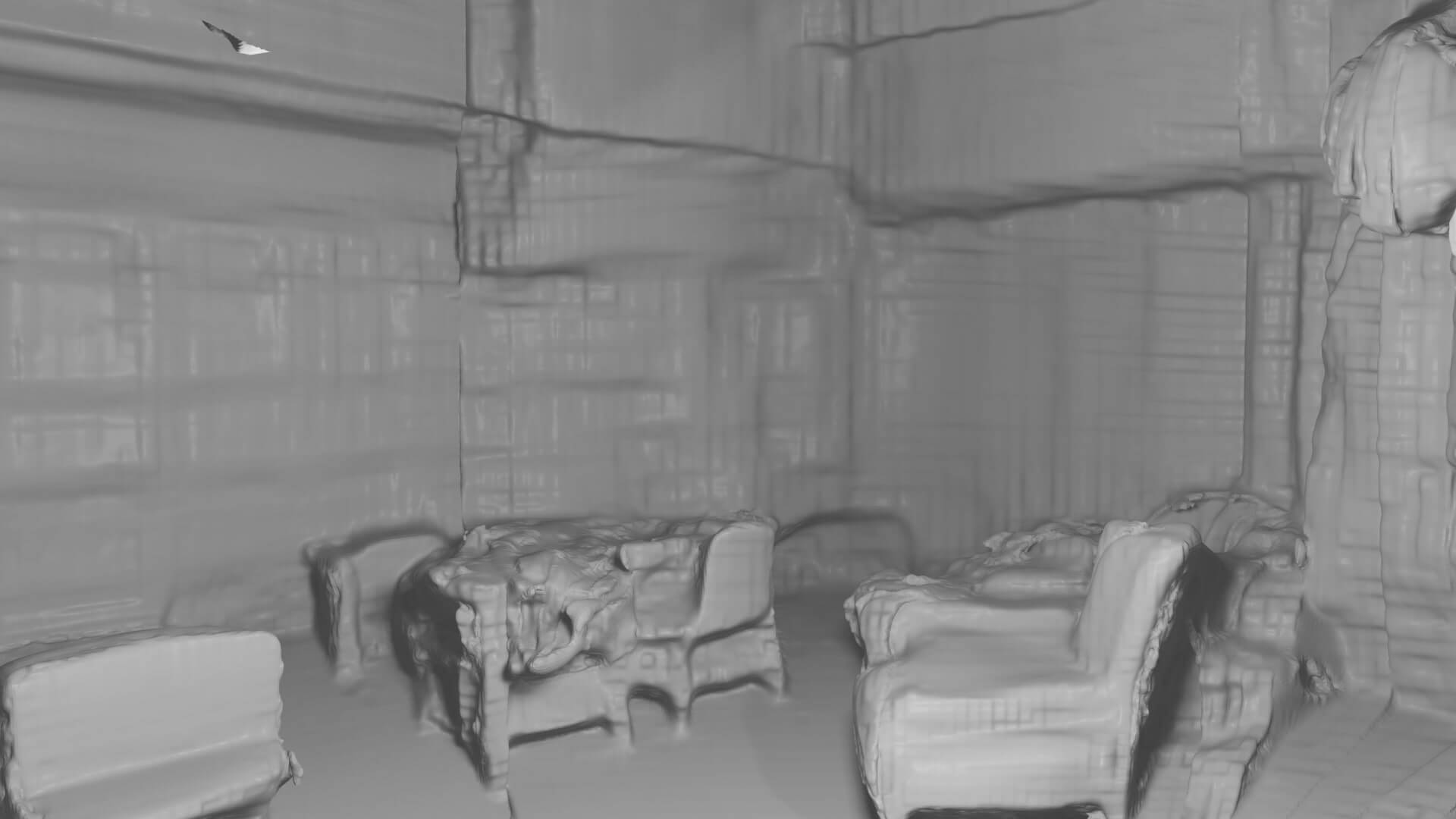} \\ 
\multicolumn{3}{c}{\textit{A library with tall bookshelves, tables, chairs, and reading lamps}} \\

\multirow{2}{*}[0.8in]{\includegraphics[height=48mm]{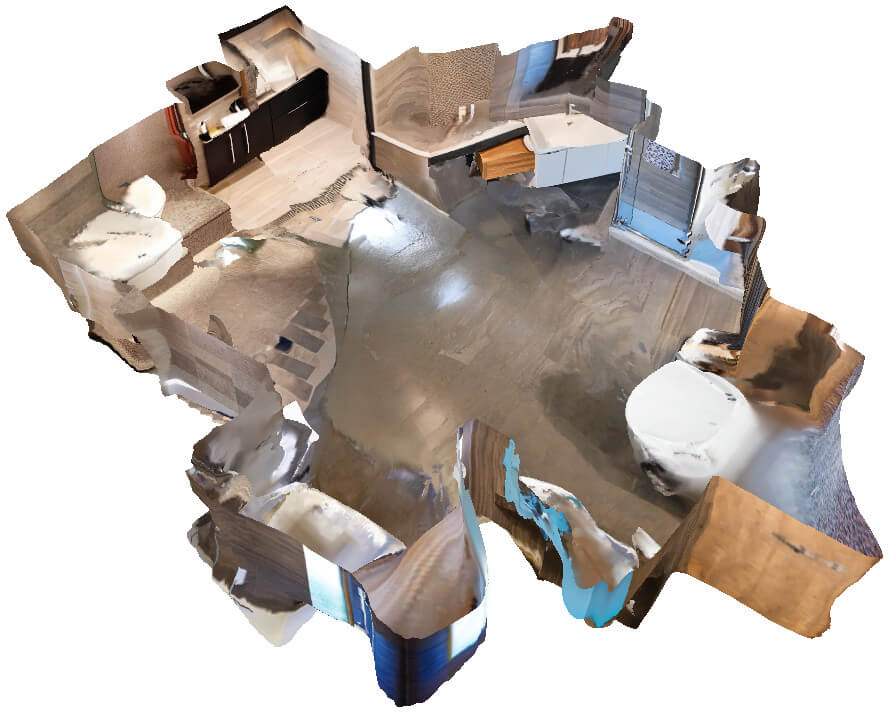}} & 
\includegraphics[height=24mm]{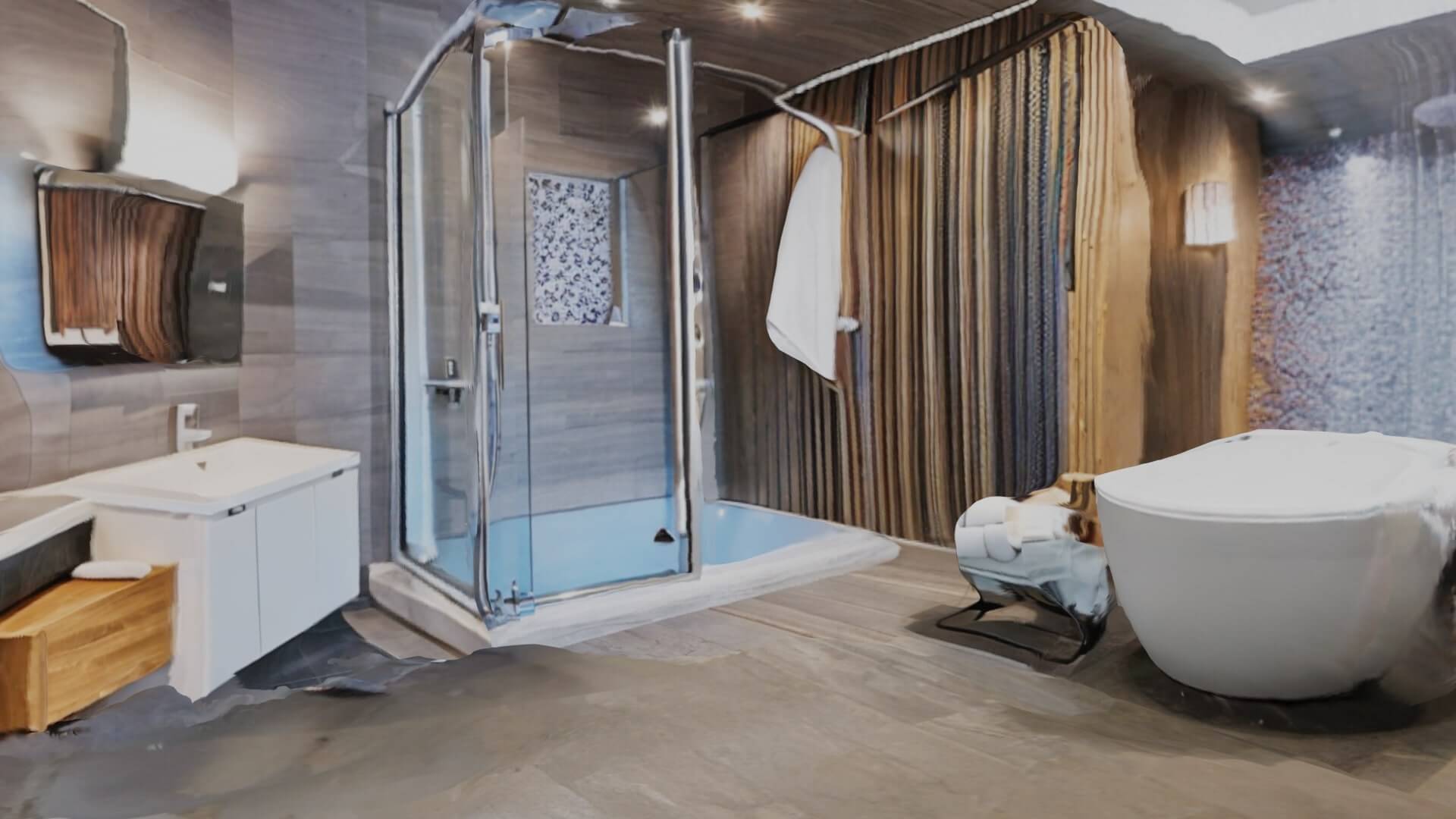} & 
\includegraphics[height=24mm]{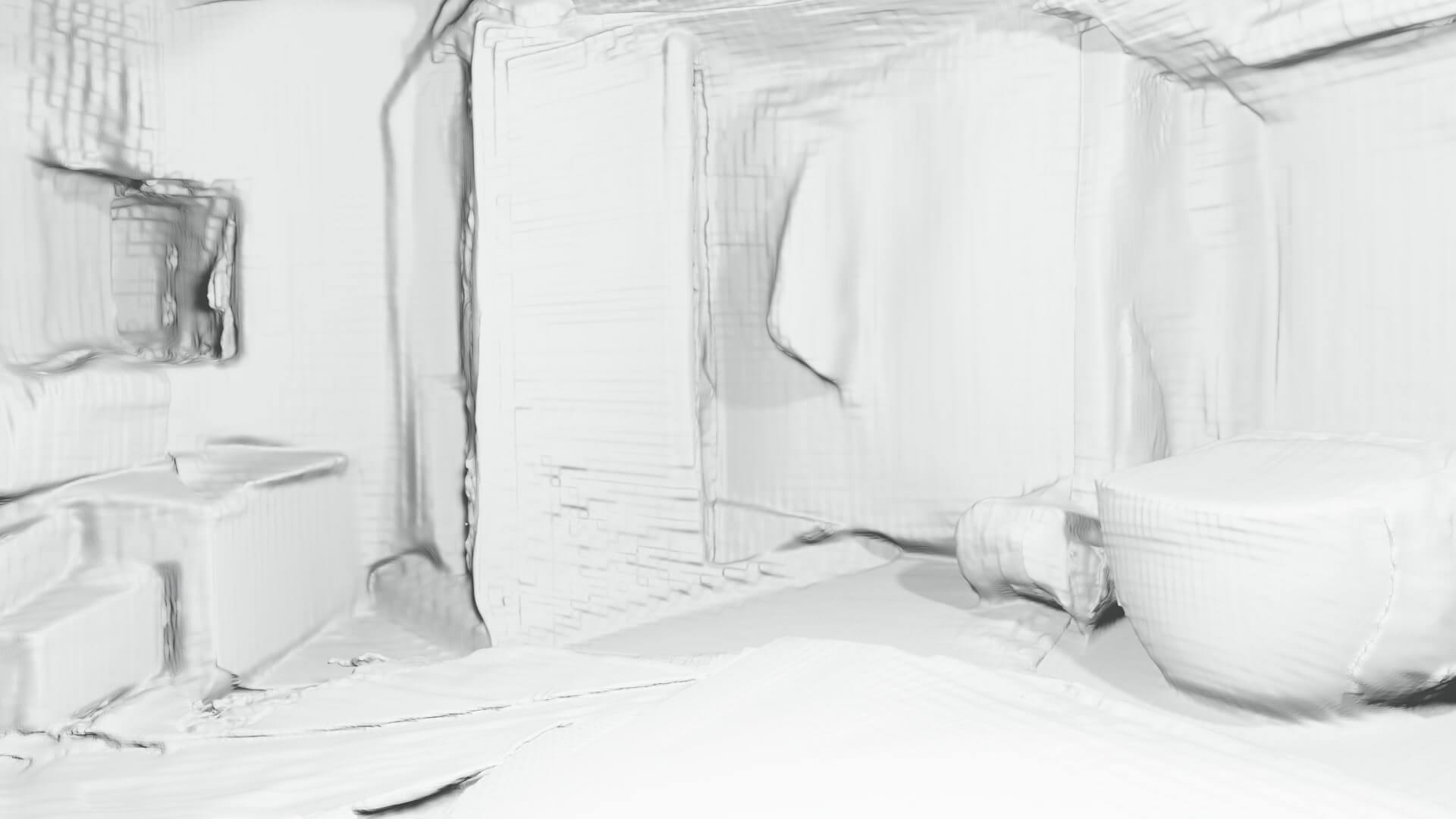} \\ 
& \includegraphics[height=24mm]{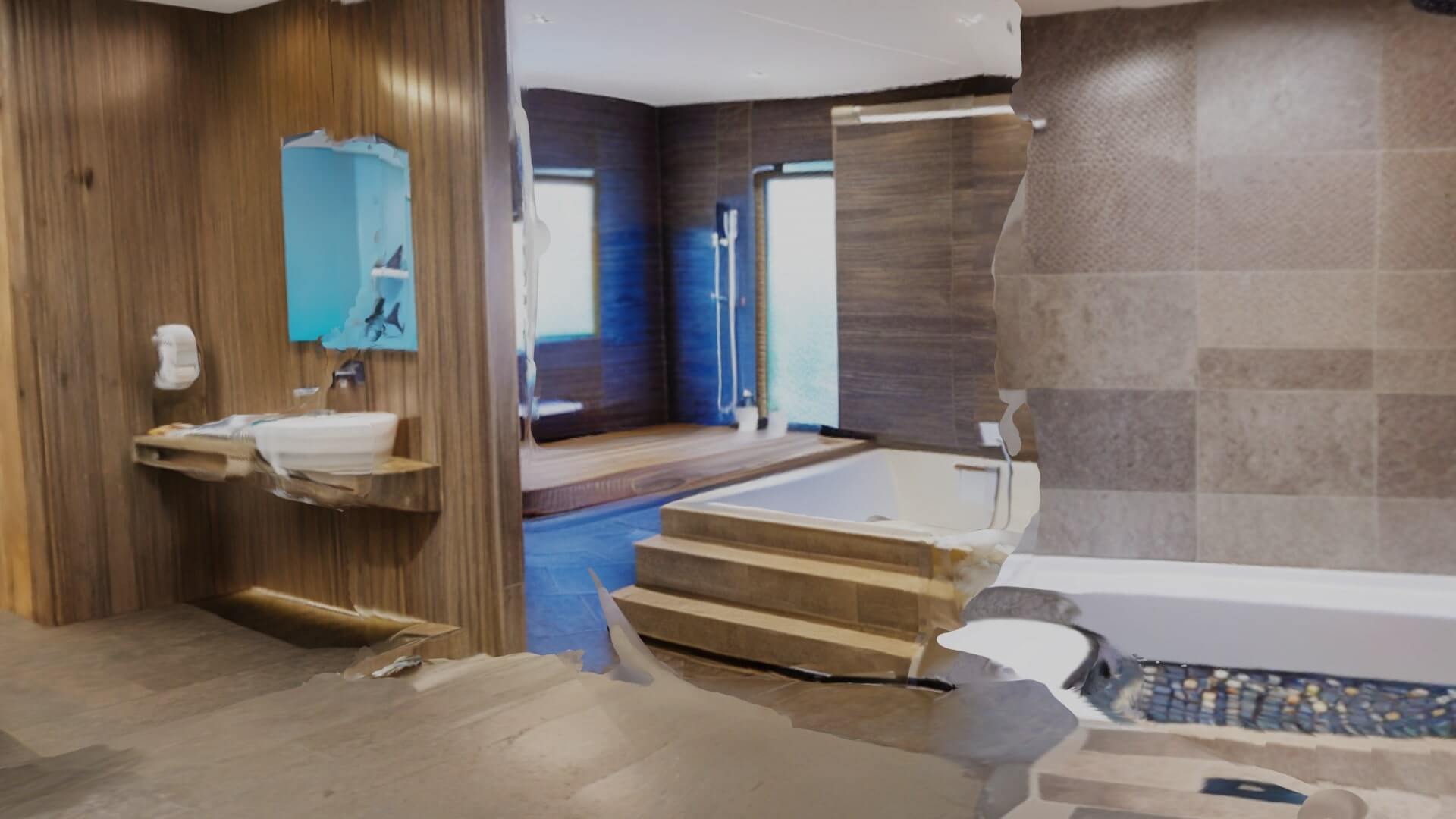} & 
\includegraphics[height=24mm]{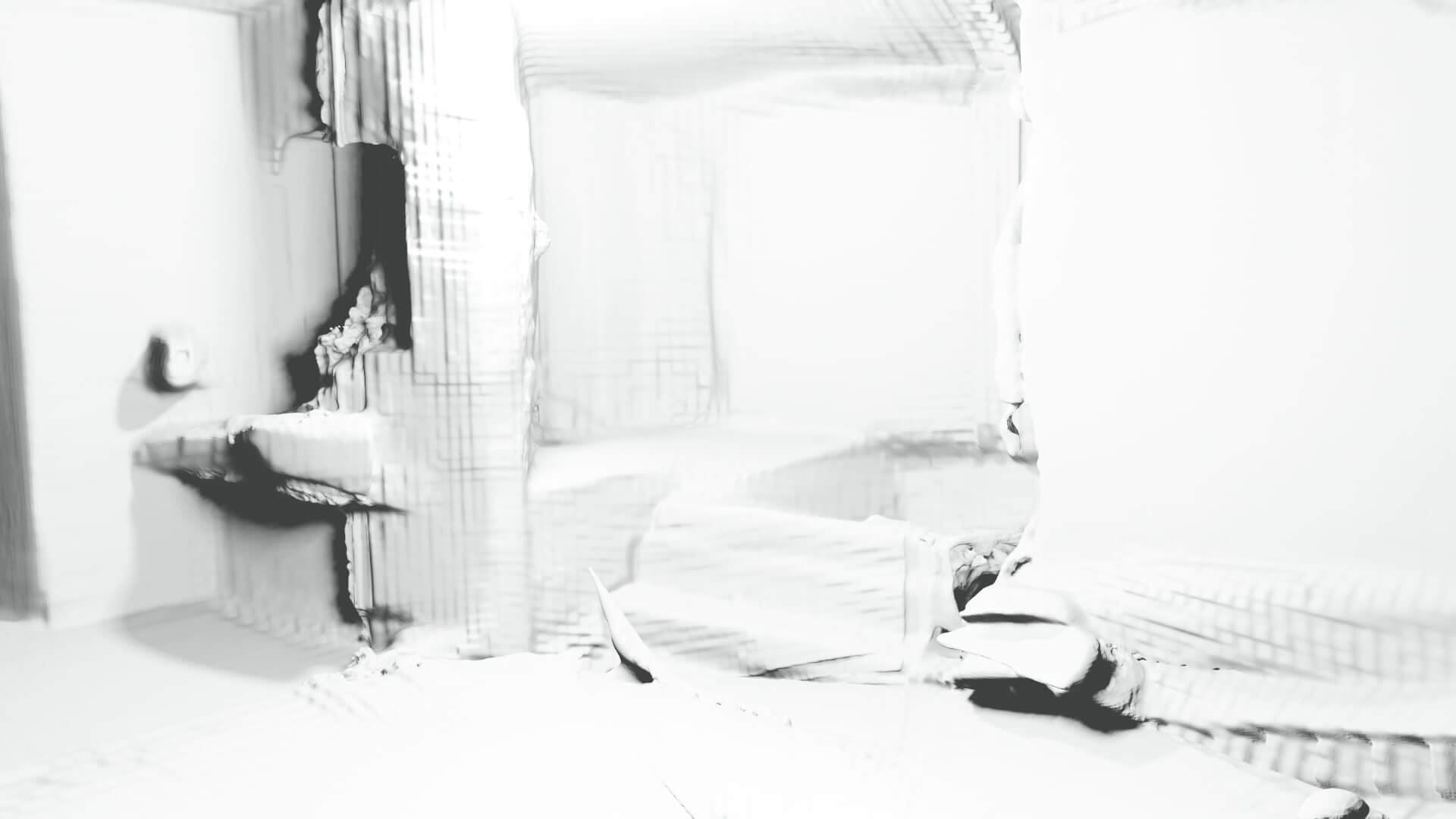} \\ 
\multicolumn{3}{c}{\textit{A large bathroom with shower, bathtub and a cozy wellness area}} \\

\end{tabular}
\caption{
\textbf{3D scene generation results of our method.}
We show color and shaded geometry renderings from generated scenes with corresponding text prompts.
Our method synthesizes realistic meshes satisfying text descriptions. 
We remove the ceiling in the top-down view for better visualization of the scene layout.}
\label{fig:supp-ours-only-final}
\end{figure*}

We show additional qualitative results of our method in Figure~\ref{fig:supp-ours-only-final}.

\end{document}